\documentclass{article}

\usepackage{arxiv}
\usepackage[utf8]{inputenc}
\usepackage{fnpct}
\RequirePackage{amsthm,amsmath,amsfonts,amssymb,dsfont, amstext, mathtools}
\usepackage{verbatim}
\usepackage{pgf,tikz}
\usepackage{mathrsfs}
\usepackage{algorithm2e}
\usepackage{algorithmic}
\usetikzlibrary{arrows}
\usetikzlibrary{decorations.pathreplacing,angles,quotes}

\usepackage{amsthm}
\usepackage{ifdraft}

\usepackage{nicefrac}

\newtheorem{theorem}{Theorem}[section]
\newtheorem*{theorem*}{Theorem}

\newtheorem*{claim*}{Claim}

\newtheorem*{proposition*}{Proposition}

\newtheorem*{lemma*}{Lemma}

\newtheorem*{conjecture*}{Conjecture}

\newtheorem*{fact*}{Fact}

\newtheorem*{hypothesis*}{Hypothesis}

\theoremstyle{definition}

\newtheorem{remark}[theorem]{Remark}

\usepackage{prettyref}
\newcommand{\savehyperref}[2]{\texorpdfstring{\hyperref[#1]{#2}}{#2}}

\newrefformat{eq}{\savehyperref{#1}{\textup{(\ref*{#1})}}}
\newrefformat{lem}{\savehyperref{#1}{Lemma~\ref*{#1}}}
\newrefformat{def}{\savehyperref{#1}{Definition~\ref*{#1}}}
\newrefformat{thm}{\savehyperref{#1}{Theorem~\ref*{#1}}}
\newrefformat{cor}{\savehyperref{#1}{Corollary~\ref*{#1}}}
\newrefformat{cha}{\savehyperref{#1}{Chapter~\ref*{#1}}}
\newrefformat{sec}{\savehyperref{#1}{Section~\ref*{#1}}}
\newrefformat{app}{\savehyperref{#1}{Appendix~\ref*{#1}}}
\newrefformat{tab}{\savehyperref{#1}{Table~\ref*{#1}}}
\newrefformat{fig}{\savehyperref{#1}{Figure~\ref*{#1}}}
\newrefformat{hyp}{\savehyperref{#1}{Hypothesis~\ref*{#1}}}
\newrefformat{alg}{\savehyperref{#1}{Algorithm~\ref*{#1}}}
\newrefformat{sdp}{\savehyperref{#1}{SDP~\ref*{#1}}}
\newrefformat{rem}{\savehyperref{#1}{Remark~\ref*{#1}}}
\newrefformat{item}{\savehyperref{#1}{Item~\ref*{#1}}}
\newrefformat{step}{\savehyperref{#1}{step~\ref*{#1}}}
\newrefformat{conj}{\savehyperref{#1}{Conjecture~\ref*{#1}}}
\newrefformat{fact}{\savehyperref{#1}{Fact~\ref*{#1}}}
\newrefformat{prop}{\savehyperref{#1}{Proposition~\ref*{#1}}}
\newrefformat{claim}{\savehyperref{#1}{Claim~\ref*{#1}}}
\newrefformat{relax}{\savehyperref{#1}{Relaxation~\ref*{#1}}}
\newrefformat{red}{\savehyperref{#1}{Reduction~\ref*{#1}}}
\newrefformat{part}{\savehyperref{#1}{Part~\ref*{#1}}}
\newrefformat{prob}{\savehyperref{#1}{Problem~\ref*{#1}}}
\newrefformat{ass}{\savehyperref{#1}{Assumption~\ref*{#1}}}

\newcommand{\Sref}[1]{\hyperref[#1]{\S\ref*{#1}}}

\usepackage[varg]{txfonts}    

\renewcommand{\mathbb}{\varmathbb} 

\renewcommand{\leq}{\leqslant}

\renewcommand{\geq}{\geqslant}

\usepackage{bm}

\usepackage{xspace}

\usepackage[pdftex,pagebackref,colorlinks,linkcolor=blue,filecolor = blue, citecolor = blue, urlcolor  = blue]{hyperref}

\usepackage{comment}

\usepackage{boxedminipage}

\renewcommand{\vec}[1]{{\bm{#1}}}

\definecolor{DSgray}{cmyk}{0,0,0,0.7}

\newcommand{\V}{\bar{v}}

\RequirePackage{graphicx}
\RequirePackage{color}
\RequirePackage{verbatim}
\RequirePackage{enumitem}  
\usepackage[toc,page]{appendix}
\usepackage{caption}
\usepackage{subcaption}
\usepackage{tabularx}
\usepackage{booktabs}
\usepackage{tablefootnote}
\usepackage{multirow}
\usepackage{makecell}

\definecolor{colour3}{RGB}{178,55,250} 
\DeclareMathOperator{\diag}{Diag}
\newcommand{\A}{\boldsymbol{A}}
\newcommand{\B}{\boldsymbol{B}}
\newcommand{\M}{\boldsymbol{M}}
\newcommand{\I}{\boldsymbol{I}}
\newcommand{\E}{\boldsymbol{E}}
\newcommand{\D}{\boldsymbol{D}}
\renewcommand{\L}{\boldsymbol{L}}
\renewcommand{\H}{\boldsymbol{H}}
\renewcommand{\P}{\boldsymbol{P}}
\newcommand{\Rmatrix}{\boldsymbol{R}}

\newcommand{\bb}{\textcolor{black}{\textbullet} }

\newcounter{noteDSctr} 

\newcounter{noteXDctr} 

\newcounter{noteMCctr} 

\newcommand{\mc}[1]{{\textcolor{black}{{{}}#1}}{}}
\newcommand{\mcc}[1]{{\textcolor{black}{{{}}#1}}{}}

\newcommand{\hk}[1]{{\textcolor{black}{{{}}#1}}{}}

\newcounter{noteHKctr} 

\title{Graph similarity learning for change-point detection \\ in dynamic networks}

\author{
 Deborah Sulem\thanks{Corresponding author.} \\
 Department of Statistics\\
 University of Oxford\\
 \texttt{deborah.sulem@stats.ox.ac.uk} \\
  \And
 Henry Kenlay \\ 
Department of Engineering Science \\
 University of Oxford\\
 \texttt{kenlay@robots.ox.ac.uk} \\
 \And
 Mihai Cucuringu\thanks{Listed alphabetically.} \\
 Department of Statistics, University of Oxford\\
 The Alan Turing Institute, London, UK\\
 \texttt{mihai.cucuringu@stats.ox.ac.uk} \\ 
\And
Xiaowen Dong$^\dagger$ \\ 
Department of Engineering Science \\
 University of Oxford\\
 \texttt{xdong@robots.ox.ac.uk} \\
}

\date{\today}

\usepackage{placeins}

\newcommand{\textbfr}[1]{\textbf{\textcolor{red}{#1}}}
\newcommand{\textbfb}[1]{\textbf{\textcolor{blue}{#1}}}

\begin{document}

\maketitle
                       
\begin{abstract} 
    Dynamic networks are ubiquitous for modelling sequential graph-structured data, e.g., brain connectome, population flows and messages exchanges. In this work, we consider dynamic networks that are temporal sequences of graph snapshots, and aim at detecting abrupt changes in their structure. This task is often termed \textit{network change-point detection} and has numerous applications, such as fraud detection or physical motion monitoring.  Leveraging a graph neural network model, we design a method to perform online network change-point detection that can adapt to the specific network domain and localise changes with no delay. The main novelty of our method is to use a siamese graph neural network architecture for learning a data-driven graph similarity function, which allows to effectively compare the current graph and its recent history. Importantly, our method does not require prior knowledge on the network generative distribution and is agnostic to the type of change-points; moreover, it can be applied to a large variety of networks, that include for instance edge weights and node attributes. 
We show on synthetic and real data that our method enjoys a number of benefits: it is able to learn an adequate graph similarity function for performing online network change-point detection in diverse types of change-point settings, and requires a shorter data  history to detect changes than most existing state-of-the-art baselines.

\end{abstract}



\section{Introduction}

The study of dynamic - or temporal, evolutionary, time-varying - networks has become very popular in the last decade, with the increasing amount of sequential data collected from structured and evolving systems, e.g. online communication platforms \cite{kumar2019}, co-voting networks \cite{wilson2016modeling} or fMRI data \cite{Cribben2017}. In fact, adding a time component to graph-structured data leads to a richer representation and allows more powerful analysis \cite{skarting2021}. This is particularly important when the network is governed by a non-stationary underlying process, which dynamics undergo abrupt switches or breaks. For instance, interaction patterns in social networks, such as Twitter or Reddit, can be very quickly modified after some event or ``shock`` \cite{rossi2020temporal}, \mc{thus providing strong motivation for incorporating a temporal dimension in the analysis}. 
Detecting such structural breaks is a common task in diverse applications, from brain connectivity state segmentation \cite{ondrus2021} to phase discovery in financial correlation networks \cite{Barnett2016}.  Moreover, most real-world dynamic networks are structured around functional groups or densely connected communities. Therefore their evolution over time has been often measured by the changes in these substructures - sometimes called \textit{community life-cycle} \cite{Rossetti_2018} - e.g. growth, \mc{decay}, merges, splits, etc.

For multivariate time series, change-point detection is a task that has been widely studied in various settings (e.g., nonparametric \cite{Zou_2014}, high-dimensional \cite{wang2017highdimensional} or online \cite{Chen2020}). The equivalent task for dynamic networks is often denoted network change-point detection (NCPD) and has recently become a popular problem with numerous successful applications in finance \cite{Barnett2016}, neuroscience \cite{OforiBoateng2019NonparametricAD} or transport networks \cite{yu2021optimal}. Depending on the type of problem at hand, dynamic networks have been represented in multiple ways, e.g., with contact sequences, interval graphs, graph snapshots (see \cite{Holme_2012} for a precise review of concepts, models and applications). In this work, we will consider the discrete representation of time-varying networks or \emph{snapshot networks}: we \mc{denote} a dynamic network  $\mathcal{N}_I = \{G_t\}_{t \in I}$ \mc{to be} a sequence of graph snapshots, where $I$ is an ordered set, chosen as $\mathbb{N}_{>0}$ for simplicity, and each $G_t, t \in I,$ is a (static) graph. We note that each $G_t$ is a general graph, that can be directed and/or have edge weights or node attributes. We define a change-point for the network $\mathcal{N}$ as a timestamp $t \in \mathbb{N}_{>0}$ such that the generative distribution of the graphs before $t$, $(G_1, \dots, G_{t-2}, G_{t-1})$ is different from the one of graphs observed \hk{from} $t$, $(G_{t},  G_{t+1}, \dots)$. When the generative distribution of the graph snapshots, a change-point for a dynamic network sequence is more broadly defined as a timestamp $t$ where a significant shift or deviation can be observed between $G_t$ and the preceding graph snapshots.

In general, a dynamic network may contain multiple change-points and \mc{the tasks of} detecting and localisating the latter therefore correspond to partitioning an observation window $[1,T], T > 0$ into segments $[1,T] = \bigcup_{i=0}^{K-1} [\tau_i, \tau_{i+1}]$ \hk{where }$\tau_0 = 1$ \hk{and} $\tau_{K} = T$, such that the generative distribution of the graph snapshots is stationary on $[\tau_i, \tau_{i+1}]$ for each $i$. Intuitively, each temporal segment $[\tau_i, \tau_{i+1}]$ can be associated with a state of the underlying process, and each change-point $\tau_i$ can be interpreted as a response of the system to an external event. Therefore, NCPD shares some \mc{similarity with the task of} anomaly detection in temporal graphs \cite{enikeeva2021changepoint}. In an online setting, one aim\hk{s to}  detect such change-points  while the graph snapshots are collected, and with minimal detection delay, while in a\hk{n} offline setting, such analysis is conducted \emph{a posteriori} on the whole data sequence. For particular graph generative models, the feasibility of the NCPD task and minimax rates of estimation have been analysed in dynamic random graph models, e.g., Bernoulli networks \cite{Padilla2019, enikeeva2021changepoint, yu2021optimal, wang2020optimal}, graphon models \cite{zhao2019changepoint}, stochastic block models \cite{wilson2016modeling, wang2013locality} and generalized hierarchical random graphs \cite{peel2014detecting}. However, most real-world dynamic networks have heterogeneous properties, e.g. sparsity, edge weights, node attributes or nonlinear dynamics \cite{li2017} - and neither their generative distribution nor the type of change that can happen are known in advance.

Many existing methods for NCPD measure the discrepancy between two subsets of graphs, and rely on a graph similarity function, kernel or distance for pairwise graph comparisons \cite{chu2018sequential, Cribben2017, zhao2019changepoint, gretton2008kernel}. However, it is often difficult to choose \emph{a priori} an appropriate measure of similarity (or dissimilarity) that can integrate all the network characteristics, \mc{while being agnostic to the generating mechanism or type of change-point}.  Consequently, without any domain knowledge, this choice is often arbitrary, and result in poor performances \cite{chu2018sequential, enikeeva2021changepoint, Kriege_2020}. Moreover, most online NCPD methods require \hk{finely tuning} several hyperparameters, such as detection thresholds \cite{yu2021optimal} and window sizes \cite{Huang2020}. To address these challenges, we propose a change-point agnostic and end-to-end method for online NCPD that in particular includes learning a data-driven 
graph similarity function.
Our method is therefore adaptive to the network distribution and different types of change-points; in particular, it can easily incorporate general graph features such as node  attributes,  edge weights or attributes,  and can adapt to sparse settings.
In summary, our contributions are the following: 
\begin{itemize}
    \item We propose a graph similarity learning model based on a siamese graph neural network \mcc{able to handle any available node attributes,} and demonstrate how it can be \mc{leveraged for the online NCPD problem} with an adequate training procedure. In particular, our learnt similarity function 
    is sensitive to both local and global displacements in the graph structure, and can effectively be employed in the context of change-point (and anomaly) detection in temporal networks.
    
    \item We use an efficient  online NCPD statistic with a short-term history of the graph snapshots that avoids detection delays and requires little additional hyperparameter tuning.
    
    \item We empirically demonstrate the advantages of our method on synthetic networks with diverse types of change-points, as well as on two challenging real-world data sets.    We notably design a self-supervised training procedure for data without ground-truth labelling of change-points. 
\end{itemize}

\paragraph{Paper outline.}  In Section \ref{sec:rel_work}, we succinctly review existing work on NCPD and present our general setup and methodology in Section \ref{sec:methodology}. In Section \ref{sec:experiments}, we evaluate our method on synthetic and real-world data sets and compare to several existing NCPD baseline methods. Finally, we conclude in Section \ref{sec:discussion} with a summary of our results and discuss possible future developments.

\section{Related works}\label{sec:rel_work}

The study of dynamic networks, and in particular NCPD, is a relatively recent area of research that has largely incorporated principles from change-point detection in time series, \mc{especially in high-dimensional settings}. Some NCPD methods estimate the parameters of a network model, e.g., the generalised hierarchical random graph \cite{peel2014detecting}, a stochastic block model \cite{deridder2016} or the preferential attachment model \cite{bhamidi18}, and conduct hypothesis tests to detect changes in the estimated parameters. Other methods maximize a penalized likelihood function, e.g., based on a non-homogeneous Poisson point process model \cite{Corneli2018MultipleCP} or a dynamic stochastic block model \cite{wilson2016modeling, bhattacharjee2020change}. However, for real-world networks, the assumption on a particular model can sometimes be too restrictive.

Several model-agnostic methods for NCPD extract features from the graph snapshots, e.g., the degree distribution \cite{miller2018size} or the joint distribution of a set of edges \cite{wang2017fast}, and use classical discrepancy measures to quantify the amount of change. Other methods relying on pairwise comparison of graphs use a graph similarity or pseudo-distance, such as the DeltaCon metric \cite{Koutra2016DeltaConPM}, the Hamming distance and the Jaccard distances \cite{donnat2018tracking}, the Frobenius and maximum norms \cite{Barnett2016}, spectral distances based on the Laplacian \cite{Huang2020, Cribben2017, hewapathirana2019change}, $\ell_2$ or  $\ell_{\infty}$ norms \cite{zhao2019changepoint} or a graph kernel \cite{desobry2005online, gretton2008kernel, harchaoui2009kernel}. Nevertheless, these graph metrics suffer from intrinsic limitations; e.g., the Hamming distance is sensitive to the graph density and the  Jaccard distance treats all edges uniformly \cite{donnat2018tracking}. Furthermore, it has been previously underlined that the choice of graph distance can significantly affect a method's results \cite{Barnett2016}, and therefore requires \textit{a-priori} knowledge or assumption on the network distribution.

One widely popular statistic in change-point detection problems is the cumulative sums (CUSUM) statistic, which has been used in different time series contexts, e.g., in the offline and high-dimensional setting (in combination with the network binary segmentation algorithm) \cite{wang2017highdimensional}, and more recently, in the online setting \cite{Chen2020}. Several NCPD methods have adapted this efficient statistic to dynamic networks, e.g., for sparse graphs \cite{wang2020optimal}, graphs with missing links \cite{dubey2021online, enikeeva2021changepoint}, in offline \cite{Padilla2019} and online \cite{yu2021optimal} settings, and proved that minimax rates of estimation can be obtained for the overall false alarm probability and the detection delay.
However, computing the CUSUM statistic necessitates a ``forward'' window to detect a change at a given timestamp, and methods based on this statistic often require to tune several hyperparameters (e.g., one or several detection thresholds).

In addition to the aforementioned limitations, most previously cited methods do not provide a principled way to incorporate node attributes or even edge weights. Interestingly, \mc{to the best of our knowledge, no prior work has ever considered graph neural networks (GNNs) for the NCPD problem,  
despite the fact that such architectures can easily handle different types of networks (e.g., signed \cite{derr2018signed} or directed \cite{huang2019signed} ), and in particular, can inherently account for any available node attributes} \cite{kipf2016semi}. 
In dynamic network modelling, graph convolutional recurrent networks \cite{gcrn2018} and dynamic graph convolutional networks \cite{dgcn2020} were introduced for predicting graph-structured sequences. In the dynamic link prediction task, methods that learn representations of dynamic networks have been proposed, using deep temporal point processes \cite{trivedi2018dyrep}, joint attention mechanisms on nodes neighborhoods and temporal domain \cite{ sankar2020}, memory feature vectors in message-passing architectures \cite{rossi2020temporal} or recurrent neural networks \cite{zhang2021dynamic, kumar2019}. For anomalous edge detection in dynamic graphs, \cite{cai21} process subgraphs around the target edges through convolution and sort pooling operations, and gated recurrent units. 
To our knowledge, only one prior work has incorporated GNN layers in a method for change-point detection, \mc{but has done so} in the context of multivariate time series \cite{zhang2020correlationaware}. However, in this method, the GNN encodes the cross-covariances between the time series' dimensions in the spatial layers, and is one part of a complex neural network architecture (the temporal dependencies being encoded by recurrent neural network layers).  

Moreover, while GNNs have proved to effectively learn representations of graphs, they can also be leveraged to learn graph similarity functions in a data-driven way and for particular tasks in a end-to-end fashion. This now popular problem is called \emph{graph similarity learning (GSL)} \cite{ma2020deep}. One common type of \hk{model} for this task is siamese networks \cite{Koch2015SiameseNN}, e.g., siamese graph neural networks \cite{ma2019similarity}) or graph matching networks \cite{Li2019GraphMN, ling2021}. These architectures allow to learn flexible and adaptive similarity functions and have been successfully applied to several tasks and graph domains, e.g. \mc{classification} of brain networks \cite{ma2019similarity, liu2019communitypreserving, ktena2017distance}, image \hk{classification} \cite{mensink2012metric}, and  detection of vulnerabilities in software systems \cite{Li2019GraphMN}.  
In this work, we will leverage such GSL models for the online NCPD task, which avoids the need for choosing \emph{a-priori} a particular graph distance, kernel or embedding.

\section{General setup and \mcc{framework}}\label{sec:methodology}

In this section, we describe our general set-up and NCPD method based on a graph similarity learning model. We will first present our network change-point statistic in Section \ref{sec:sim_NCPD}, leveraging a similarity function learnt by a GSL model described in Section \ref{sec:sGNN}, through an adequate training and validation procedures (see Section \ref{sec:training}). Before presenting our methodology, we introduce some useful notation.  

\paragraph{Notation.} 
We denote $G = (\A, \E) \in \mathbb{G}$ a graph with $n \geq 1$ nodes denoted by $\{u_1, \dots, u_n\}$, adjacency matrix $\A \in \mathbb{R}^{n \times n}$ and node attributes (or features) matrix $\E \in \mathbb{R}^{n \times d} \cup \{\emptyset\}$, $d\geq 1$. We say that the graph is attributed if $\E \neq \emptyset$, and \hk{un}attributed otherwise. If  $\A \in \mathbb{R}_{\geq 0}^{n \times n}$, we also say that the graph is unsigned. 
 We denote $\mathcal{N}_T = \{G_t\}_{1 \leq t \leq T}$ a dynamic network with $T \geq 1$ snapshots, where each graph $G_t$ has the same set of nodes with a fixed ordering. Let $\I_n$ and $\mathds{1}_n$ be respectively the \hk{$n \times n$} identity matrix and the all-one vector of size $n$. For a matrix $\M$, we denote $\M_{ij}$ an entry, $\M_{i:}$ its $i$-th row and $\M_{:j}$ its $j$-th column. We also denote $\|\M\|_F$ and  $\|\M\|$ respectively the Frobenius norm and operator norm (i.e., the largest singular value). For a vector $\vec{v}$, we denote $\|\vec{v}\|$ its Euclidean norm. For any \hk{positive} integer $J$, let $[J]$ denote the set $\{1,2,\dots, J\}$.

\subsection{Graph similarity function for network change point detection}\label{sec:sim_NCPD}

We consider a single dynamic network $\mathcal{N}_T = \{G_i\}_{1\leq t \leq T}$ with an unknown number of change-points $1 < \tau_1 < \dots < \tau_K < T$, $K \geq 1$, such that, for any $k \in [K]$ we have
\begin{align*}
    &G_i \overset{i.i.d}{\sim} \mathcal{G}_{k-1}, \quad  \tau_{k-1} \leq i < \tau_{k}, \\
    &G_i \overset{i.i.d}{\sim} \mathcal{G}_k, \quad \hspace{2ex} \tau_k \leq i < \tau_{k+1},
\end{align*}
with $\tau_0 = 1$ and $(\mathcal{G}_{0}, \dots, \mathcal{G}_{K})$ distinct graph generating distributions. We assume that 
$\forall k \geq 1, \tau_k - \tau_{k-1} \hk{\geq} L_0$, with $L_0 > 0$ a known lower bound of the minimal spacing between two consecutive change-points. We recall that in our setting, the set of nodes in each graph snapshot $G_t$ is fixed and its ordering is kept unchanged along the sequence.

Assume for now that we have a graph similarity function $s : \mathds{G} \times \mathds{G} \to [0,1]$ such that for any $G_{t_1} \sim \mathcal{G}_{i_{1}}, G_{t_2}  \sim \mathcal{G}_{{i_2}}$, $s(G_{t_1}, G_{t_2}) > 0.5$ if $ \mathcal{G}_{i_1} = \mathcal{G}_{i_2}$ and  $s(G_{t_1}, G_{t_2}) \leq 0.5$ otherwise. One can then detect change-points in $\mathcal{N}_T $ by monitoring the following average similarity statistic
\begin{align}\label{eq:CP_stat}
    Z_t(s, L) = \frac{1}{L} \sum_{i=1}^L s(G_t, G_{t-i}), \quad t \geq L,
\end{align}
where $L < L_0$ is a hyperparameter that controls the length of the past 
window, and declare a change-point at any timestamp $t$ such that
\begin{align}\label{eq:detection_rule}
     &Z_{t'}(s, L) > 0.5 , \quad t-L \leq t' < t,  \\
     &Z_{t}(s, L) \leq 0.5 \nonumber.
\end{align}
This general method can be applied to recover an arbitrary number of change-points in the dynamic network in an online setting and without any detection delay, i.e., as soon as the data is collected. In practice, one can choose a graph similarity function or kernel \mc{$s(\cdot,\cdot)$} and a detection threshold $\theta$, e.g., using a validation criterion \cite{ranshous2015anomaly} or a significance test procedure using stationary bootstrap \cite{Cribben2017}, and declare a change-point (or an anomaly) in the dynamic network whenever $Z_{t}(s, L) > \theta$. Note that the properties of this method heavily depend on the chosen similarity function and its discriminative power. 

Our NCPD method consists \hk{of} using the statistic $ Z_t(s, L)$ and the detection rule \eqref{eq:detection_rule}, together with a data-driven graph similarity function $s(\cdot,\cdot)$ learnt by a s-GNN model, which we describe in the next section.

\begin{remark}
Our method can also be employed in an offline setting, where one aims at localising changes in a dynamic network after the whole sequence has been collected, with a slight change of the detection rule. For instance, for a dynamic network with a single change-point, one can localise the latter at $\hat{\tau}$, such that
\begin{align}
    &\hat{\tau} = \arg \min_{t \in [L,T]}  Z_t(s, L),  \nonumber \\
    \text{or} \quad &\hat{\tau} = \arg \max_{t \in [L+1,T]}  |Z_t(s, L) - Z_{t-1}(s, L)|. \label{eq:loc_single}
\end{align}
Additionally, our method could be adapted to a setting where  a small detection delay (e.g., of order $L$) may be tolerated. In this case, we could replace \eqref{eq:CP_stat} by a more robust change-point statistic that also uses a future (or forward) window, e.g., $(G_t, G_{t+1},\dots G_{t+L})$. For instance, we could use a two-sample test statistic on the two sets of graphs $(G_{t-1}, \dots, G_{t-L})$ and $(G_{t}, \dots, G_{t+L})$ such as the maximum kernel Fisher discriminant ratio \cite{harchaoui2009kernel} or the maximum mean discrepancy (MMD)  \cite{gretton2008kernel}, for which an unbiased estimate is given by
\begin{align*}
     Z_t^{MMD} = \sqrt{\frac{1}{L(L+1)} \sum_{i, j=1}^{L+1} \left(s(G_{t-i},G_{t-j}) + s(G_{t-1+i},G_{t-1+j}) - s(G_{t-i},G_{t-1+j})\right)}.
\end{align*}
Note that this estimate would correspond to the empirical MMD measure between two sets of graphs mapped into a reproducing kernel Hilbert space if the function $s(\cdot,\cdot)$ was a graph kernel function \cite{gretton2008kernel}.
\end{remark}

\subsection{Graph similarity learning via siamese graph neural networks}\label{sec:sGNN}
Siamese graph neural networks (s-GNN) are architectures designed to compare pairs of graphs,  e.g., for learning a graph similarity function or distance. They can notably be used in graph classification and graph matching tasks \cite{ma2019similarity, ktena2017distance} in both supervised and unsupervised settings. More precisely, a general s-GNN takes as input a pair $(G_1, G_2)$, embeds $G_1$ and $G_2$ with the same graph encoder (or equivalently, two \emph{siamese} encoders that share the same weights), then combines the embeddings in a \hk{symmetric similarity} module. 
The variability of s-GNN architectures mainly lies in the design of these two modules (see for instance \cite{ktena2017distance, ma2019similarity, ling2021}).

\begin{figure}[t]
     \begin{subfigure}[b]{0.5\textwidth}
     \centering
        \includegraphics[width=\textwidth, trim=0.cm 0.cm 0cm  0.cm,clip]{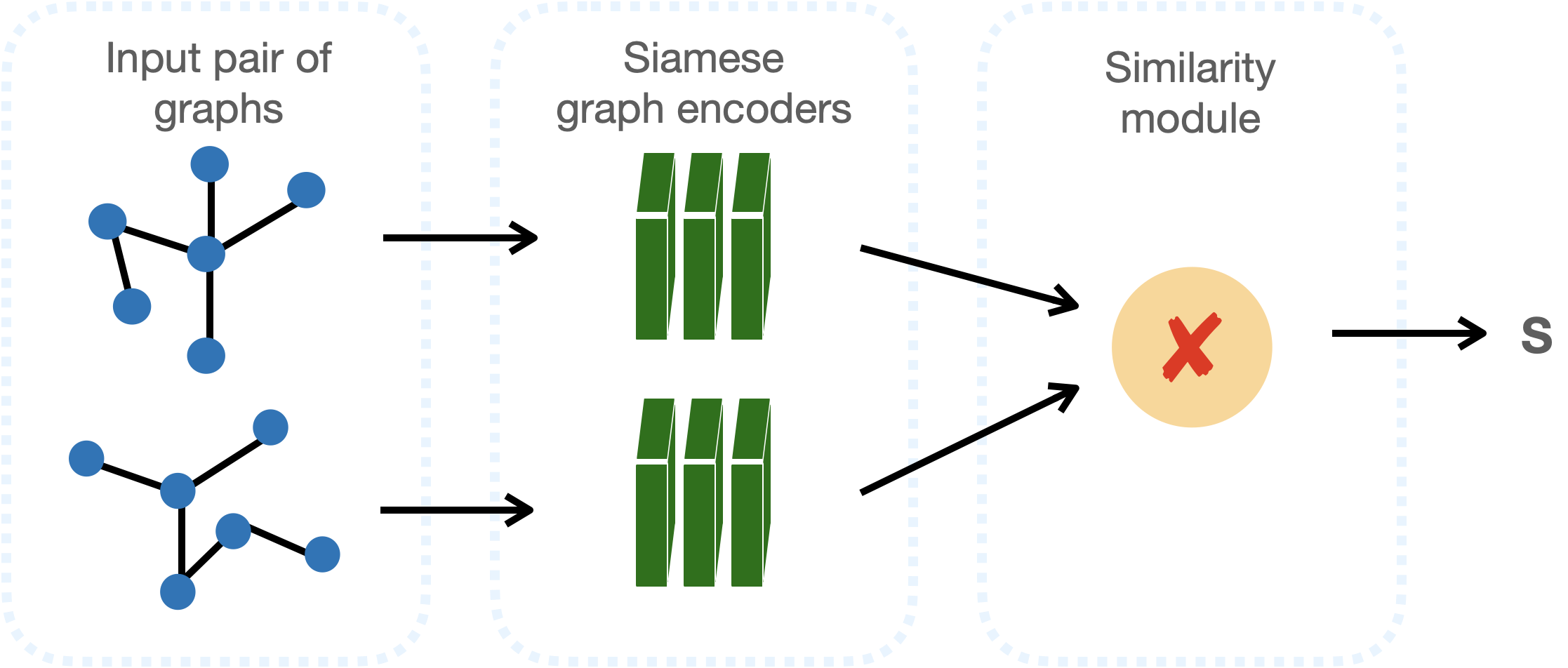}
        \caption{Siamese graph neural network}
         \label{fig:siamese}
    \end{subfigure}
    \hfill\hspace{-0.5cm}
    \begin{subfigure}[b]{0.4\textwidth}
     \centering
        \includegraphics[width=\textwidth, trim=0.cm 0.cm 0.3cmv 0.cm,clip]{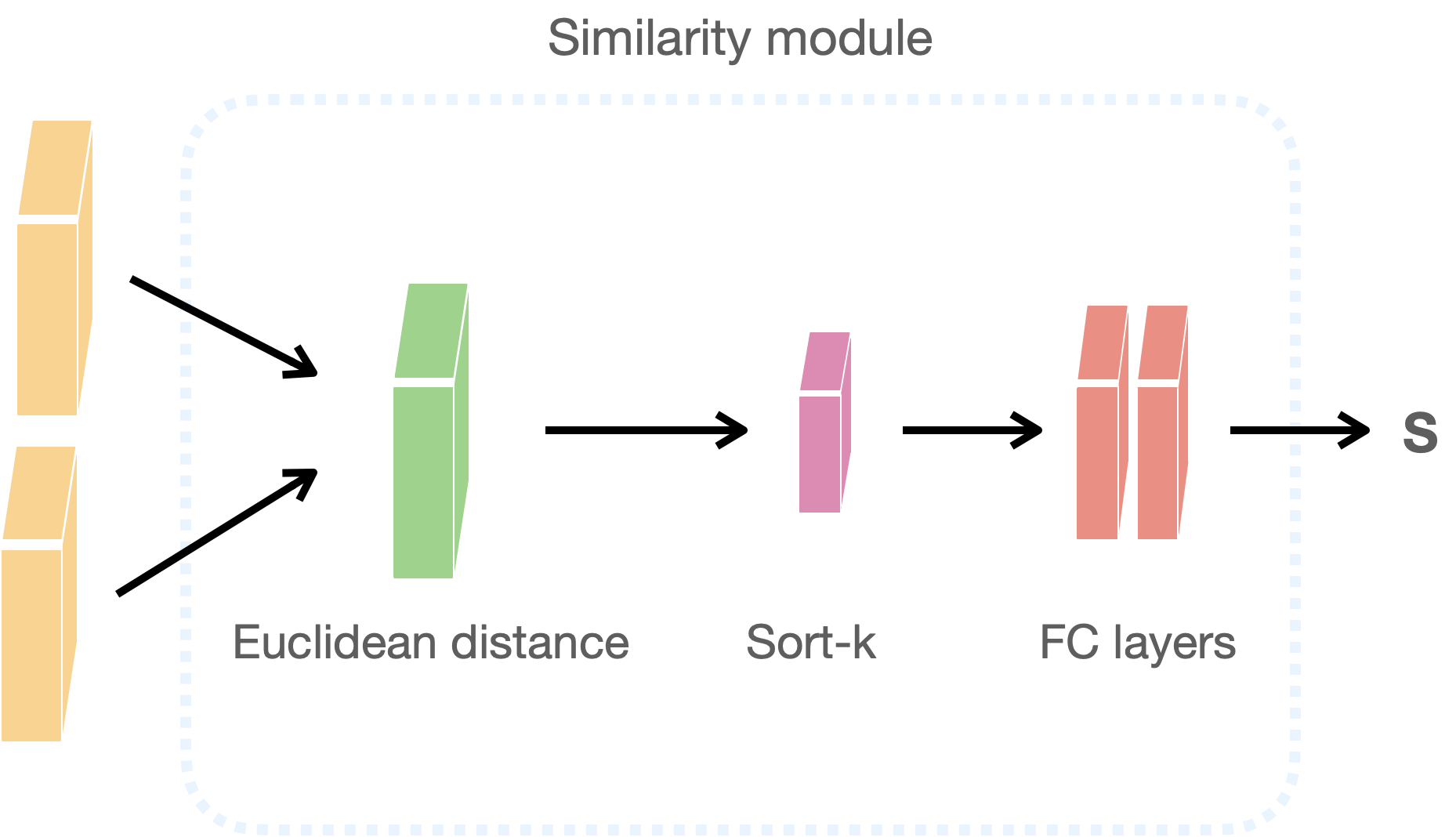}
        \caption{Similarity module}
        \label{fig:similarity_module}
    \end{subfigure}
    \hfill
        \caption{Architecture of our graph similarity learning model. The general pipeline (\subref{fig:siamese}) is a siamese GNN where the output module is a similarity module (\subref{fig:similarity_module}). We design the latter with Euclidean distance, Sort-$k$ pooling operations, and fully-connected layers, for measuring the proximity of snapshots in dynamic networks. } 
        \label{fig:GSN}
\end{figure}

In our NCPD method, we propose a s-GNN architecture summarized in Figure \ref{fig:GSN}, for learning a similarity score $s(G_{t_1},G_{t_2})$ in $[0,1]$ on the space of graph snapshots $(G_1, G_2, \dots, G_t, \dots)$ from the dynamic network. 
For this purpose, we design a similarity module for comparing the node-level embeddings output by a generic graph encoder (e.g., a graph convolutional network \cite{kipf2016semi}, a graph attention network \cite{velickovic2018graph}, a GraphSage network \cite{hamilton2018inductive} or a graph isomorphism network (GIN) \cite{xu2019powerful}). Our similarity module consists of Euclidean distance,  pooling and fully-connected layers, as detailed in Figure \ref{fig:similarity_module}. 
The design of this module allows to select local regions of interest of the graph and limits the number of parameters by using Sort-$k$ pooling \cite{zhang2018end} and two fully-connected layers. 

For the sake of simplicity, we use a simple graph convolutional network (GCN) \cite{kipf2016semi} for undirected and unsigned graphs as the graph encoder in our architecture. However, this block can be replaced by any \textit{ad-hoc} graph encoder. With a GCN, the embedding of a graph $\H^{(j)}$ at each layer $j \in [J], J \geq 1$ is computed as follows
\begin{align}\label{eq:gcn}
    \H^{(j)} = \sigma\left( \Tilde{\A} \H^{(j-1)} \mathbf{W}^{(j)} + \mathbf{B}^{(j)} \right),
\end{align}
where $ \mathbf{W}^{(j)} \in \mathbb{R}^{h_{j-1} \times h_j}$ is a weight matrix, $h_j$ is the number of hidden units of layer $j$, $\mathbf{B}^{(j)} \in \mathbb{R}^{h_{j}}$ is a bias vector, $ \Tilde{\A} = \Tilde{\D}^{-1/2} (\A + \I_n) \Tilde{\D}^{-1/2}$ is the normalized augmented adjacency matrix with degree matrix $\Tilde{\D} = \diag((\A+\I_n) \mathds{1}_n)$, and $\sigma$ is the point-wise ReLU activation  function, i.e., $\sigma(x) = \max(x,0)$. The input of the first layer,  $\H^{(0)}$, is either the node attributes matrix $\E \in \mathbb{R}^{n \times d}$ if the input graph is attributed or a positional encoding matrix (see below). Finally, the output of the GCN is the node-level embedding matrix  $\H^J \in \mathbb{R}^{n \times h_J}$ at the last layer. Therefore, for a pair of graphs $(G_{t_1}, G_{t_2})$, this siamese encoder module computes a pairs of graph embeddings,  $(\H_1, \H_2) := (\H^J(G_{t_1}), \H^J(G_{t_2}))$, and the vectors $(\H_1)_{i:}$ and $(\H_2)_{i:}$ correspond to the representations of the node $i$ respectively in $G_{t_1}$ and $G_{t_2}$. Intuitively, a large distance between these two embeddings can indicate that node $i$ plays distinct structural roles in $G_{t_1}$ and $G_{t_2}$. 

Then, the pair of embeddings $(\H_1, \H_2)$ is processed by a similarity module, which first computes a vector of Euclidean distance between the nodes' embeddings, and secondly, applies a Sort-$k$ pooling operation \cite{zhang2018end} to select its $k$ largest entries, i.e.,
\begin{align*}
    \P = (f_{r_1}, \dots, f_{r_k} ), \quad f_{i} = \|(\H_1)_{i:} - (\H_2)_{i:}\|_2 \in \mathbb{R}_{\geq 0}, \quad 1 \leq i \leq n,
\end{align*}
where $r_1, \dots, r_k$ correspond to the indices of the (sorted) $k$ largest elements of $\{f_{i}\}_{i\in [n]}$. 
\hk{Next, the pooled vector $P$ is processed by two fully connected layers, each of them containing an affine transformation, batch normalisation and ReLU activation function. Finally, the output of the second fully connected layer is pooled using a sum-pooling layer followed by a} sigmoid activation function, so that the final output of the similarity module (and the s-GNN), $s(G_{t_1}, G_{t_2}) \in [0,1]$, a non-negative similarity score between the two input graphs. This score can be transformed into a similarity label via 
\mcc{
\vspace{-3mm}
\[  
\hat{y}(G_{t_1}, G_{t_2}) = 
\left\{
\begin{array}{lll}
1 & \text{(i.e., $G_{t_1}$ and $G_{t_2}$ are similar or have the same generative distribution)}, & \text{if }  s(G_{t_1},G_{t_2}) > 0.5 \\
0 & \text{(i.e., $G_{t_1}$ and $G_{t_2}$ are dissimilar or have different generative distributions)}, &  \text{otherwise}.  \\
\end{array} 
\right. \]
}

Note that using a Sort-$k$ pooling layer in the design of the similarity module has two main advantages in our context.

\bb First, since it follows the (node-wise) Euclidean distance operation, it therefore selects the nodes that have the largest discrepancies between their embeddings in the two graphs. Therefore, if a structural change in the dynamic network affects only a few nodes, 
this change can be picked up by this pooling operation, without being diminished by the absence of change in the rest of the network. 
\mcc{This component could be further built upon for identifying which \textit{local} part of the network is mainly driving the change-point, thus enhancing the explainability of the proposed  pipeline.}

\bb  Second, Sort-$k$ pooling reduces the number of parameters while preserving the most important information for measuring potential and local graph changes. More generally, replacing max or sum pooling by sorted pooling have been proven to increase the accuracy and generalization power of neural networks, in particular in settings with limited data availability, such as one-shot learning  \cite{horvath2020sorted}, and can also be used for downsampling graphs \cite{lee2019pooling}. In the network change-point-agnostic detection task, we incorporate this pooling layer to mitigate our lack of information on the change-points. 

\begin{remark}\label{rem:invariance}
 It is often a desirable property of GNN models with graph-level (resp. nodel-level)  output to be invariant (resp. equivariant) to nodes' permutations. This is due to the fact that nodes in a graph are generally considered exchangeable, or in other words, the order of node set in the adjacency and node attributes matrix is arbitrary. In our method, the s-GNN model takes as input pairs of graph snapshots from a dynamic network sequence (see Section \ref{sec:sim_NCPD}), where every snapshot contains the same set of nodes with the same ordering. Therefore, in our context, the invariance property of the learned graph similarity function denotes that the latter is invariant to any permutation of the nodes that is applied on both inputs. More precisely, for any permutation of the node set $\sigma: [n] \to [n]$, denoting $\sigma * G$ the resulting transformation of a graph $G$ under $\sigma$ (i.e., permutation of the rows and columns of the adjacency and node attributes matrices), the invariance property writes $s(\sigma(G_1), \sigma(G_2)) = s(G_1, G_2)$. This is indeed the case for our method since the node-wise operations, i.e, the graph encoder and the Euclidean distance, are equivariant.  Then, since the Sort-$k$ pooling layer is permutation-invariant, i.e. $\P(\H) = \P(\sigma(\H))$, so is the final similarity score.
\end{remark}

Moreover, when the dynamic network is unattributed, i.e., each graph snapshot contains only structural information $G_i = (\A_i, \emptyset)$, one needs to choose an appropriate initialisation of the node features matrix $H^{(0)}$ as input of the s-GNN.  
Following existing methodology \cite{you2019positionaware}, we compute Positional Encodings (PE), i.e., synthetic node attributes that capture their relative positions in the graph structure. In fact, it has been previously noted that the choice of PE is critical for the expressivity of GNN models \cite{dwivedi2022graph}. Therefore, in our experiments, we will use and compare four types of encoding, the first three being existing techniques that have been introduced in different graph learning settings, and the last being one that we believe may also be appropriate for certain NCPD tasks.
\begin{enumerate}
    \item \textbf{Degree encoding \cite{bruna2017community}:} the attribute of a node is a scalar equal to its degree in the graph, i.e. $H^{(0)} = \hk{\A \mathds{1}_n} \in \mathbb{R}^{n \times 1}$.
    \item \textbf{Random-Walk encoding \cite{li2020distance, dwivedi2022graph}:} for $k \geq 1$, the vector of attributes of a node $n_i, 1\leq i \leq n$ is defined as
    \begin{align*}
        \H^{(0)}_{i:} = [\Rmatrix_{ii},\Rmatrix^2_{ii}, \dots \Rmatrix^k_{ii}] \in \mathbb{R}^k, 
    \end{align*}
    where $\Rmatrix = \A \D^{-1}$ is the random-walk operator and $k\geq 1$ is a hyperparameter.
    \item \textbf{Laplacian encoding \cite{dwivedi2021generalization}:} the node attributes are the principal eigenvectors of the symmetric normalised Laplacian matrix $\L = \I_n - \D^{-1/2} \A \D^{-1/2}$. We note that this is similar to the first steps of spectral clustering algorithms. More precisely, using the factorisation $\L = U^T \Lambda U$ where $U, \Lambda$ respectively contain the ordered set of eigenvectors and eigenvalues of $L$, the Laplacian encodings are defined as
    \begin{align*}
        \H^{(0)} = [U^T_{:1}, U^T_{:2}, \dots, U^T_{:k} ] \in \mathbb{R}^{n \times k}, 
    \end{align*}
    where $k\geq 1$ is a hyperparameter. 
    
    \item \textbf{Identity encoding: } we define the initial feature matrix as $\H^{(0)} = \I_n$, which corresponds to learning a unique initial input embedding for each node at the first layer of the s-GNN. We argue that this is an appropriate choice for the graph siamese encoder in our setting. In fact, these encodings in general break the equivariance property of GNN models; however, this is not the case here since this property has a modified definition when the graphs are snapshots of a dynamic network (see Remark \ref{rem:invariance}). We recall that we have assumed that the set of nodes is constant in the dynamic network,   
    and the global ordering of the nodes, although arbitrary, is common to all graph snapshots.  Finally, it has been previously noted in graph learning tasks that taking into account the nodes' identities \cite{donnat2018tracking} can be beneficial. We will in particular use these encodings for the real-world dynamic network with a small number of nodes in Section \ref{sec:exp_physical}.
\end{enumerate}
We note that more complex strategies for computing positional encodings include learning them along the training procedure of the s-GNN \cite{dwivedi2022graph}. However, we do not consider these latter approaches, which significantly increase the model complexity.

Finally, our s-GNN architecture can classically be trained in a supervised way if a data set of labelled pairs of graphs is available, with $\mathcal{D} = \{(G_1^i, G_2^i, y_i)\}_i$ with $y_i \in \{0,1\}$ indicating if the graphs come from the same distribution or not. For example, one can optimise its parameters by minimising the cross-entropy loss function
\begin{align*}
    L(G_1, G_2, y) =  - y \log s(G_1, G_2) - (1-y) \log (1- s(G_1, G_2)),
\end{align*}
\hk{via gradient descent.} In the NCPD task, designing such a data set requires an adequate sampling scheme of the snapshots in the dynamic network as well as an \textit{ad-hoc} validation procedure. In the next section, we present our supervised learning strategy for the s-GNN.

\subsection{Training and validation procedures for NCPD}\label{sec:training}

In the NCPD task, a supervised setting corresponds to the case where a training subsequence of the dynamic network containing ground-truth change-points is available. In this setting, these change-point labels can then be used to design training and validation sets for our GSL model, 
with these sets containing  triplets $(G_i^1, G_i^2, y_i)$, where $y_i \in \{0,1\}$ is a similarity label ($y_i = 1$ corresponding to "similar"). In this section, we describe our strategy for sampling such triplets from the dynamic network sequence, for both the training and validation steps. 

First, we divide the sequence of graph snapshots into training, validation and test subsequences, e.g., using consecutive windows of respectively x\%, y\% and z\% timestamps. In the training and validation sequences, we sample labelled pairs of graphs according to the two following schemes.
\begin{enumerate}
    \item \textbf{Random scheme (training set):} we consider the set of all (non-ordered) pairs of graphs in the training sequence and label each pair $(G_{t_1}, G_{t_2})$ with $y = 1$ if  there is no change-point between $t_1$ and $t_2$, and 
    $y_i = 0$ otherwise. Then we uniformly sample a fixed number of pairs with label 1 ("positive" examples) and the same number of pairs with label 0 ("negative" examples), without replacement. The number of pairs is a hyperparameter of the method, which can be chosen heuristically as $10 \times T$,  if $T$ is not too large.
    \item \textbf{Windowed scheme (validation set):} we consider the set of all (non-ordered) pairs of graphs in the network sequence that are not distant from each other by more than $L$ timestamps, and label them with the same procedure as in the \textbf{Random scheme}. 
\end{enumerate}

We note that the different sampling mechanisms for the pairs in the training and validation sets are designed to satisfy a double objective of our learning procedure: we aim to learn an adequate graph similarity function and to detect change-points in a network sequence using the latter. For the first objective, the \textbf{Random scheme} allows to create pairs of graph snapshots that are far away in the sequence in order to avoid strong correlations between the training pairs and undesired "transition" effects in real networks, e.g., when the change of generative distribution is not abrupt. Moreover, with the scheme, we can avoid label imbalance in the training set by sampling the same number of positive and negative pairs, which we assume is favorable for the s-GNN. For the second objective, the \textbf{Windowed scheme} builds a validation set of pairs that imitates the test setting of our GSL model. In fact, in our online NCPD method (see Section \ref{sec:sim_NCPD}), the evaluation of the graph similarity function $s(.,.)$ in the statistic \eqref{eq:CP_stat} (and detection rule \eqref{eq:detection_rule}) only applies for pairs of graphs within a sliding window of size $L$. In particular, in the test setting, the pairs of graphs that are compared by $s(\cdot,\cdot)$ are highly correlated and the number of positive pairs is  much larger than the number of negative examples, since there are generally only few change-points in the dynamic network. We finally note that in both the \textbf{Random} and \textbf{Windowed schemes}, the sampled pairs may have in common (at most) one graph snapshot.

Consequently, in a supervised NCPD setting, we can train our s-GNN model in a supervised way as a binary classifier of pairs (see Section \ref{sec:sGNN}) using the previous sampling strategies. In an unsupervised NCPD setting, i.e., when the dynamic network does not contain any ground-truth label of change-point, we need to resort to a novel \emph{self-supervised learning} technique \cite{liu2021graph}.  
In this case, we first pre-estimate a set of change-points in the training and validation sequences and then use the previous sampling schemes to draw training and validation pairs of graphs. This learning procedure is applied on the financial network data set in Section \ref{sec:exp_financial}, where our algorithm for pre-estimating a set of change-points is based on a spectral clustering technique.

\section{Numerical experiments}\label{sec:experiments}

In this section, we test and evaluate the performances of our s-GNN method in the online NCPD task, first in a controlled setting of synthetic dynamic networks (Section \ref{sec:exp_synthetic}), then on real-world correlation networks (Sections \ref{sec:exp_financial} and \ref{sec:exp_physical}).  
Moreover, since one of these data sets does not contain ground-truth change-points, we also introduce a self-supervised learning procedure for our method (see Section \ref{sec:exp_financial}). 

\subsection{Performance metrics}\label{sec:metrics}

For dynamic network data sets with ground-truth labels of change-points, we evaluate the performance of NCPD methods using the following metrics, in the single or multiple change-point settings:
\begin{itemize}
    \item \textbf{Localisation error (single change-point)} defined as $\text{Error}_{\text{CPD}} = |\hat{\tau} - \tau|$, where $\tau, \hat{\tau}$ are respectively the ground-truth and estimated change-points.
    \item \textbf{Adjusted F1-score \cite{xu18} (multiple change-points)} on the classification of timestamps as change-point (label 1) or not change-point (label 0). We tolerate an error of $\pm 5$ timestamps on this task, i.e. all the timestamps within a window of length 11 centered at the ground-truth change-point are given a ground-truth label 1, and a valid detection occurs whenever one of these timestamps is classified as change-point.
\end{itemize} 

For the the data set without ground-truth labels in Section \ref{sec:exp_financial}, we qualitatively discuss our findings, \mcc{place them in a financial context,} and compare them with previous analysis of similar data.  Additionally, in the synthetic data experiments in Section \ref{sec:exp_synthetic}, we also evaluate the ability of our graph similarity function $\hat{s}$ to discriminate between graphs sampled from the same or different distributions, i.e., to classify pairs of graphs generated from either the same or different random graph models. We measure this property in terms of the accuracy score.

\subsection{Baselines}\label{sec:baselines}

We will compare our data-driven graph similarity function to graph distances, similarity function and kernel previously used in the context of NCPD and graph two-sample-test.   
\begin{itemize}
    \item \textbf{Frobenius distance} \cite{Barnett2016, Nie2021WeightedGraphBasedCP, Bao2018NetworkDB, dubey2021online}, defined as $d_F(\A,\B) = \|\A - \B\|_F$, for two matrices $\A,\B$ with \hk{equal} dimensions. Here, we will apply this distance to the adjacency matrices of two graphs. Note that one can also apply it on the graph Laplacian matrices \cite{Bao2018NetworkDB}, and that this distance has also been used in a minimax testing perspective between two graph samples \cite{Ghoshdastidar2020}. 
    
     \item \textbf{Procrustes distance} between Laplacian principal eigenspaces  \cite{hewapathirana2019change}. This distance corresponds to the Frobenius distance between the matrices of eigenvectors corresponding to the $k$ largest eigenvalues of the symmetric graph Laplacian $\L = \I_n - \D^{1/2}\A \D^{1/2}$, after performing an alignment step. The number of eigenvectors $k$ can be pre-specified or chosen by finding the optimal low-rank approximation of $\L$. 
     
     \item \textbf{DeltaCon similarity \cite{Koutra2016DeltaConPM}}. This graph similarity function is based on the Matusita distance applied to the Fast Belief Propagation graph operators, defined for a graph as $S = [\I_n + \epsilon^2 \D - \epsilon \A]^{-1}$ with $\epsilon > 0$. We use the implementation of this similarity function provided in the python package \textsc{netrd}\footnote{\url{https://netrd.readthedocs.io/en/latest/index.html\#}}.

    \item \textbf{Weisfeiler-Lehman (WL) kernel \cite{wlkernel2011}}. 
    This graph kernel is notably used in the two-sample-test problem for sets of graphs \cite{gretton2008kernel}.
    We use the implementation from the GraKel python package \cite{grakel}, and fix the number of iterations of the WL kernel algorithm to 5 in our experiments.
    
\end{itemize}
We will use the previous baselines in the statistic \eqref{eq:CP_stat} and detect change-points using a threshold chosen on a validation set. We also compare our NCPD pipeline to methods that do not rely on an explicit graph metric for detecting change-points.
\begin{itemize}
    
    \item \textbf{Network change-point detection with spectral clustering (SC-NCPD)} \cite{Cribben2017}. This method first partitions the node set of each snapshot with a spectral clustering algorithm and compute an inner product between averages of spectral features across a backward (or \emph{past}) and a forward (or \emph{future}) windows. In this method, the number of clusters and the lengths of the windows are pre-specified. 
    
    \item \textbf{Laplacian anomaly detection (LAD)} \cite{Huang2020}. This method applies both to the anomaly detection and change-point detection tasks for dynamic networks, and is based on the anomaly score
        \begin{align*}
        Z_t = 1 - |\Tilde{\sigma}_t \sigma_t|,
    \end{align*}
    where $\sigma_t$ is the vector of top-$k$ singular values of the unormalized Laplacian of the graph $G_t$ and $\Tilde{\sigma}_t$ aggregates (e.g. averages) the top-k singular values of each snapshots in a past window of size $L$, i.e., $(\sigma_{t-L}, \dots, \sigma_{t-1})$. 
     The number of singular values $k$ and the length of the window are pre-specified hyperparameters. 
     
    \item \textbf{Network cumulative sums statistic (CUSUM) \cite{yu2021optimal}.} This method uses a backward and a forward windows of sizes $L'$ to compute a sequence of CUSUM matrices
    \begin{align}\label{eq:cusum}
        C_t = \frac{1}{\sqrt{2 L'}} \left(\sum_{s=t-L'+1}^t \A_s - \sum_{s=t+1}^{t+L'} \A_s \right), \quad L' \leq t \leq T - L'.
    \end{align}
    Following the methodology in \cite{yu2021optimal}, we divide the dynamic network into two samples, $N_A = \{G_{2t}\}_{1 \leq t \leq T/2}$ and $N_B = \{G_{2t-1}\}_{1 \leq t \leq T/2}$, containing the snapshots respectively at even and uneven timestamps. This algorithm monitors two statistics based on the CUSUM matrices \eqref{eq:cusum} of these samples: the Frobenius norm of the Universal Singular Value Threshold (USVT) estimator $\widetilde{B}(t)$ of the CUSUM matrix computed from $N_B$, and the dot product between $\widetilde{B}(t) / \|\widetilde{B}(t)\| $ and the CUSUM matrix computed from $N_A$. To avoid tuning the additional threshold parameters, we do not apply the USVT step (or equivalently choose $\tau_1 = 0$ and $\tau_2 = 1$ in USVT). Moreover, we only use the second statistics since the first one is very close to the next baseline.
    
    \item \textbf{Operator norm of network CUSUM (CUSUM 2) \cite{enikeeva2021changepoint}.} We adapt this offline method to the online problem by computing the CUSUM matrix over a past and future windows of size $L'$.
    The NCPD statistics is then $z_t = \|Z_t\|$.
\end{itemize}

For these baselines, we fix the number of clusters or singular values to $k = 6$ and the size of windows to $L'=L/2$ when both the past and future are used in the NCPD statistic. We also note that these methods are applied to non-attributed dynamic networks and therefore only use the sequence of adjacency matrices $(\A_t)_t$. However, only one of our network data sets is attributed (see Section \ref{sec:exp_financial}) and in this case, the node attributes are ignored by the baseline methods.

\subsection{Synthetic data}\label{sec:exp_synthetic}

In this section, we generate dynamic networks from a dynamic stochastic block model \cite{zhao2019changepoint, yu2021optimal, Padilla2019, bhattacharjee2020change} with a unique change-point. More precisely, we generate sequences of unattributed graphs $(G_1, \dots, G_T)$ with $T = 100$ such that for each $t \in [T]$, each graph is independently drawn from a Stochastic Block Model (SBM) with $n = 400$ nodes and
\begin{align*}
    &G_t \overset{i.i.d}{\sim}  \mathcal{G}_1,\quad \text{ if } t < \tau, \\
    &G_t \overset{i.i.d}{\sim}  \mathcal{G}_2,\quad  \text{ if } t \geq \tau,
\end{align*}
where $\mathcal{G}_1, \mathcal{G}_2$ are two SBM distributions. We recall that an SBM with $K \geq 1$ communities can be defined by a connectivity matrix $C = (p - q) I_K + q \mathds{1}_K \mathds{1}_K^T$ with intra- and inter-cluster connectivity parameters $p,q\in [0,1]$, and a membership matrix $\Theta \in \{0,1\}^{n \times K}$. The parameter $p$ (respectively $q$) corresponds to the probability of existence of an edge between two nodes in the same community (respectively in two different communities), while each row $\Theta_i$ of the membership matrix indicates the community a node $n_i$ belongs to.

\begin{figure}
        \begin{subfigure}[b]{0.3\textwidth}
         \centering
          \includegraphics[width=\textwidth, trim={0cm 0cm 3cm 0cm}, clip]{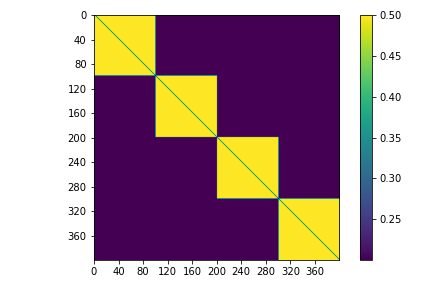}
     \end{subfigure}
            \begin{subfigure}[b]{0.3\textwidth}
         \centering
          \includegraphics[width=\textwidth, trim={0cm 0cm 3cm 0cm}, clip]{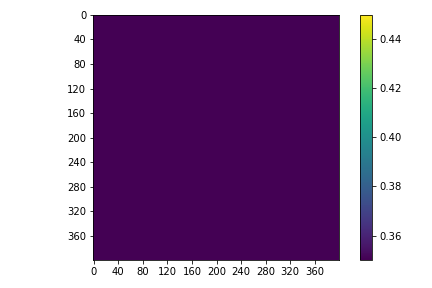}
     \end{subfigure}
        \begin{subfigure}[b]{0.37\textwidth}
         \centering
          \includegraphics[width=\textwidth, trim={0cm 0cm 0cm 0cm}, clip]{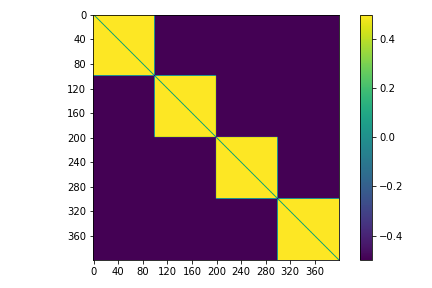}
     \end{subfigure}
        \begin{subfigure}[b]{0.3\textwidth}
         \centering
          \includegraphics[width=\textwidth, trim={0cm 0cm 3cm 0cm}, clip]{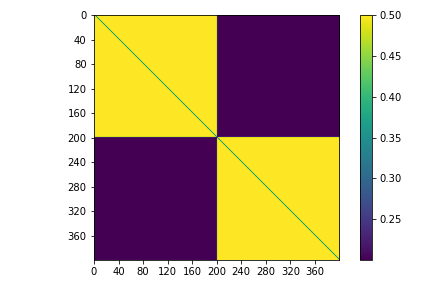}
          \caption{``Merge" scenario.}
         \label{fig:b1}
     \end{subfigure}
            \begin{subfigure}[b]{0.3\textwidth}
         \centering
          \includegraphics[width=\textwidth, trim={0cm 0cm 3cm 0cm}, clip]{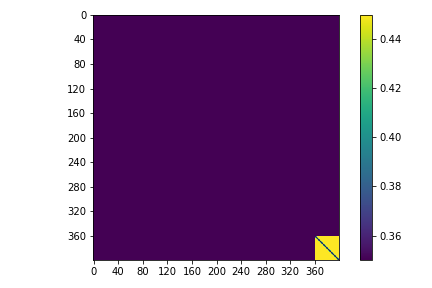}
          \caption{``Birth" scenarios.}
         \label{fig:b2}
     \end{subfigure}
    \begin{subfigure}[b]{0.37\textwidth}
         \centering
          \includegraphics[width=\textwidth, trim={0cm 0cm 0cm 0cm}, clip]{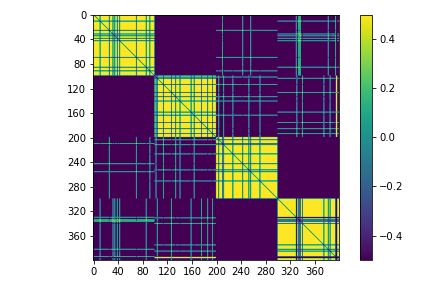}
          \caption{``Swap" scenario.}
         \label{fig:b3}
     \end{subfigure}
     \caption{Expectation of the adjacency matrices of the graphs in the SBMs $\mathcal{G}_1$ (first row) and $\mathcal{G}_2$ (second row) , i.e., before and after the change-point, in our three types of synthetic scenarios, ``Merge" (\subref{fig:b1}), ``Birth" (\subref{fig:b2}) and ``Swaps" (\subref{fig:b3}).}
    \label{fig:adjacencies}
\end{figure}

We consider four different change-point scenarios related to three possible types of events in a community life-cycle \cite{Rossetti_2018}, namely \textbf{``Merge"}, \textbf{``Birth"} and \textbf{``Swaps"}. These community events are illustrated in Figure \ref{fig:adjacencies} by heatmaps of the expected adjacency matrices in $\mathcal{G}_1$ and $\mathcal{G}_2$. We note that in all the following settings, the graph snapshots will be relatively sparse.

\begin{itemize}
    \item \textbf{Scenario 1 (``Merge").} 
    the two SBMs $\mathcal{G}_1$ and $\mathcal{G}_2$ have respectively four and two equal-size clusters with inter-cluster connectivity parameter $q = 0.02$  and  intra-cluster connectivity parameter $p > q$ which we vary. We design several difficulty levels of this scenario by changing the value of $p$: the larger $p$ is, the easier the detection problem is.  
    
    \item \textbf{Scenario 2 (``Birth 1").} 
    This scenario mimics the appearance of a community in a dynamic network. In this case, $\mathcal{G}_1$ is the distribution of an Erdos-Renyi model with parameter $q = 0.03$ and $\mathcal{G}_2$ is a SBM with two communities of size $n-s$ and $s$, $1\leq s \leq n/2$, and connectivity matrix 
    \begin{align*}
         C =  \begin{pmatrix}
         q & q \\
         q & p
        \end{pmatrix},
    \end{align*}
    with $p = 0.1$. 
    We vary the difficulty of this detection scenario by changing the size of the second cluster $s$:  
    the bigger $s$ is, the easier the detection problem is.
    
    \item \textbf{Scenario 3 (``Birth 2").} This scenario uses the setting of Scenario 2 but in this case, the size of the dense subgraph is fixed to $s = 100$ and the difficulty level is controlled by the connectivity $p$.
    \hk{We consider $p > q$ such that the larger} $p$ is, the easier the detection problem.
    
    \item \textbf{Scenario 4 (``Swaps").} 
    In this scenario, the connectivity parameters of the two SBMs are equal but their membership matrices differ. We simulate a recombination of communities where pairs of nodes exchange their community memberships, i.e., two nodes "swap" their community of attachment. The two SBMs have four equal-size clusters with inter-connectivity parameter $q = 0.05$ and  intra-connectivity parameter $p = 0.1$.  We test different difficulty levels by varying the proportion $h$ of pairs of nodes swapping their memberships; 
    the bigger the $h$, the easier the detection problem. 
\end{itemize}
Note that Scenario 1 can be considered as a \emph{global} change of the network structure, while the other scenarios correspond to a \emph{local} topological change (i.e., localised on a subset of nodes). For each scenario and each difficulty level, we generate 50 sequences with one change-point uniformly sampled in the interval $[25, 75]$. Moreover, for the pair classification task (see Section \ref{sec:metrics}), in each scenario, we also independently generate 1000 labelled pairs of graphs $\{(G_1^i, G_2^i, y_i)\}_i$, where for each $i$, $G_1^i \sim \mathcal{G}_k,  G_2^i \sim  \mathcal{G}_l$ with $k,l \in \{1,2\}$ and $y_i = 1$ if $\mathcal{G}_k = \mathcal{G}_l$ and $y_i=0$ otherwise. Each of these data sets of pairs is balanced 
and we use respectively 60\%, 20\% and 20\% of the pairs for training, validation and test. 
In the NCPD task, we estimate the unique change-point with the detection rule \eqref{eq:loc_single}, and use a window size $L = 6$. Additional details on the experimental setting can be found in Appendix \ref{app:details_exp}.

\begin{figure}
    \centering
        \begin{subfigure}[b]{0.7\textwidth}
         \centering
          \includegraphics[width=\textwidth]{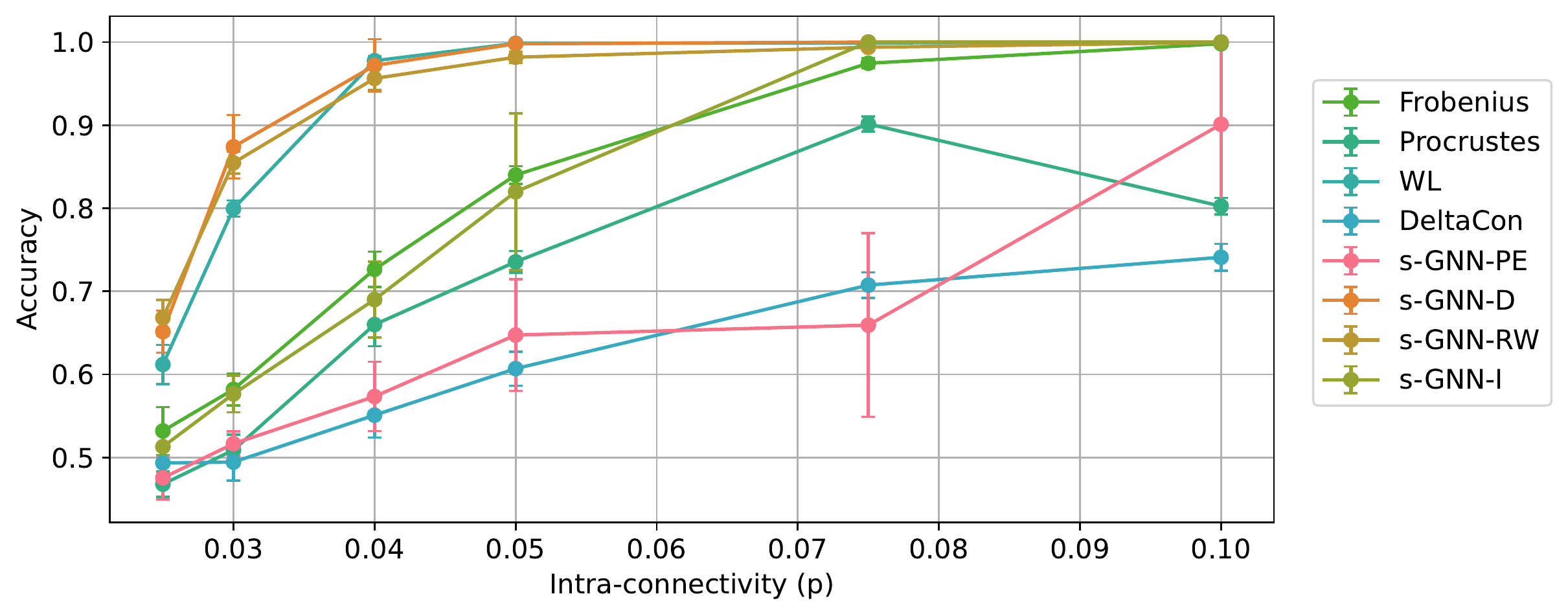}
           \vspace{-5mm}
           \caption{Classification accuracy vs the intra-connectivity parameter $p$.}
         \label{fig:sc1}
     \end{subfigure} 
          \begin{subfigure}[b]{0.7\textwidth}
         \centering
         \includegraphics[width=\textwidth]{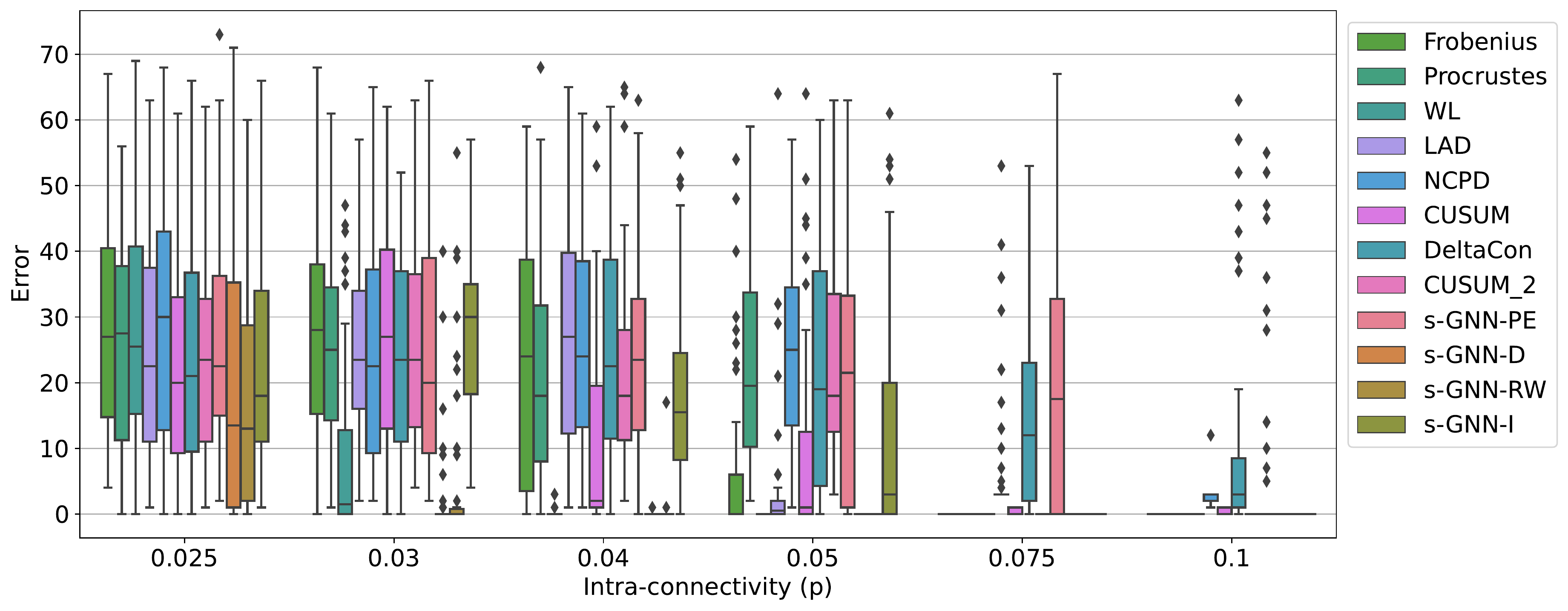}
          \vspace{-5mm}
          \caption{Change-point localisation error for different intra-connectivity parameters $p$.}
         \label{fig:sc1bis}
     \end{subfigure}
     \caption{Performances of our s-GNN method and baselines on the classification  \label{fig:sc2} and detection  \label{fig:sc2bis} tasks in the ``Merge" scenario. The first task consists in classifying pairs of graphs sampled from the same or different SBM distributions using a graph similarity function or a graph distance. The second task consists in localising a single change-point in a dynamic SBM sequence. \mc{We remark that for very large values of $p$, many methods attain zero error.}} 
    \label{fig:scenario1}
\end{figure}

\begin{figure}
    \centering
    \begin{subfigure}[b]{0.7\textwidth}
         \centering
         \includegraphics[width=\textwidth, trim=0cm 0cm 0cm 0cm , clip]{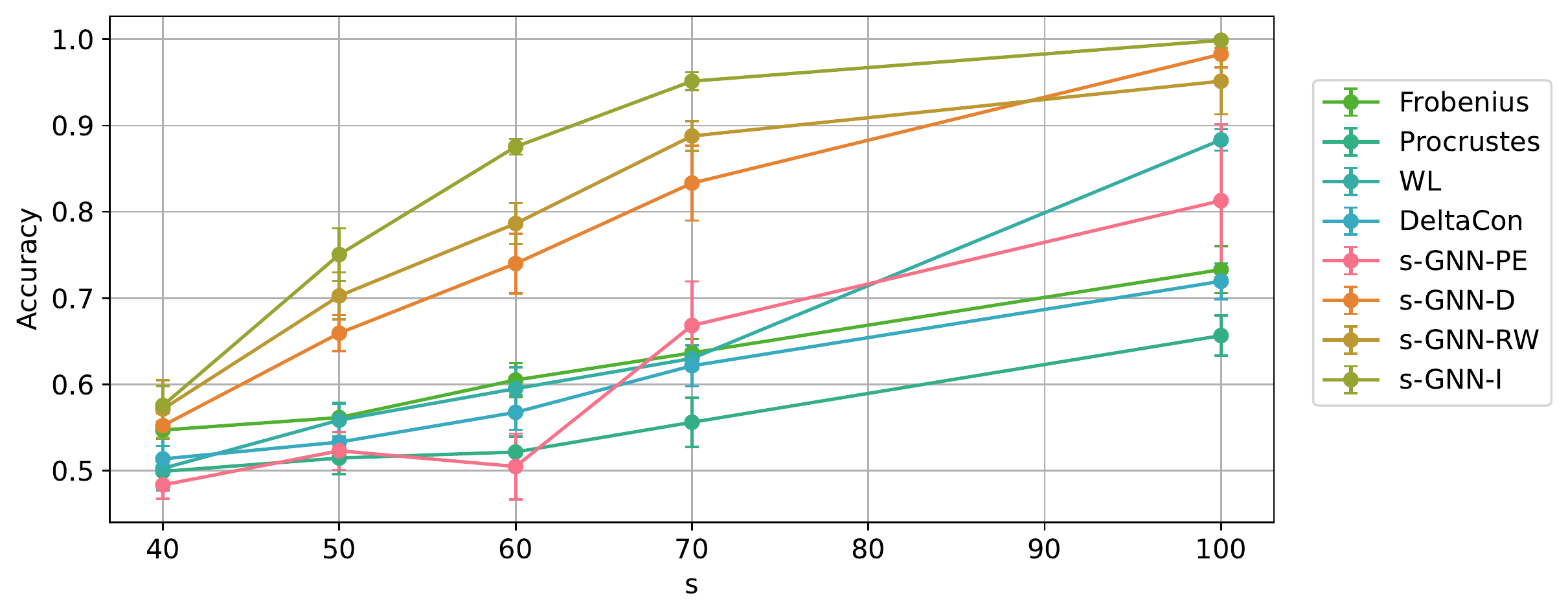}
         \vspace{-5mm}
         \caption{Classification accuracy of pairs vs the community size $s$.}
          \label{fig:sc2}
     \end{subfigure} 
     \begin{subfigure}[b]{0.7\textwidth}
         \centering
         \includegraphics[width=\textwidth]{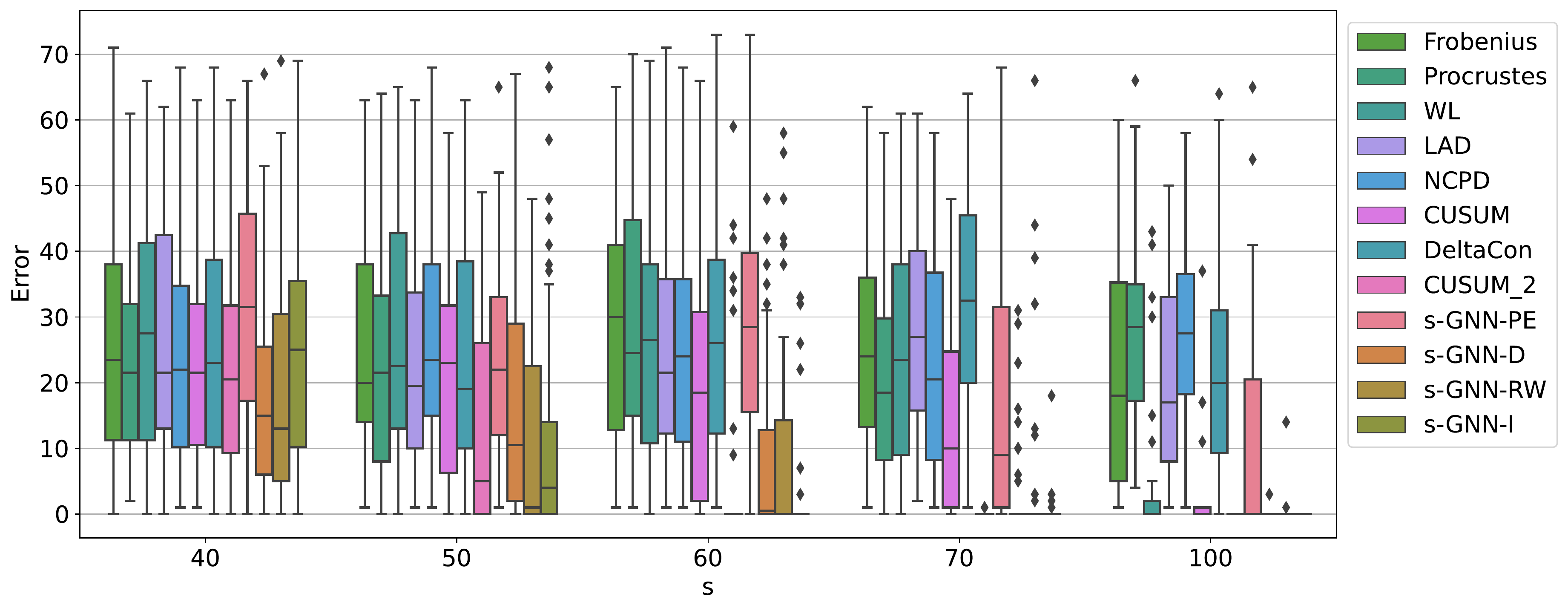}
         \vspace{-5mm}
         \caption{Change point localisation error for different community sizes.}
          \label{fig:sc2bis}
     \end{subfigure}
     \caption{Performances on the classification   (\subref{fig:sc2}) and detection  (\subref{fig:sc2bis})  tasks in the ``Birth 1" scenario.}  
    \label{fig:scenario2}
\end{figure}

\begin{figure}
    \centering
    \begin{subfigure}[b]{0.8\textwidth}
         \centering
         \includegraphics[width=0.9\textwidth]{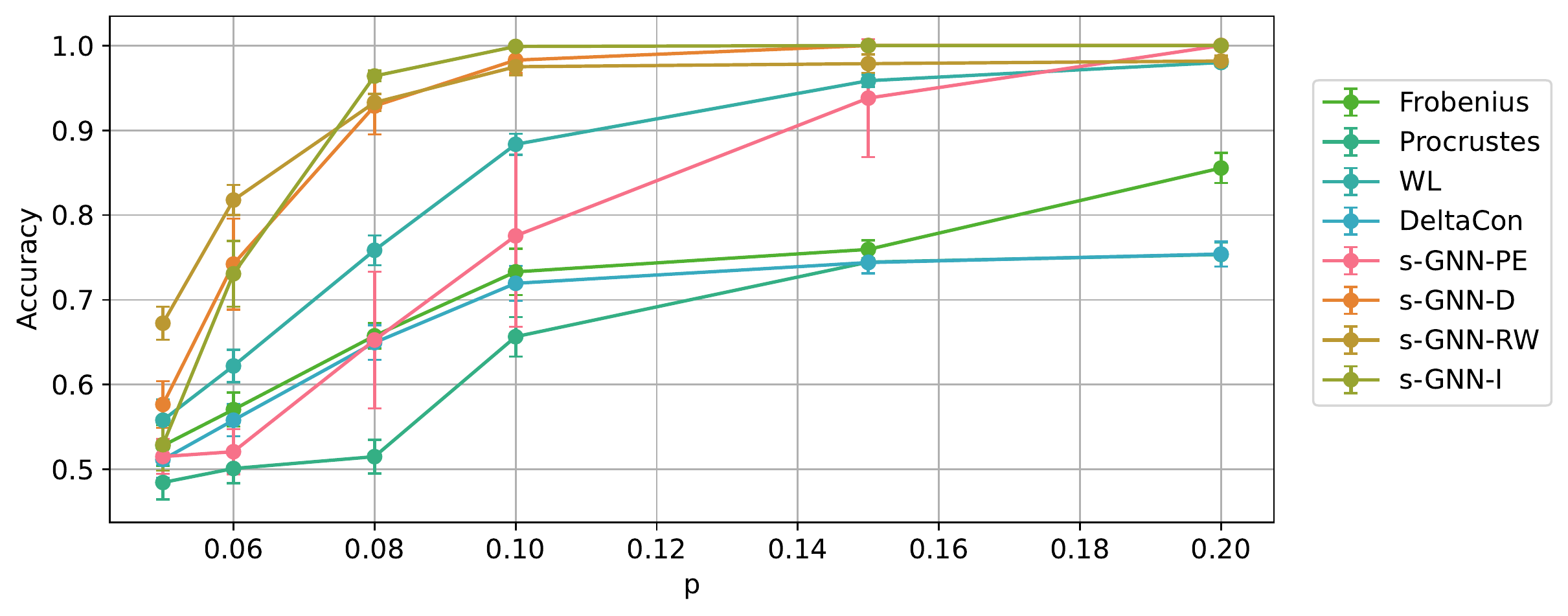}
         \vspace{-3mm}
         \caption{Classification accuracy vs the intra-connectivity parameter $p$.}
          \label{fig:sc2bis1}
     \end{subfigure}
        \begin{subfigure}[b]{\textwidth}
         \centering
         \includegraphics[width=0.75\textwidth]{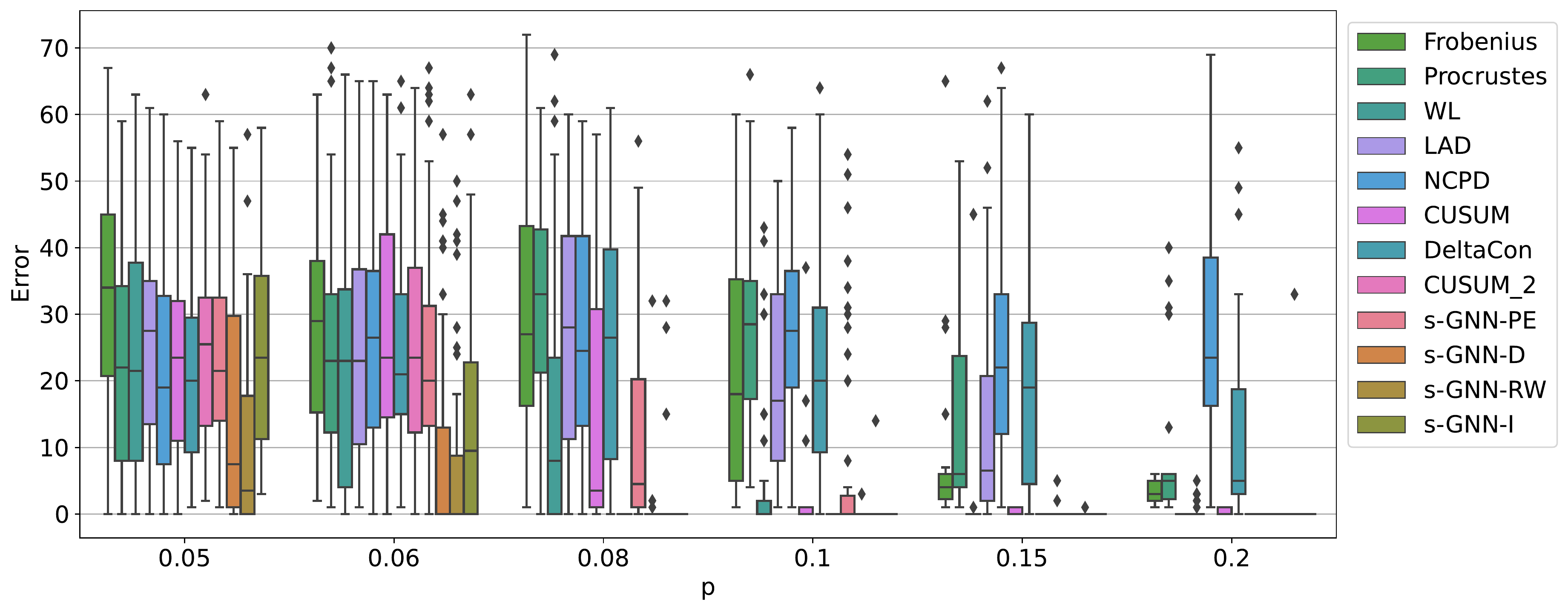}
          \vspace{-3mm}
         \caption{Change point localisation error for different intra-connectivity parameters $p$.}
          \label{fig:sc2bis2}
     \end{subfigure}
      \vspace{-3mm}
     \caption{Performances on the classification (\subref{fig:sc2bis1}) and detection  (\subref{fig:sc2bis2})  tasks in the  ``Birth 2" scenario. The first task consists in classifying pairs of graphs sampled from the same or different random graph models using a graph similarity function or a graph distance. The second task consists in localising a single change-point in a dynamic network sequence.  \mc{We remark that for very large values of $p$, many methods attain zero error.}}
    \label{fig:scenario2bis}
\end{figure}

\begin{figure}
    \centering
    \begin{subfigure}[b]{0.8\textwidth}
         \centering
         \includegraphics[width=0.9\textwidth]{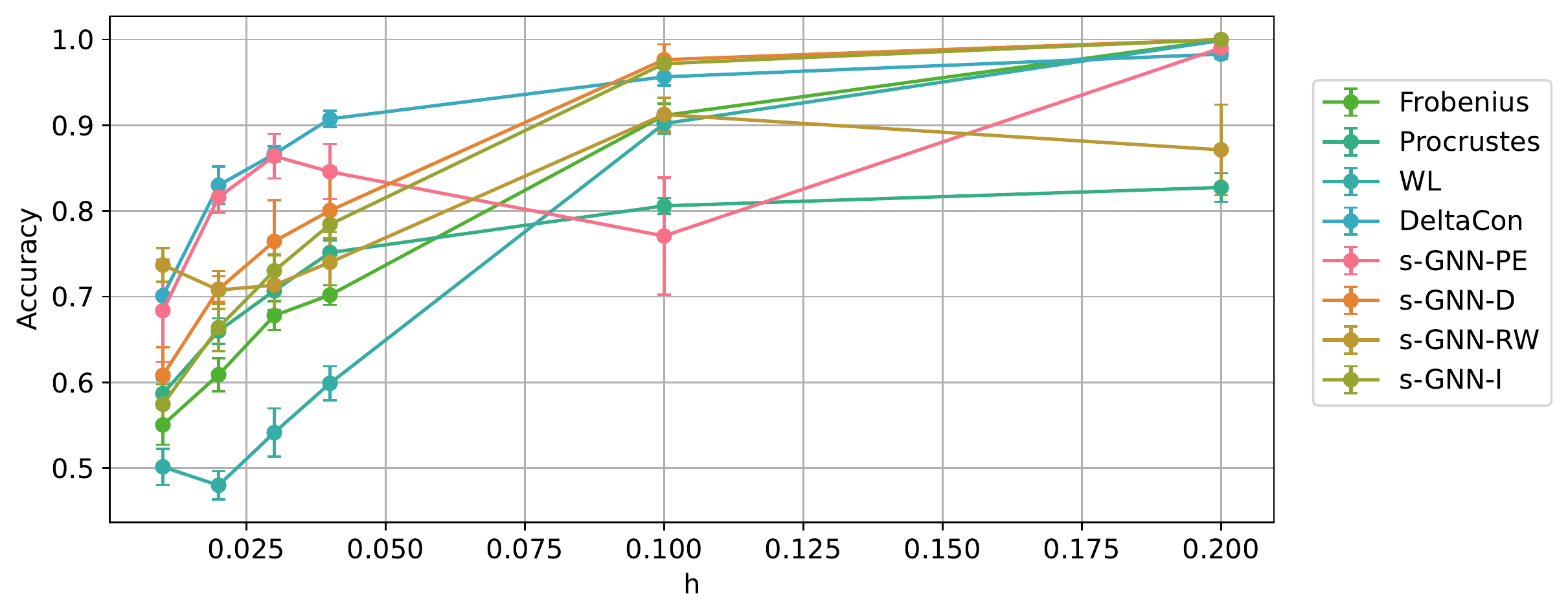}
         \vspace{-3mm}
         \caption{Classification accuracy vs the exchange rate $h$.}
         \label{fig:sc3}
     \end{subfigure}
     \hfill
    \begin{subfigure}[b]{\textwidth}
         \centering
         \includegraphics[width=0.75\textwidth]{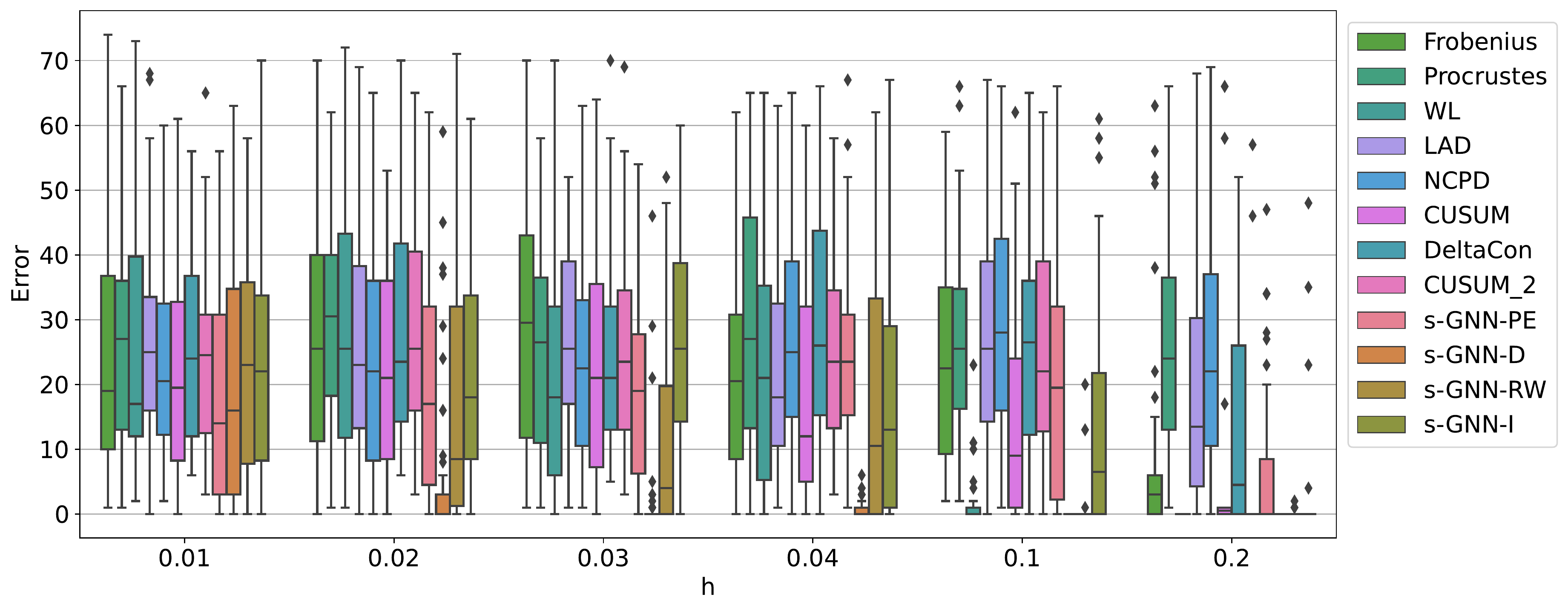}
          \vspace{-3mm}
         \caption{Change point localisation error for different exchange rates $h$.}
         \label{fig:sc3bis}
     \end{subfigure}
     \hfill
     \vspace{-5mm}
     \caption{Performances on the classification (\subref{fig:sc3}) and detection  (\subref{fig:sc3bis}) tasks in the ``Swaps" scenario.} 
    \label{fig:scenario3}
\end{figure}

We test our NCPD method with the four variants of positional encodings (\textbf{Degree}, \textbf{Random Walk}, \textbf{Laplacian} and \textbf{Identity}) defined in Section \ref{sec:sGNN} and report the results of each scenario in Figures \ref{fig:scenario1}, \ref{fig:scenario2}, \ref{fig:scenario2bis} and \ref{fig:scenario3}. In almost all scenarios and difficulty levels, our method outperforms the other baselines, except for the variant with Laplacian attributes. The drop of performance using the latter type of encodings has been previously attributed to the sign ambiguity in Laplacian eigenvectors \cite{dwivedi2022graph}. Moreover, the degree and random walks encodings generally seem to be better than the identity encodings, except for the last scenario. We conjecture that this is due to the fact that in the first three scenarios, nodes in the same cluster are exchangeable in the SBM model, while in the last scenario, this symmetry is broken by the membership exchange mechanism. For networks with a lot of symmetry, the \textbf{Identity} encoding might introduce additional noise.

We also observe that the strongest baselines are CUSUM and CUSUM2, which have better performances if larger window sizes $L$ are used, while our method is not sensitive to this hyperparameter (see Appendix \ref{app:sensitivity}). In particular, our method performs well even for a short history of data and therefore could also detect change-points that are close to each other in a multiple change-point setting. Consequently, these experiments show that using a data-driven graph similarity function leads to better performances in the NCPD task than existing baselines, in various change-point scenarios.
 
 \vspace{-5mm}

\FloatBarrier

\subsection{Correlation network from stock returns data} \label{sec:exp_financial}

This data set comprising the cross-correlation networks of daily stock returns, computed over a one-month interval, from the S\&P 500 index in a period of about 20 years (February 2000 - December 2020). Data sets of stock returns have been previously analysed in different contexts. For online NCPD, \cite{yu2021optimal} and \cite{ Barnett2016} consider the covariance matrices of S\&P 500 weekly log-returns respectively on the period between 1950 and 2000, and between 1982 and 2000, while \cite{dubey2021online} analyses the weekly log-returns of 29 stocks from the Dow Jones Industrial Average index, from April 1990 to January 2012. Closely related to our problem, \cite{chakraborti2020phase} and \cite{samal2021network} cluster market behaviours in the USA S\&P 500 and Japan Nikkei 225 stock networks during the period from 1985 to 2016.

In this analysis, we consider 685 stocks (therefore nodes in the dynamic network) 
alongside additional information of their economic activity during each month. \mc{The correlation networks are built from the time series of  open-to-close (intraday) and close-to-open (overnight) returns. Typically, there are 21 trading days in a calendar month, hence each stock has associated a time series of length 42, since each day of the month contributes with two returns. The resulting stock correlation matrix is the starting point for our network construction. In addition, we employ the following stock properties as node attributes
\begin{itemize} 
\item volatility:  the standard deviation of the above 42  open-to-close and close-to-open returns, based on which the correlation network was built, 
\item volume\_ADV: average daily volume, in shares, over the 21 days of the month, 
\item sharesOut\_ADV: average  shares outstanding, over the 21 days of the month. 
\end{itemize} 
}
We then construct an attributed and unweighted dynamic network $\mathcal{G}_{F} = ((\A_t, E_t))_{1\leq t \leq 244}$ with 244 snapshots, and for each $t$, $\A_t \in \{0,1\}^{685 \times 685}$ and $E_t \in \mathbb{R}^{685 \times 4}$, by truncating the cross-correlation matrices between stocks. More precisely, we set to 1 the matrix entries that are below the 0.1-quantile and above the 0.9-quantile among the entries of all correlation matrices. We note that after this preprocessing step, each graph snapshot is  connected and contains self-loops. A similar procedure  
has been applied in \cite{yu2021optimal}, while other works transform the correlation matrices into complete weighted graphs, e.g., by squaring the correlation coefficients  \cite{chakraborti2020phase} or computing the inverse of the ultra-metric distance \cite{samal2021network}. Here we adopt the sparsifying approach to avoid dealing with a large complete graph.   
In Table \ref{tab:stats_tn}, we report some properties of the resulting network. Finally, we standardize the node attributes matrices $\{E_t\}_{1\leq t \leq 244}$ across the timestamps: for each column (i.e. each attribute) and each matrix, we center and scale its values by the mean and standard deviation of all the values of this attribute in the graph snapshots.

Although previous work reported changes in the behaviour of the stock market following different economic or global events \cite{chakraborti2020phase}, there is no ground-truth knowledge of change-points for this dynamic correlation network. However, there is strong evidence that some major events, such as the ones listed in Table \ref{tab:market_events}, have impacted the dynamics of stock returns and their correlation \cite{Barnett2016}.
Therefore, we consider a self-supervised training procedure \cite{liu2021graph} that first pre-estimates a set of change-points in order to train our s-GNN with the procedure described in Section \ref{sec:methodology}.

We first divide our dynamic network into consecutive windows of 50\%, 20\% and 30\% graph snapshots as training, validation and test sequences. Then we pre-estimate change-points in the training and validation sequences using the following methodology.
It is common practice to cluster stocks into market sectors  \cite{chakraborti2020phase}, and we conjecture that this cluster structure is reflected in the correlation network and is a proxy for the underlying state of the financial market at a given time. Therefore,  
we consider the following three-step procedure:
\begin{enumerate}
    \item We estimate a cluster structure for each correlation matrix 
    using a spectral clustering algorithm based on the Symmetric Signed Laplacian \cite{gallier2016spectral}. 
    \item We compare the obtained node partitions in each pair of matrices 
    using the Adjusted Rand Index and use the latter as a pairwise measure of similarity between correlation graphs. Then we apply a spectral clustering algorithm based on the Normalised Symmetric Laplacian on the graph snapshots (i.e., each snapshot is given a label, interpreted as a state or behaviour of the stock market).
    \item We estimate change-points by ``smoothing" the snapshots' labels: we compute the centroid timestamp of each cluster of snapshots and relabel the latter with the labels of their closest centroid. These new labels now define a partition of the temporal window into consecutive intervals, and therefore pre-estimate change-points in network training sequence.
\end{enumerate}

In the first step, we cluster each correlation matrix into $k = 13$ clusters; this value is chosen by evaluating the silhouette index of the result clustering for different number of clusters $k \in \{10, \ldots, 20\}$. In the second step, we cluster the similarity matrix between the graph snapshots based on the ARI (see Figure \ref{fig:aris} in Appendix \ref{app:details_exp}) into $C = 9$ clusters. We note that in the first step, the clusters correspond to sets of nodes (i.e., stocks) in each graph, while in the second step, the clusters are sets of graph snapshots. The estimated change-points obtained in the third step are plotted in Figure \ref{fig:preCPS}.

Then for the training set, we sample $N = 3000$ pairs of graphs using the ``Random scheme" (see Section \ref{sec:training}) and for the validation set, we use the ``Windowed scheme" with a window of size 12, which leads to 684 pairs. For choosing the hyperparameters of the s-GNN, we test every configuration of  values with the learning rate in the set \hk{$\{ 10^{-5}, 10^{-4}, 10^{-3}, 10^{-2} \},$}
 the weight decay in \hk{$\{10^{-6}, 10^{-5}\}$,}
the dropout rate in $\{0.05, 0.2, 0.4\}$, the \hk{output }size of the Sort-$k$ layer in $\{50, 100, 200\}$, the number  of hidden units in $\{32, 64, 128\}$\hk{. We} select the \hk{model} with the highest F1-score on the validation set \hk{to make predictions on the test set}.

\begin{table}
    \centering
    \begin{tabular}{l l l l}
     \multicolumn{4}{c}{\textbf{Financial network ($T = 244$, $n = 685$)}} \\
     \toprule
            & Mean & Median & Standard deviation \\
          \toprule
         Number of edges per graph & $46.3 \times 10^3$  & $40.3 \times 10^3$  & $23.6 \times 10^3$ \\
         Edge density & 0.20 & 0.17 & 0.10 \\
         Average degree & 135 & 96 & 111 \\
         Average shortest path length & 1.8 & 1.8 & 0.1 \\
         Diameter & 2.8 & 3.0 & 0.5 \\
         \bottomrule
    \end{tabular}
    \caption{Mean, median and standard deviation of network statistics for the snapshots in the correlation network of S\&P index stock returns.}
    \label{tab:stats_tn}
\end{table}

\begin{table}
\centering
\begin{tabular}{ll} \toprule
    Major crashes& Period Date \\ \midrule
    9/11 Financial Crisis & 11/09/2001 \\
    Stock Market Downturn Of 2002 & 09/10/2002  \\
    US Housing Bubble & 2005-2007  \\
    Lehman Brothers Crash & 16/09/2008 \\
    DJ Flash Crash & 06/05/2010 \\
    Tsunami/Fukushima & 11/03/2011 \\
    Black Monday / Stock Markets Fall & 08/08/2011 \\
    Chinese Black Monday & 24/08/2015 \\
    Dow Jones plunge & 02/2018 - 03/2018 \\
    WHO public emergency state (COVID-19) & 30/01/2020 \\\bottomrule
\end{tabular}
\caption{Dates of major crashes and bubbles in the USA  market.}
\label{tab:market_events}
\end{table}

\begin{figure}
    \centering
     \begin{subfigure}[b]{0.95\textwidth}
         \centering
         \includegraphics[width=\textwidth, trim={0cm 0cm 0cm 0cm}, clip]{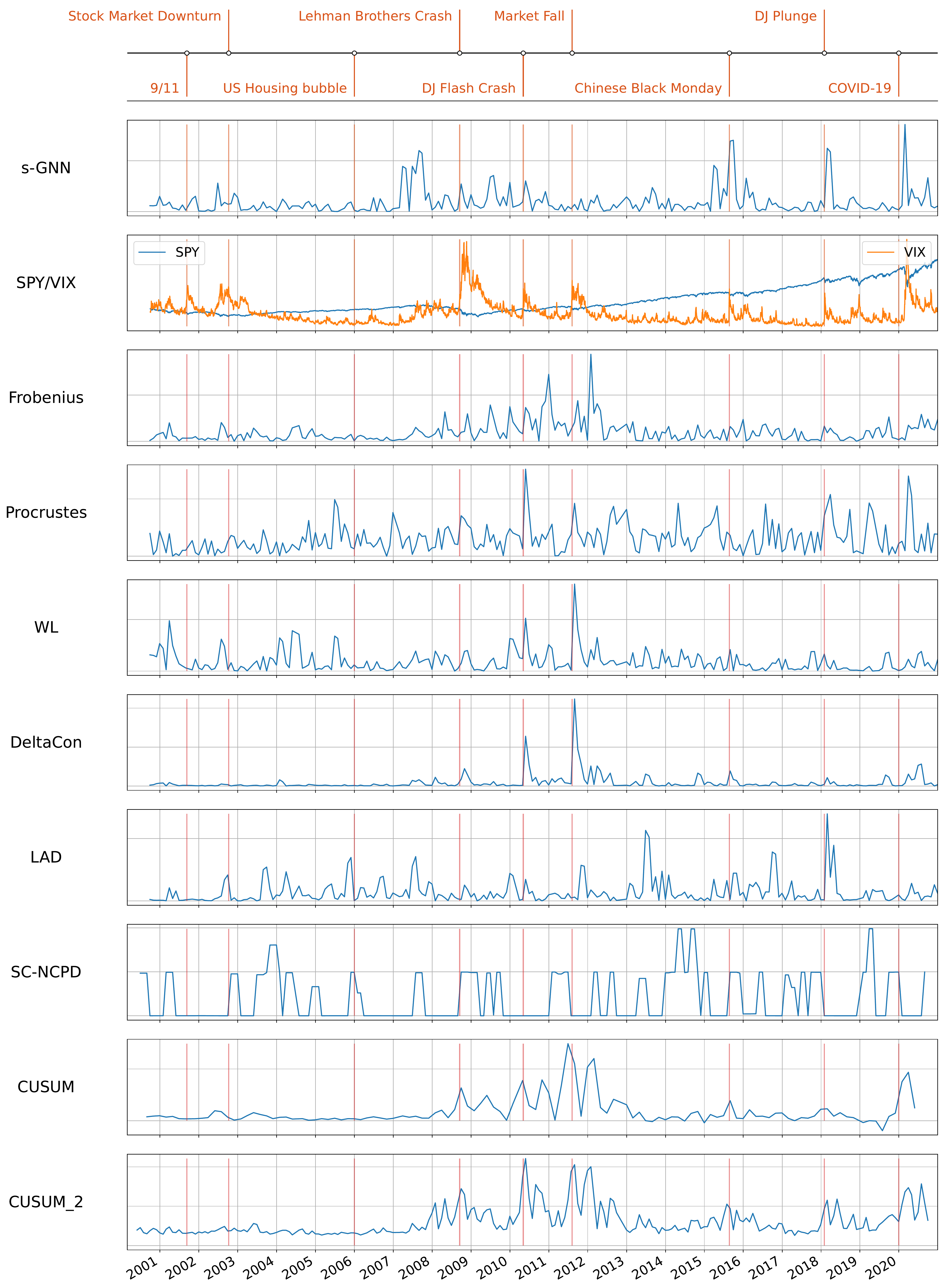}
     \end{subfigure}
     \hfill
     \caption{Change-point detection statistics obtained with our method and the baselines on the dynamic correlation network of S\&P 500 stock returns from February 2000 to December 2020. This period covers a training period from February 2000 to August 2010, a validation period from September 2010 to October 2014 and a test period from November 2014 to December 2020. Main financial events are indicated with vertical red bars. The different rows correspond, from top to bottom, to a timeline of known market events, our NCPD statistic, the SPY and VIX volatility indices and the baselines' NCPD statistics.} 
    \label{fig:stats_fin}
\end{figure}

Finally, we compute our NCPD statistic \eqref{eq:CP_stat} with a window size $L = 6$ over the whole network sequence (i.e., training, validation and test), and qualitatively interpret the time series of its increments $\Delta Z_t(s,L) = |Z_t(s,L) - Z_{t-1}(s,L)|$, plotted in Figure \ref{fig:stats_fin}, alongside the baselines, the SPY and VIX volatility indices and a timeline with the major market events listed in Table \ref{tab:market_events}. 
We observe that our detection statistic exhibits some large peaks at several timestamps coinciding or soon after some market events. The biggest peaks appear in August 2007, August 2015, February 2018 and February 2020 and could be associated with the following financial \mcc{events} and world disruptions.
\begin{enumerate}
    \item The financial crisis of 2007-2008, which began as early as February 2007 when the stock prices in China and the US dramatically fell. 
    \item The Chinese Black Monday in August 2015.
     
    \item The Dow Jones index's plunge in February-March 2018.  
    \item The declaration of \hk{a Public Health Emergency of International Concern} by the World Health Organisation in January 2020 after the emergence of COVID-19.
\end{enumerate}
A secondary peak is also observed in March 2002, which could be a consequence of the 9/11 financial crisis. 

In comparison, the baselines are not able to detect as many events.  Almost all baselines detect change-points between 2010 and 2012, a period when several financial crashes happened such as the 2010 Dow Jone flash crash, 
the Fukushima nuclear incident in March 2011 and the Stock Markets Fall of August 2011. However, most baselines fail to detect anything outside this two-year period. One exception holds for the \textbf{CUSUM 2}, which indicates network disruptions at roughly six periods: during the financial crisis of 2007-2008, Dow Jones Flash Crash in 2010, the Stock Market Fall in 2011, the Chinese Black Monday in 2015, the Dow Jones plunge in 2018 and the consequences of the COVID-19 pandemic in 2020. However, this method delimits some periods of disruptions rather than clear change-points.

\subsection{Correlation networks from physical activity monitoring}\label{sec:exp_physical}

This public data set\footnote{\url{http://www.pamap.org/demo.html}} was built for benchmarking time series classifiers on physical activity monitoring \cite{reiss2012introducing, reiss2012}. This data contains multivariate time series recorded from eight subjects wearing 3D inertial measurement units (IMUs) and performing a protocol of 12 different physical activities such as sitting, walking, descending and ascending stairs and vacuum cleaning. The time series correspond to measurements from 3 IMUs positioned on the subjects' wrist, chest and ankle and containing 3-axis MEMS sensors (an accelerometer, a gyroscope and a magnetometer) with a sampling period of 0.01s.  
Thus, the dimension of the time series is $27$, and there are 8 \hk{time series} in total (one per subject). 

Although this data has also been analysed in the change-point detection task for time series \cite{zhang2020correlationaware}, to our knowledge, it has not been used in the context of NCPD. However, previous work noted that the correlations between pairs of axis are particularly useful for differentiating activities based on translations such as walking, running or ascending stairs \cite{reiss2012}. Therefore, similarly to Section \ref{sec:exp_financial}, for each subject (i.e., each multivariate time series), we build a dynamic correlation network from the $27$ time series, where each node therefore corresponds to an IMU sensor's axis and is associated with a body part.

More precisely, we segment the time series into non-overlapping windows of 100 observations (i.e., a window  of length one second) and compute the correlation matrices of these time series over these windows. Then, the correlation matrices $\{C_t\}_t$ are transformed into binary adjacency matrices $A_t = \mathds{1}_{|C_t|>\eta}, t \in [T]$ with a chosen threshold $\eta =0.2$. We thus obtain an unweighted, unattributed, dynamic network with 27 nodes for each of the 8 subjects. Moreover, each graph snapshot is labelled with the activity performed by the subject during the corresponding temporal window. Therefore, a change of activity between two consecutive snapshots corresponds to a change-point for the network. Table \ref{tab:act_per_subj} summarises the characteristics of each network.

\begin{table}
    \centering
    \begin{tabular}{ccccccccccccccc} \toprule
    \multirow{2}{*}{Subject} & \multicolumn{12}{c}{Activity performed} & \multirow{2}{*}{\makecell{Number of \\ change points}} & \multirow{2}{*}{\makecell{Number of \\ timestamps}} \\ \cmidrule{2-13}
     & 1 & 2 & 3 & 4 & 5 & 6 & 7 & 12 & 13 & 16 & 17 & 24 &  &  \\ \midrule
    1 & \checkmark & \checkmark & \checkmark & \checkmark & \checkmark & \checkmark & \checkmark & \checkmark & \checkmark & \checkmark & \checkmark & \checkmark & 13 & 2490 \\
    2 & \checkmark & \checkmark & \checkmark & \checkmark & \checkmark & \checkmark & \checkmark & \checkmark & \checkmark & \checkmark & \checkmark & \checkmark & 13 & 2618 \\
    3 & \checkmark & \checkmark & \checkmark & \checkmark & \bf{--} & \bf{--} & \bf{--} & \checkmark & \checkmark & \checkmark & \checkmark & \bf{--} & 10 & 1732 \\
    4 & \checkmark & \checkmark & \checkmark & \checkmark & \bf{--} & \checkmark & \checkmark & \checkmark & \checkmark & \checkmark & \checkmark & \bf{--} & 11 & 2302 \\
    5 & \checkmark & \checkmark & \checkmark & \checkmark & \checkmark & \checkmark & \checkmark & \checkmark & \checkmark & \checkmark & \checkmark & \checkmark & 13 & 2709 \\
    6 & \checkmark & \checkmark & \checkmark & \checkmark & \checkmark & \checkmark & \checkmark & \checkmark & \checkmark & \checkmark & \checkmark & \checkmark & 13 & 2487 \\
    7 & \checkmark & \checkmark & \checkmark & \checkmark & \checkmark & \checkmark & \checkmark & \checkmark & \checkmark & \checkmark & \checkmark & \bf{--} & 12 & 2314 \\
    8 & \checkmark & \checkmark & \checkmark & \checkmark & \checkmark & \checkmark & \checkmark & \checkmark & \checkmark & \checkmark & \checkmark & \checkmark & 13 & 2606 \\ \bottomrule
    \end{tabular}
    \caption{Properties of the dynamic networks obtained from the physical activity monitoring data. Some activities are not listed because they have not been performed by any of the subjects in this experiment.}
    \label{tab:act_per_subj}
\end{table}

We then define two NCPD tasks to evaluate our method on, defined as follows.
\begin{itemize}
    \item \textbf{Individual-level NCPD.} Each dynamic network (subject) is used separately and segmented into train, validation and test set. Then we train and test one s-GNN model per network, therefore the learnt graph similarity function is subject-dependent. We note that in this setting, the test sequence contains graphs with unseen activity labels.
    \item \textbf{Cross-individual NCPD.} The eight dynamic networks are combined into a train, validation and test set and we train only one s-GNN model for all sequences. In this case, our method learns only one graph similarity function for all the subjects and its performances evaluated on all of them.
\end{itemize}
More precisely, in the  \textbf{Individual-level NCPD} task, we randomly split each dynamic network into 70\% training and validation, and 30\% test, by isolating a test interval and a validation interval. The latter is a window of size 60 centered around a uniformly sampled change-point, while the lower end of the test interval is uniformly sampled in the whole sequence. We then sample 4000 pairs from the training sequence and 1000 pairs from the validation interval, and train and evaluate our method on each subject separately. In the \textbf{Cross-individual NCPD} task, we design the following two evaluation settings.
\begin{enumerate}
    \item Random split: we concatenate all the training, validation and test sequences from the previous task, as well as the training and validation sets of pairs. Then we subsample respectively 15000 and 3000 pairs with the \textbf{Random scheme} (see Section \ref{sec:training}). Note that in this case, the s-GNN is trained and tested on sub-sequences from every dynamic network.
    
    \item Leave-one-subject-out (LOSO): in this setting we keep the whole sequence of one subject for testing, and train and validate on the seven remaining sequences. For the latter, we select validation intervals as in the previous task, and sample 3000 training and 1000 validation pairs in each network. Then we subsample 15000 and 3000 pairs from the aggregated training and validation sets.

\end{enumerate}

One can get an insight on the feasibility of these two tasks by looking at (a) the similarity of adjacency matrices within each dynamic network, grouped by activity labels; (b) the similarity of adjacency matrices within the same activity, grouped by network (i.e., subject). We report in Appendix \ref{app:similarity_act} some insights on the dissimilarity between activities and subjects, measured in terms of the Frobenius distance. In Figure \ref{fig:av_corr_0}, we plot the average adjacency matrices in each activity (the average of all matrices corresponding to the same activity) from the first subject. We note that some of these matrices are quite similar, like the ones for activities $\{1,2,3\}$, that correspond respectively to sitting, lying and standing, which are all static activities. In Figure \ref{fig:av_dist_sub_0}, we plot the Frobenius distance between these matrices. We observe that the lowest values of this distance matrix are mostly located on the diagonal, indicating that the graphs that have the same label are more similar to each other than graphs with different labels. In Figure \ref{fig:dist_act}, we represent the Frobenius distances between the graphs of different subjects with the same label, for four activities. We observe that for activities 5 and 7, the distances between matrices of different subjects are bigger than the ones from the same subject, however this difference is not always significant and does not seem to appear for activities 1 and 17. Figure \ref{fig:av_dist_subj} confirms this observation: we note that the average Frobenius distance between the graphs with the same label and from different subjects is smaller than the average distance between graphs with different labels.  

For our change-point statistic, we use sliding windows of $L=20$ timestamps. We report the performances of our method and baselines in terms of the adjusted F1-score \cite{xu18} in Table \ref{tab:res_exp1}, with a tolerance window of $\pm 5$ timestamps. Our method has the best performance in most evaluation settings, and largely outperforms the baselines in the LOSO scheme. This latter result seems to indicate that the s-GNN is able to learn a graph similarity function that is generalisable to unseen subjects, while the good performances in the \textbf{Individual-level} task suggests that it can generalise to unseen activities.

\begin{table}
    \centering
  \begin{tabularx}{\textwidth}{ 
   >{\centering\arraybackslash}X 
   >{\centering\arraybackslash}X 
   >{\centering\arraybackslash}X 
   >{\centering\arraybackslash}X 
   >{\centering\arraybackslash}X 
   >{\centering\arraybackslash}X  }
          \toprule
\multicolumn{6}{c}{\textbf{Individual-level NCPD}} \\
\midrule
         Subject &  s-GNN-I & Frobenius & SC-NCPD & CUSUM & CUSUM 2  \\
         \midrule
1 &  0.76 (0.20) &  0.62 (0.31) &  \textbfr{0.82 (0.14)} &  0.54 (0.30) &  \textbfb{0.81 (0.20)} \\
2 &  \textbfr{0.91 (0.11)} &  0.45 (0.22) &  0.61 (0.08) &  0.45 (0.12) &  \textbfb{0.84 (0.11)} \\
3 &  \textbfb{0.60 (0.18)} &  0.37 (0.14) &  \textbfr{0.67 (0.15)} &  0.21 (0.27) &  0.34 (0.26) \\
4 & \textbfr{0.73 (0.18)} &  0.58 (0.26) &  \textbfb{0.70 (0.08)} &  0.60 (0.07) &  0.59 (0.22) \\
5 &  \textbfr{0.85 (0.19)} &  0.61 (0.22) &  0.72 (0.16) &  0.36 (0.24) &  \textbfb{0.72 (0.13)} \\
6 &  \textbfb{0.74 (0.19)} &  0.73 (0.22) & \textbfr{0.75 (0.17)} &  0.56 (0.30) &  0.58 (0.16) \\
7 &  \textbfr{0.90 (0.13)} & \textbfb{0.79 (0.19)} &  0.67 (0.35) &  0.57 (0.23) &  0.72 (0.37) \\
8 &  0.72 (0.24) &  \textbfr{0.88 (0.13)} &  0.65 (0.14) &  0.57 (0.28) &  \textbfb{0.82 (0.13)} \\
         \bottomrule
    \end{tabularx}
    \begin{tabularx}{\textwidth}{ 
    >{\centering\arraybackslash}X 
   >{\centering\arraybackslash}X
   >{\centering\arraybackslash}X
   >{\centering\arraybackslash}X
   >{\centering\arraybackslash}X
   >{\centering\arraybackslash}X  }
          \toprule
\multicolumn{6}{c}{\textbf{Cross-individual NCPD}} \\
\midrule
Setting   & s-GNN-I & Frobenius & SC-NCPD & CUSUM & CUSUM 2 \\
         \midrule
Random split  &     \textbfr{0.81 (0.07)} & 0.75 (0.03) &  0.75 (0.03) &  0.59 (0.12) &  \textbfb{0.80 (0.04)}  \\
LOSO  &     \textbfr{0.89 (0.02)} & 0.70 (0.20) & \textbfb{0.77 (0.06)} &  0.62 (0.11) &  0.75 (0.12) \\
        \bottomrule
    \end{tabularx}
    \caption{Adjusted F1-score of our method and  baselines in the \textbf{Individual-level} and \textbf{Cross-individual} NCPD tasks on the physical activity monitoring data. \mc{The red, respectively blue, bold values in each row  denote the top best, respectively second best, performing methods}. The values in the parentheses denote the standard deviation over 10 repetitions of the random splits train/validation/set, except for the leave-one-subject-out (LOSO) setting for which mean and standard deviation are computed over the 8 folds.}
    \label{tab:res_exp1}
\end{table}

\section{Conclusion} \label{sec:discussion}

In this work, we proposed a novel method for detecting change-points in dynamic networks using a data-driven graph similarity function. The latter is learnt by a siamese GNN model, trained and validated on pairs of graphs from the network sequence. This similarity function allows to effectively compare the current graph to its short-term history for detecting potential displacements, with a simple  online statistic. We demonstrated the benefits of our method in synthetic experimental settings of dynamic SBMs, and on two real-world data sets of correlation networks, and concluded that our method is more accurate at distinguishing graphs with different generative distributions and detecting change-points, compared to existing baselines.

As previously noted, one main challenge posed by using a deep-learning based model for NCPD is the training and validation procedures, which necessitate either a data set with change-point labels, or an adequate unsupervised or self-supervised learning procedure. Since the former is quite rare, a future direction for this work could to develop the latter approach, for instance using data augmentation strategies \cite{Carmona2021NeuralCA} for introducing artificial change-points in the training set. Another possible extension would be to adapt our framework to more general types of dynamic networks, e.g., snapshots with varying node sets or with missing edges.  
\mc{In certain application domains, it may well be the case that change-points phenomena are localized only in certain parts of the network (as considered in some of our synthetic experiments), and are not affecting the global structure}. \mcc{To this end, yet another interesting addition to the current framework is to be able to pinpoint specifically which part of the network is mainly driving the change-point, to enhance explainability.}

\mcc{Finally, extending the methodology to different types of networks, such as directed networks, is an interesting direction to explore, especially in the context of recent work in the literature that encodes various measures of causality or lead-lag associations in multivariate time series as directed graphs \cite{leadLagTS,Runge2017}. The structure of such weighted directed graphs may evolve over time, which motivates the need for techniques for change-point detection, a setting where adapting traditional spectral methods for change-point detection would be challenging, due to the asymmetry of the adjacency matrix.}

\bibliographystyle{unsrt}
\bibliography{bib}

%


\newpage

\begin{appendix}

\section{Additional details and analysis in the synthetic experiments}\label{app:details_exp}

In this section, we provide additional details on the hyperparameter selection procedure in the synthetic networks of Section \ref{sec:exp_synthetic}, and two supplementary experiments, namely a sensitivity analysis of our method to the window size parameter $L$ and to the choice of pooling layer in the similarity module (see Section \ref{sec:methodology}).

\subsection{Hyperparameter selection}

In each scenario and difficulty level, we train our s-GNN over 100 epochs and and validate using the F1-score. We select one set of hyperparameters of the s-GNN (i.e., learning rate, number of hidden units, dropout rate and size of Sort-$k$ layer) per scenario by searching over a grid of values in one difficulty level, i.e., $p = 0.03$  in Scenario 1, $s=60$ in Scenario 2,   $p = 0.06$  in Scenario 3, and $h= 0.1$ in Scenario 4. For choosing the hyperparameters  we test every configuration of  values with the learning rate in the set $\{0.001, 0.01\}$, the dropout rate in $\{0.01, 0.05, 0.1\}$, the size of the Sort-$k$ layer in $\{20, 40, 100\}$, the number  of hidden units in $\{16, 32, 64\}$, and select the one with the highest F1-score on the validation set.

For each graph distance baseline $d$, we use the training and validation sets to choose a classification threshold $\theta$ such that the estimated label $\hat{y}_i = 1$ if $d(G_1^i, G_2^i) < \theta$ and $\hat{y}_i = 0$ otherwise (or reversely for the WL kernel).
 
\subsection{Sensitivity to the window size}\label{app:sensitivity}

We evaluate the sensitivity of our NCPD method and the baselines to the window size parameter $L$ in the \textbf{``Merge"} scenario from Section \ref{sec:exp_synthetic}. We recall that this hyperparameter corresponds to the amount of past (and future for some baselines) information needed to compute the NCPD statistic. It is therefore also the minimal distance between change-points that a method can detect. In this analysis, we test the performances of the methods using different window sizes in a  a synthetic setting from Section \ref{sec:exp_synthetic}. More precisely, we consider Scenario 1 (``Merge") and three difficulty levels ($p = 0.3, 0.4, 0.5$). 

\begin{figure}
    \centering
    \begin{subfigure}[b]{0.8\textwidth}
         \centering
         \includegraphics[width=\textwidth]{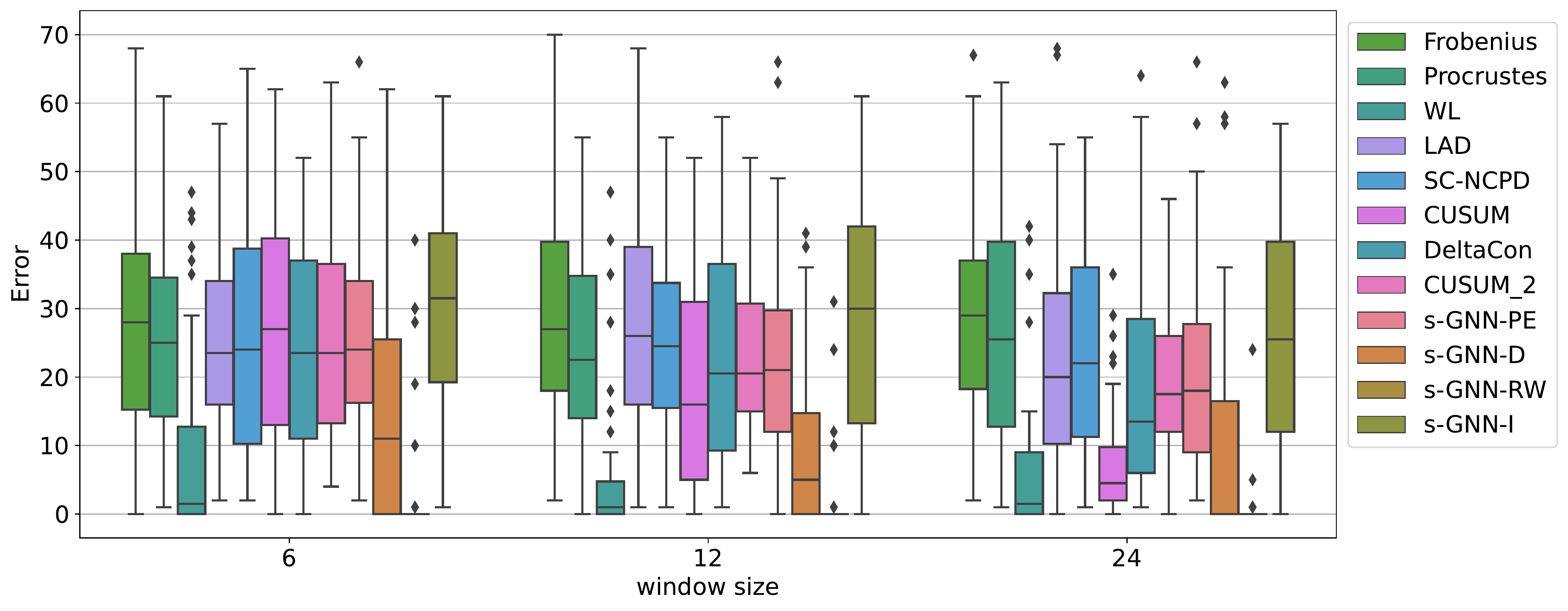}
         \caption{Difficult level}
          \label{fig:diff1}
     \end{subfigure}
     \hfill
     \begin{subfigure}[b]{0.8\textwidth}
         \centering
         \includegraphics[width=\textwidth]{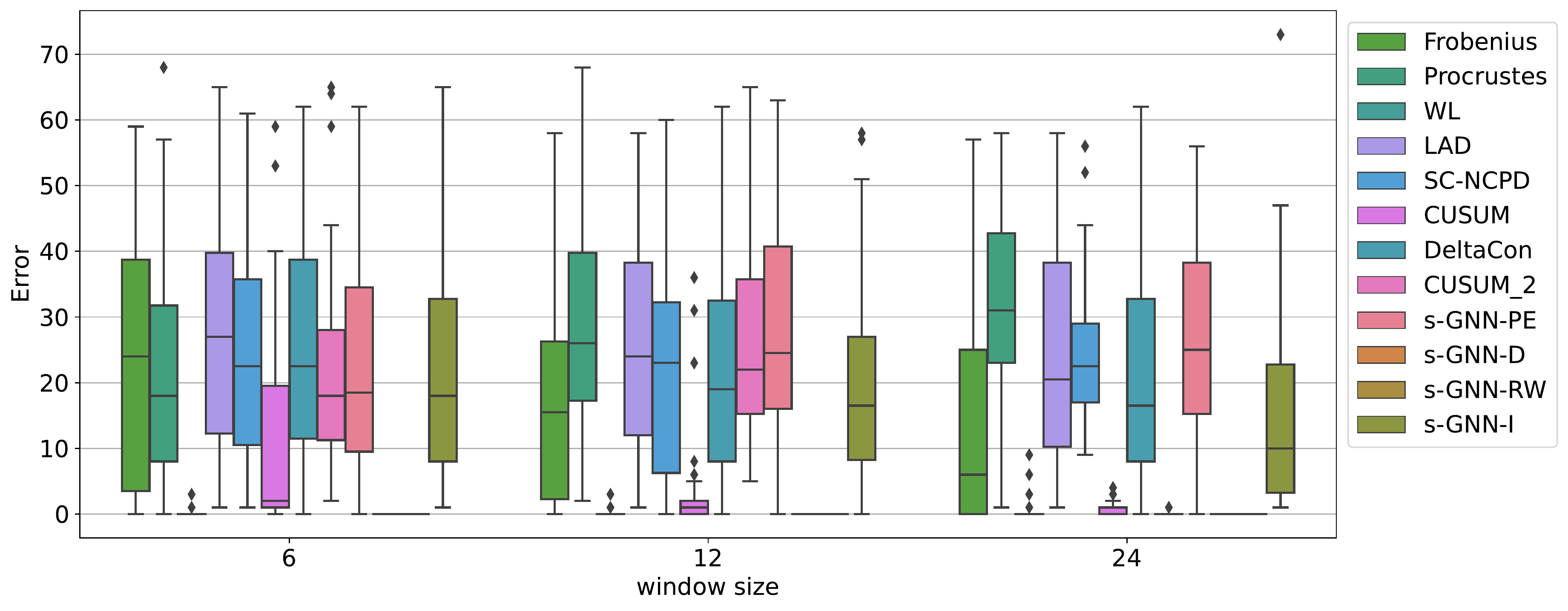}
    \caption{Moderate level}
          \label{fig:diff2}
     \end{subfigure}
     \hfill
          \begin{subfigure}[b]{0.8\textwidth}
         \centering
         \includegraphics[width=\textwidth]{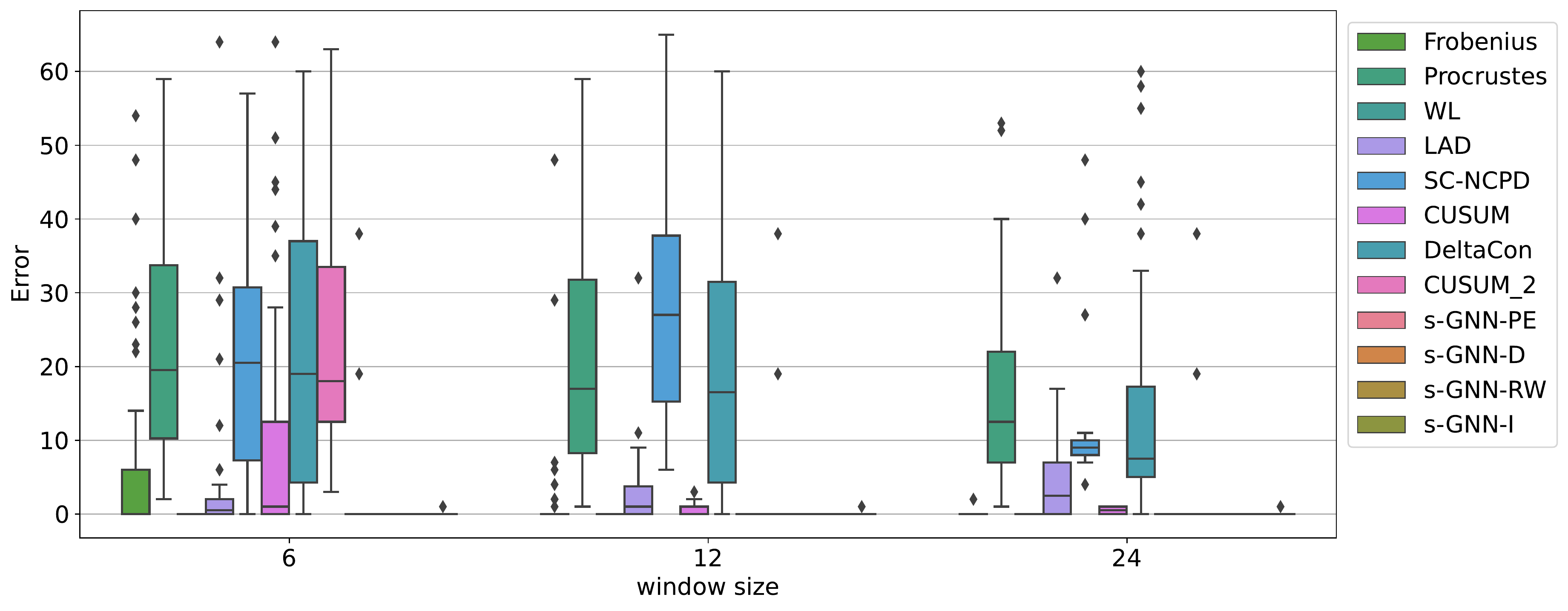}
    \caption{Easy level}
          \label{fig:diff3}
     \end{subfigure}
     \hfill
     \caption{Performances on the detection task in the \textbf{``Merge"} scenario for 3 window sizes $L = 6, 12, 24$ in three difficulty levels: difficult ($p = 0.3$) (\subref{fig:diff1}), moderate  ($p = 0.4$) (\subref{fig:diff2}), easy ($p = 0.5$) (\subref{fig:diff3}).}
    \label{fig:sensitivity_window}
\end{figure}

We report our findings in Figure \ref{fig:sensitivity_window}. We note that our method is not very sensitive to the window size, in particular our best variant (\textbf{s-GNN-RW}) outperforms the baselines for all window sizes. We also remark that the two methods based on the CUSUM statistic (\textbf{CUSUM} and \textbf{CUSUM 2}) have better performances for larger $L$, and this effect is larger than for the other baselines. In conclusion, the choice of window size in our NCPD statistic \eqref{eq:CP_stat} does not have a big impact on the performance of our method, and therefore does not require to be finely tuned. 

\subsection{Sensitivity to the pooling layer}

\begin{figure}
    \centering
    \includegraphics[width=0.7\textwidth]{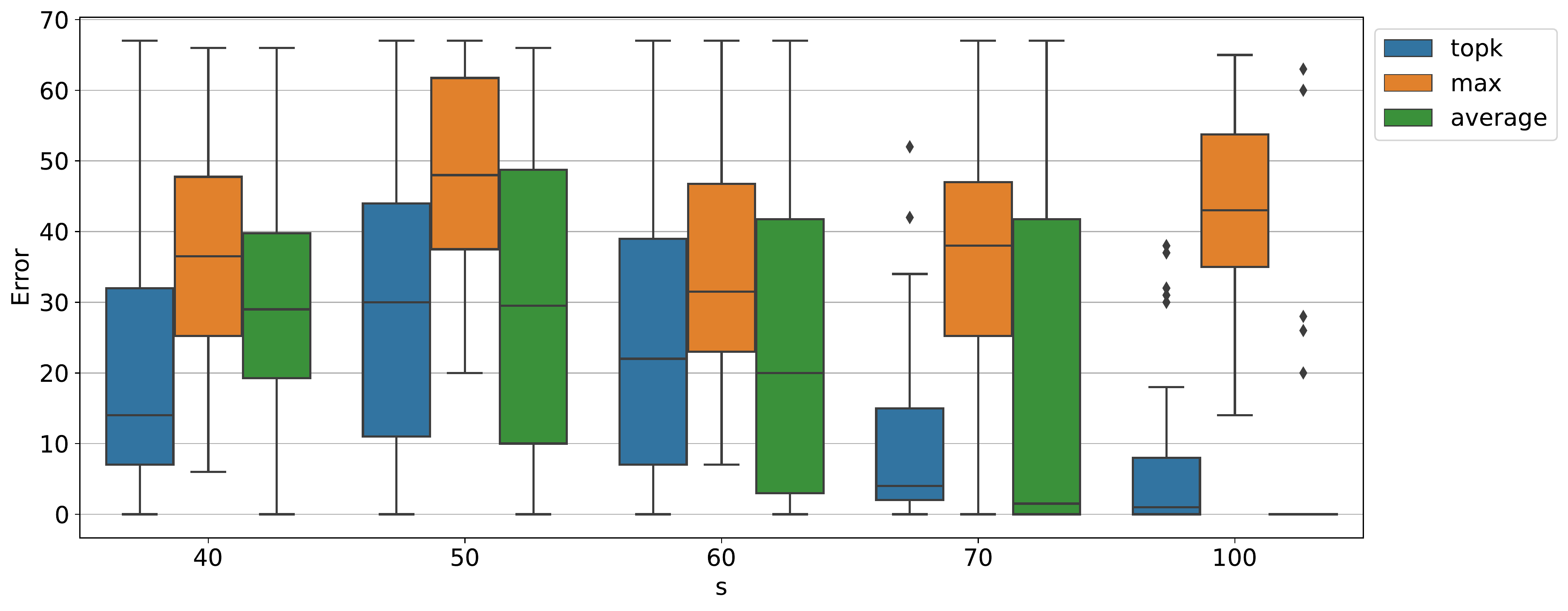}
    \caption{Localisation error of the s-GNN in the \textbf{``Birth 1"} scenario with different choices of pooling layers in the similarity module, namely Sort-$k$, Max and Average pooling.}
    \label{fig:sens_pooling}
\end{figure}

In this section, we test the importance of using a Sort-$k$ pooling layer in our similarity module (Figure \ref{fig:similarity_module}). We consider the \textbf{``Birth 1"} scenario from Section \ref{sec:exp_synthetic} and compare the performance of our method with Sort-$k$ pooling ($k=100$) with the same method with Max or Average pooling. We report our findings in Figure \ref{fig:sens_pooling}. We note that Max pooling does not have good performances in this experiment and Average pooling has a higher variance than Sort-$k$, except in the last (and easiest setting). It may be due to the fact that Max pooling is less robust to the sparsity of the network than Sort-$k$ and Average pooling, and that the latter cannot detect local changes in the graphs since it averages the displacement over the whole set of nodes. Therefore, we can conclude that Sort-$k$ pooling is more adapted to detect small distribution changes, while being robust to the sparsity of edges.

 \section{Additional results on the S\&P 500 stock returns data set}\label{app:stocks}
 
 In this section, we first report additional figures illustrating the procedure for pre-estimating change-points in the training and validation sequences of the dynamic correlation network; secondly we analyse the network using the eigen-entropy $H(t)$, as previously done in \cite{chakraborti2020phase} on similar data. This latter analysis also gives an insight on the possible market phases (i.e., the period in-between our estimated change-points).
 
\begin{figure}
    \centering
    \includegraphics[width=\textwidth]{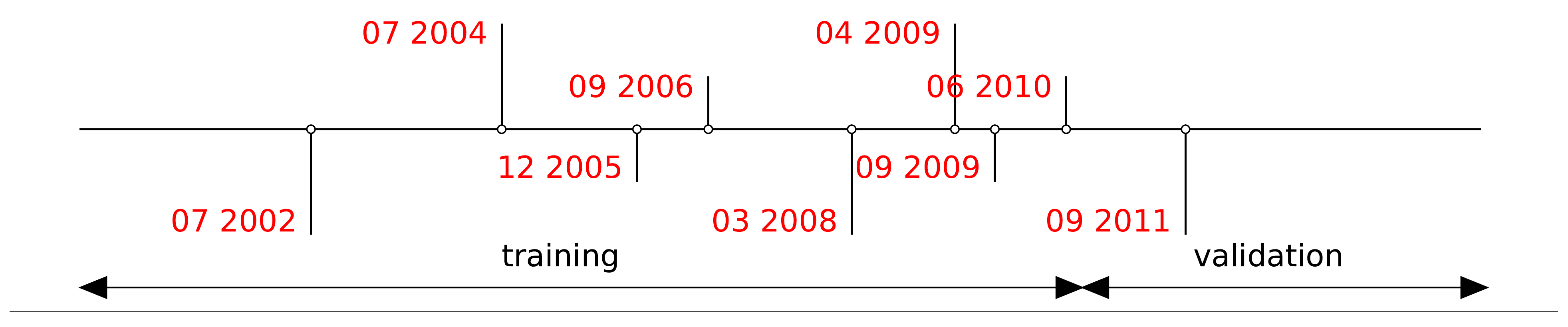}
    \caption{Pre-estimated change-points on the training and validation sequences of the  S\&P 500 stock returns correlation network.}  
    \label{fig:preCPS}
\end{figure}

 In Figure  \label{fig:aris}, we plot the heatmap of the similarity matrix between the graph snapshots in the training and validation sequences. We recall from Section \ref{sec:exp_financial} that the similarity score between pairs of snapshots is measured in terms of the Adjusted Rand Index values between the stock partitions obtained for each snapshots. We note that this similarity matrix seems to have a cluster structure; in particular, high similarity scores can be found during the period of the financial crisis from 2007 to 2011 and in 2001-2002. The clustering procedure of this matrix with spectral clustering and a post-processing step (see \ref{sec:exp_financial}) leads to the pre-estimated change-points plotted in Figure \ref{fig:preCPS}.
 
 \begin{figure}
    \centering
    \includegraphics[width=0.5\textwidth]{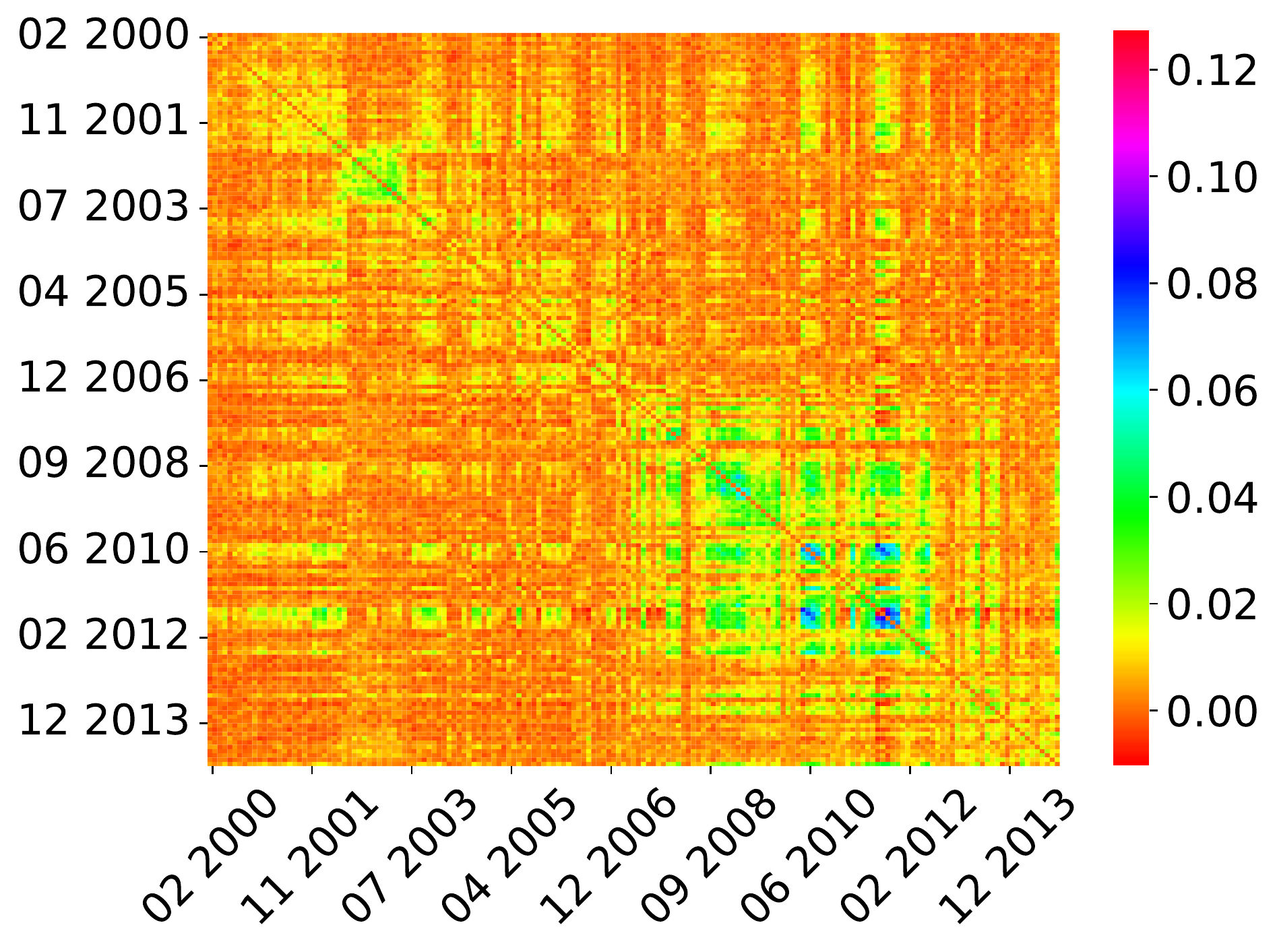}
    \caption{Matrix of Adjusted Rand Index values between the partitions obtained for each pair of graph snapshots in the correlation network of S\&P 500 stock returns. \mcc{The first two digits denote the month, followed by the year.} } 
    \label{fig:aris}
\end{figure}

\paragraph{Analysis of the eigen-entropy}

\begin{figure}
    \centering
    \includegraphics[width=0.9\linewidth]{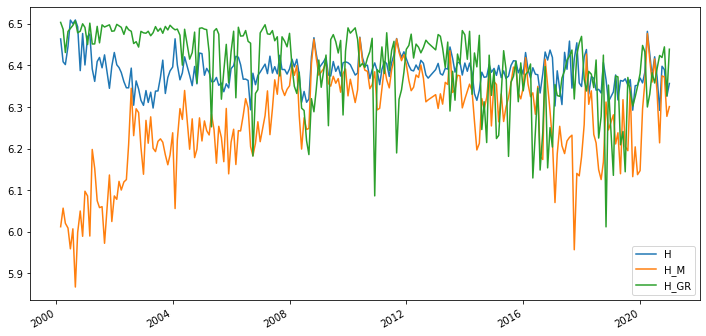}
    \caption{Eigen-entropy of the correlation graph, the market mode and the group plus random mode over time.}
    \label{fig:eigen_entropy}
\end{figure}

\begin{figure}
    \centering
    \includegraphics[width=0.32\linewidth]{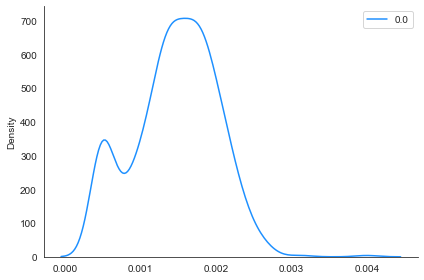}
        \includegraphics[width=0.32\linewidth]{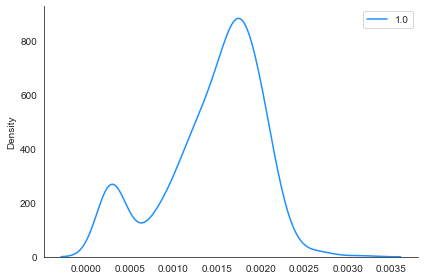}
            \includegraphics[width=0.32\linewidth]{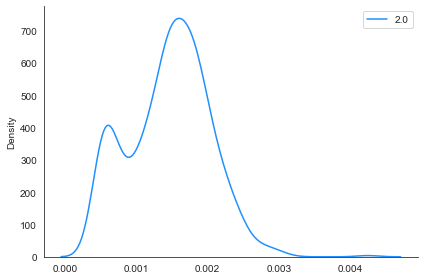}
                \includegraphics[width=0.32\linewidth]{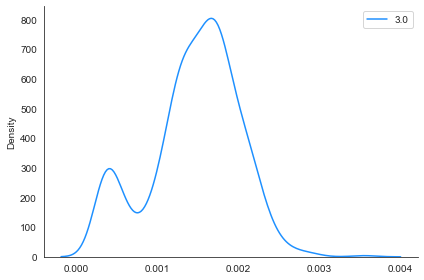}
        \includegraphics[width=0.32\linewidth]{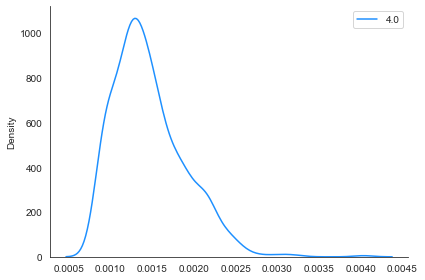}
            \includegraphics[width=0.32\linewidth]{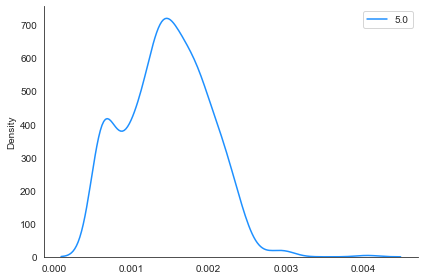}
                \includegraphics[width=0.32\linewidth]{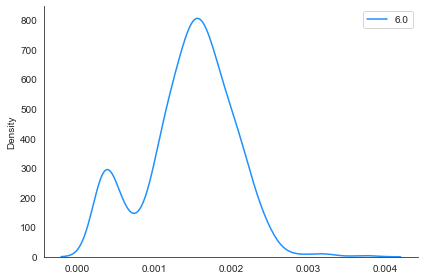}
        \includegraphics[width=0.32\linewidth]{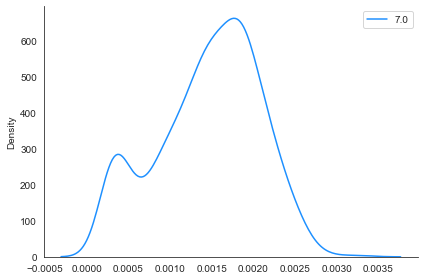}
        \includegraphics[width=0.32\linewidth]{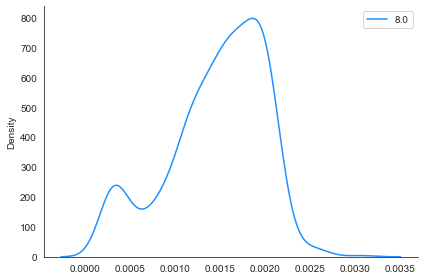}
        \includegraphics[width=0.32\linewidth]{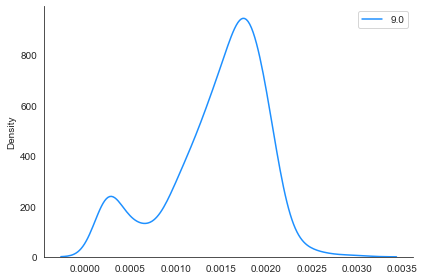}
        \includegraphics[width=0.32\linewidth]{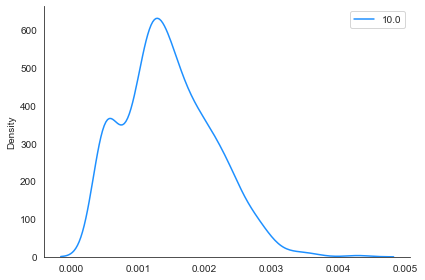}
    \caption{Histogram of the values in the average eigen-centrality vector of each class (phase) of graphs (the subplots correspond to different classes).}
    \label{fig:centrality_dist}
\end{figure}

\begin{figure}
    \centering
    \begin{subfigure}[b]{0.45\textwidth}
    \includegraphics[width=\linewidth]{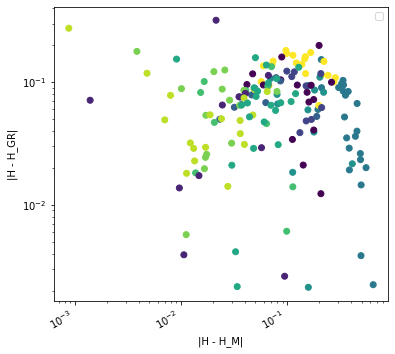}
    \caption{Entropy differences $|H - H_{GR}|$ versus $|H - H_{M}|$.}
    \label{fig:phase_space}
    \end{subfigure}
    \hfill
    \begin{subfigure}[b]{0.45\textwidth}
    \includegraphics[width=\linewidth]{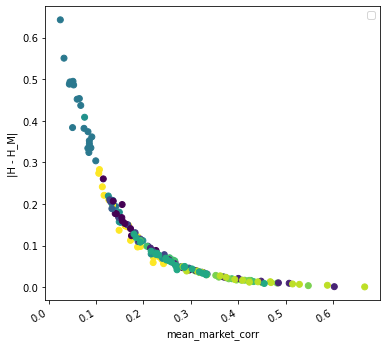}
    \caption{Entropy differences $|H - H_{M}|$ versus mean market correlation.}
    \label{fig:entropy_mmc}
    \end{subfigure}
    \hfill
    \caption{Entropy differences $|H - H_{GR}|$ versus $|H - H_{M}|$ (in log scale) (\subref{fig:phase_space}) and entropy differences $|H - H_{M}|$ versus mean market correlation (\subref{fig:entropy_mmc}) for the graphs in the financial correlation dynamic network. The colors indicate the class of the graphs in our partition.}
    \label{fig:phase_entropy}
\end{figure}

We first analyse the correlation structures of our graphs using the eigen-entropy $H(t)$ \cite{chakraborti2020phase}. The eigen-entropy of a graph is defined as the entropy of the eigen-centrality vector, which is a $L_1$-normalised version of the principal eigenvector of the graph adjacency matrix. This principal eigenvector is related to the relative ranks of the different stocks in the market and its entropy measures the market ``disorder". The correlation matrices $C(t)$'s can be further decomposed into a market mode (principal eigen matrix) $C(t)_M$ and a composite group plus random mode $C(t)_{GR}$ and their corresponding eigen-entropy $(H_M(t), H_{GR}(t)$ can be also computed (see Figure \ref{fig:eigen_entropy}). This allows to define a 3D-phase space where the graphs can be separated into types (e.g. market anomalies, crashes, normal behaviour or highest disorder in \cite{chakraborti2020phase}). A 2D visualisation of our graphs with their corresponding labels is given in Figure \ref{fig:phase_space}. The distribution of the average eigen-centrality vectors in each class of graphs also indicates that the correlation structure changes in the different phases (see Figure \ref{fig:centrality_dist}).  \cite{chakraborti2020phase} also observe a scaling behaviour by comparing the absolute entropy difference $|H - H_{GR}|$ and the mean market correlation (see Figure \ref{fig:entropy_mmc}).

\section{Additional experimental results on the physical activity monitoring data}

\subsection{Similarity between activities and subjects}\label{app:similarity_act}

\begin{figure}
    \centering
    \begin{subfigure}[b]{0.29\textwidth}
         \centering
    \includegraphics[width=\linewidth, trim={0cm 0cm 3cm 0.8cm}, clip]{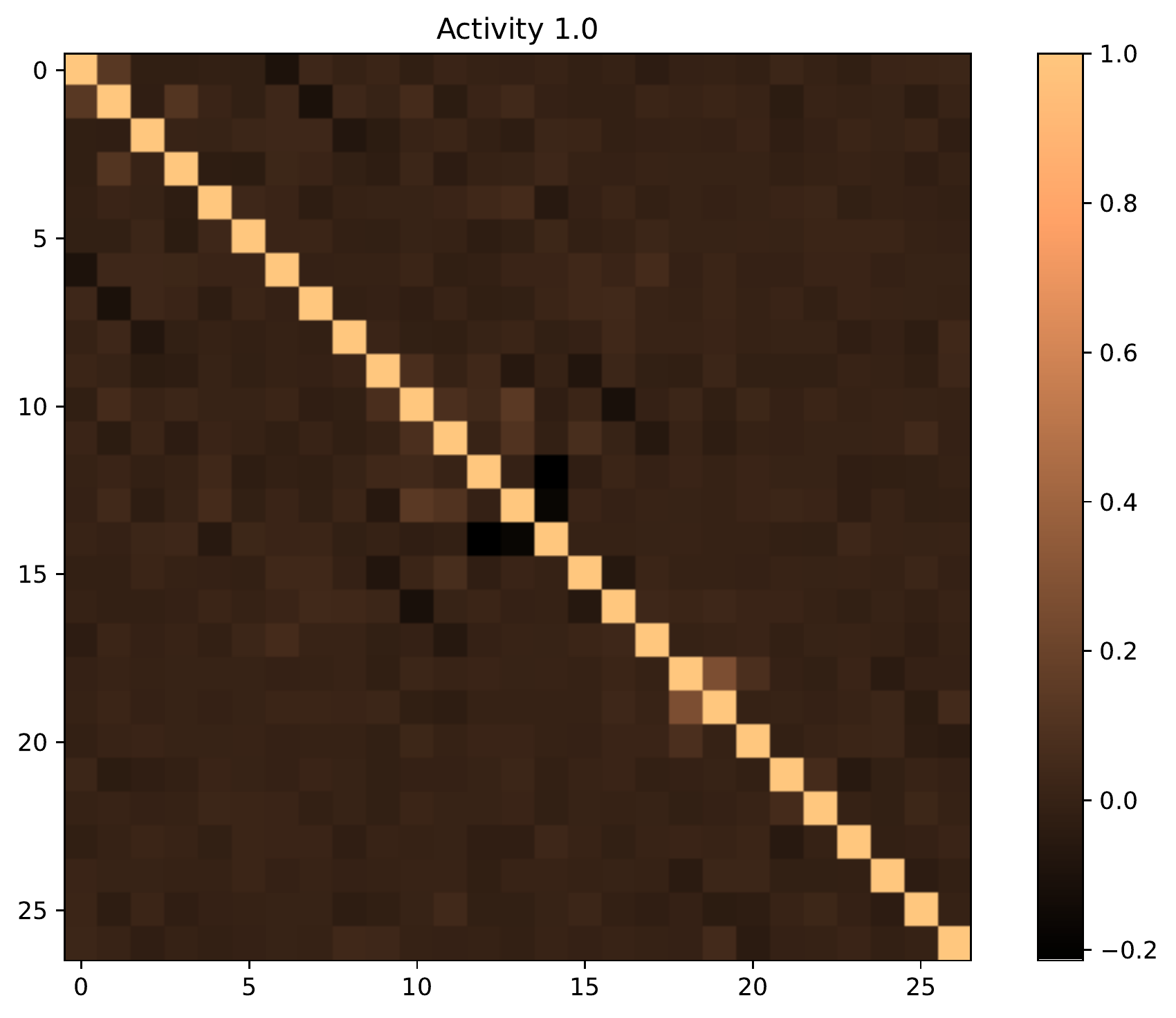}
    \caption{Activity 1}
    \end{subfigure}
    \hfill
    \begin{subfigure}[b]{0.29\textwidth}
         \centering
    \includegraphics[width=\linewidth, trim={0cm 0cm 3cm 0.8cm}, clip]{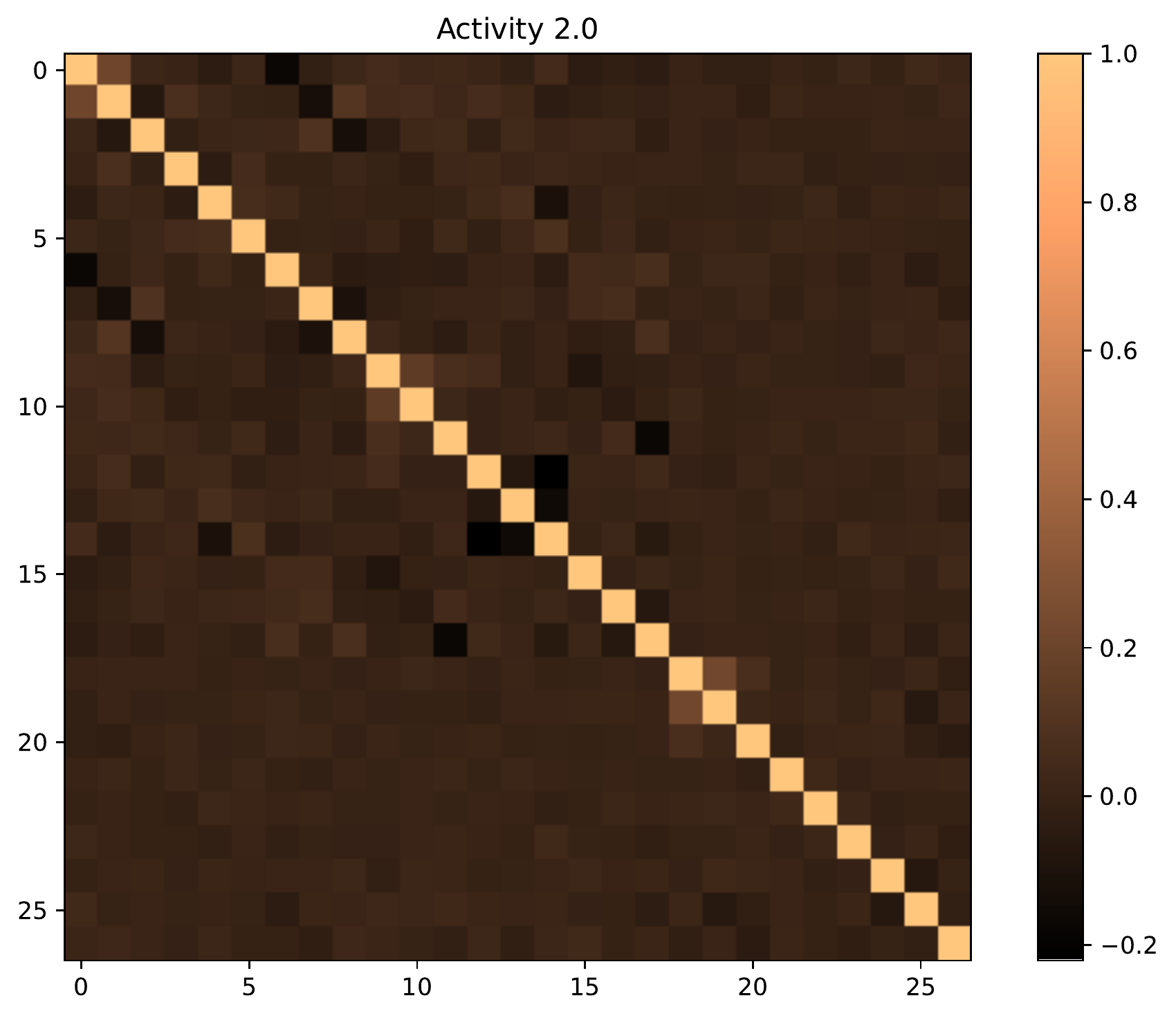}
    \caption{Activity 2}
    \end{subfigure}
    \hfill
    \begin{subfigure}[b]{0.35\textwidth}
         \centering
    \includegraphics[width=\linewidth, trim={0cm 0cm 0cm 0.5cm}, clip]{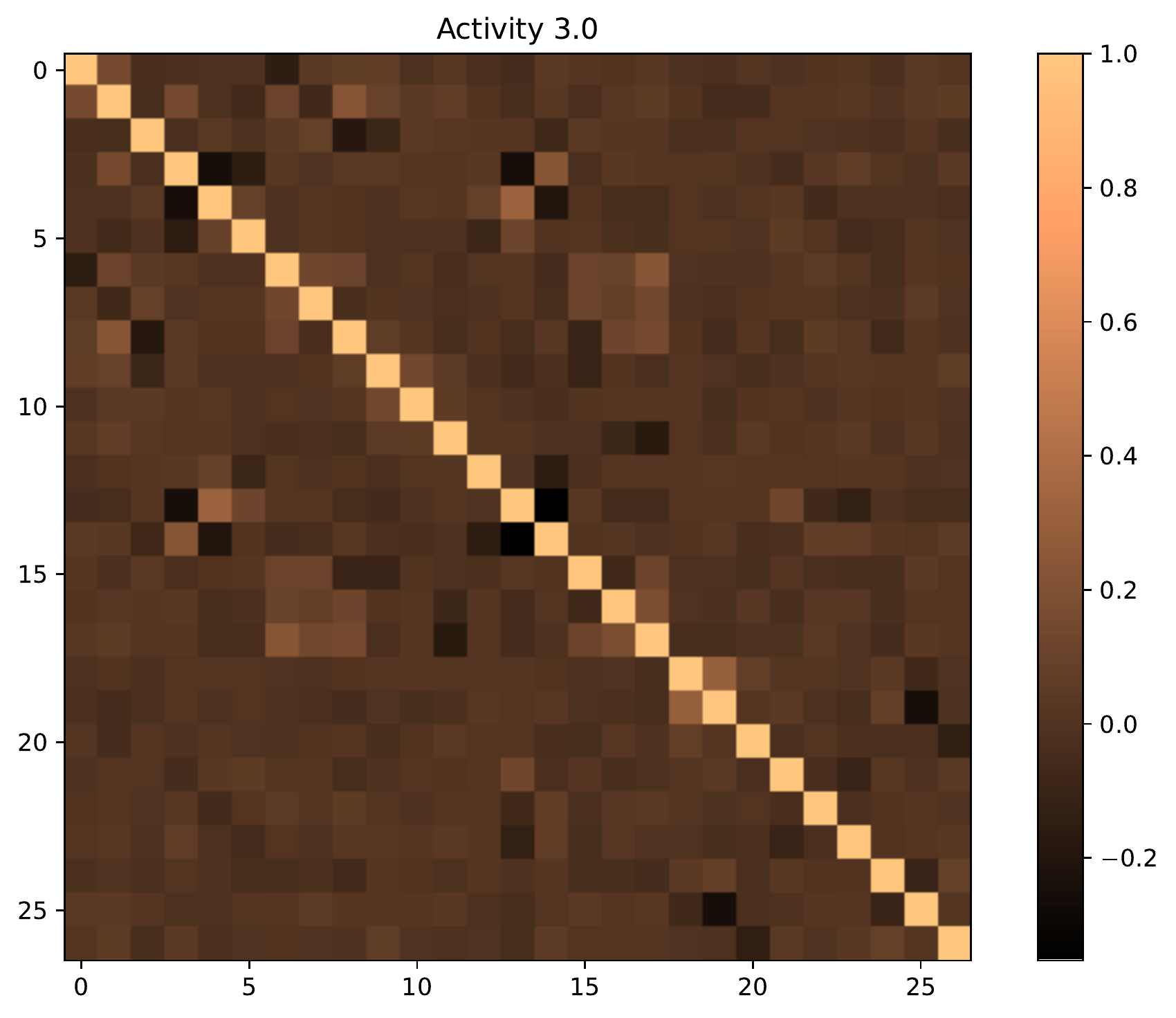}
    \caption{Activity 3}
    \end{subfigure}
    \hfill
        \begin{subfigure}[b]{0.3\textwidth}
         \centering
    \includegraphics[width=\linewidth, trim={0cm 0cm 3cm 0.8cm}, clip]{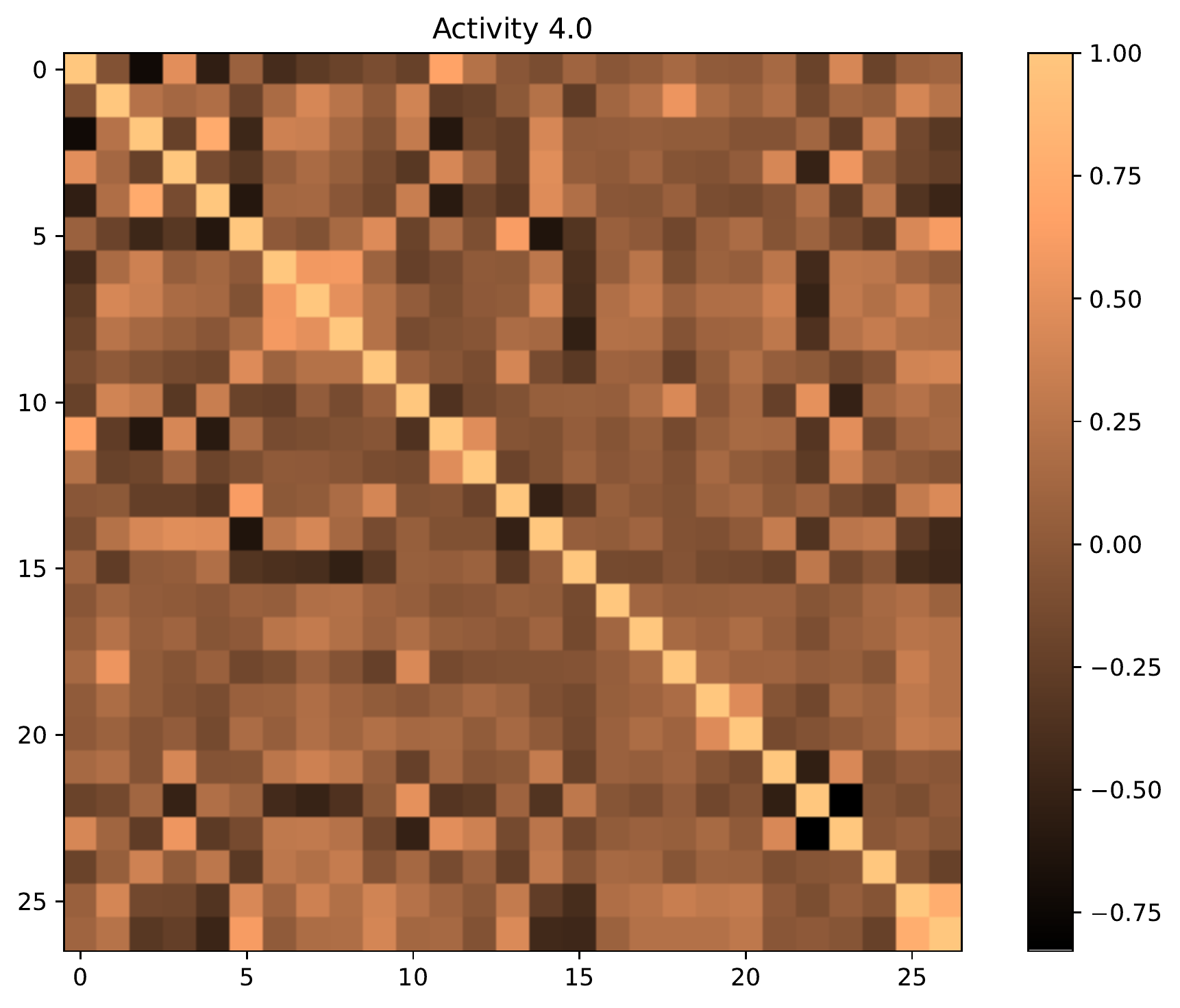}
    \caption{Activity 4}
    \end{subfigure}
    \hfill
        \begin{subfigure}[b]{0.3\textwidth}
         \centering
    \includegraphics[width=\linewidth, trim={0cm 0cm 3cm 0.8cm}, clip]{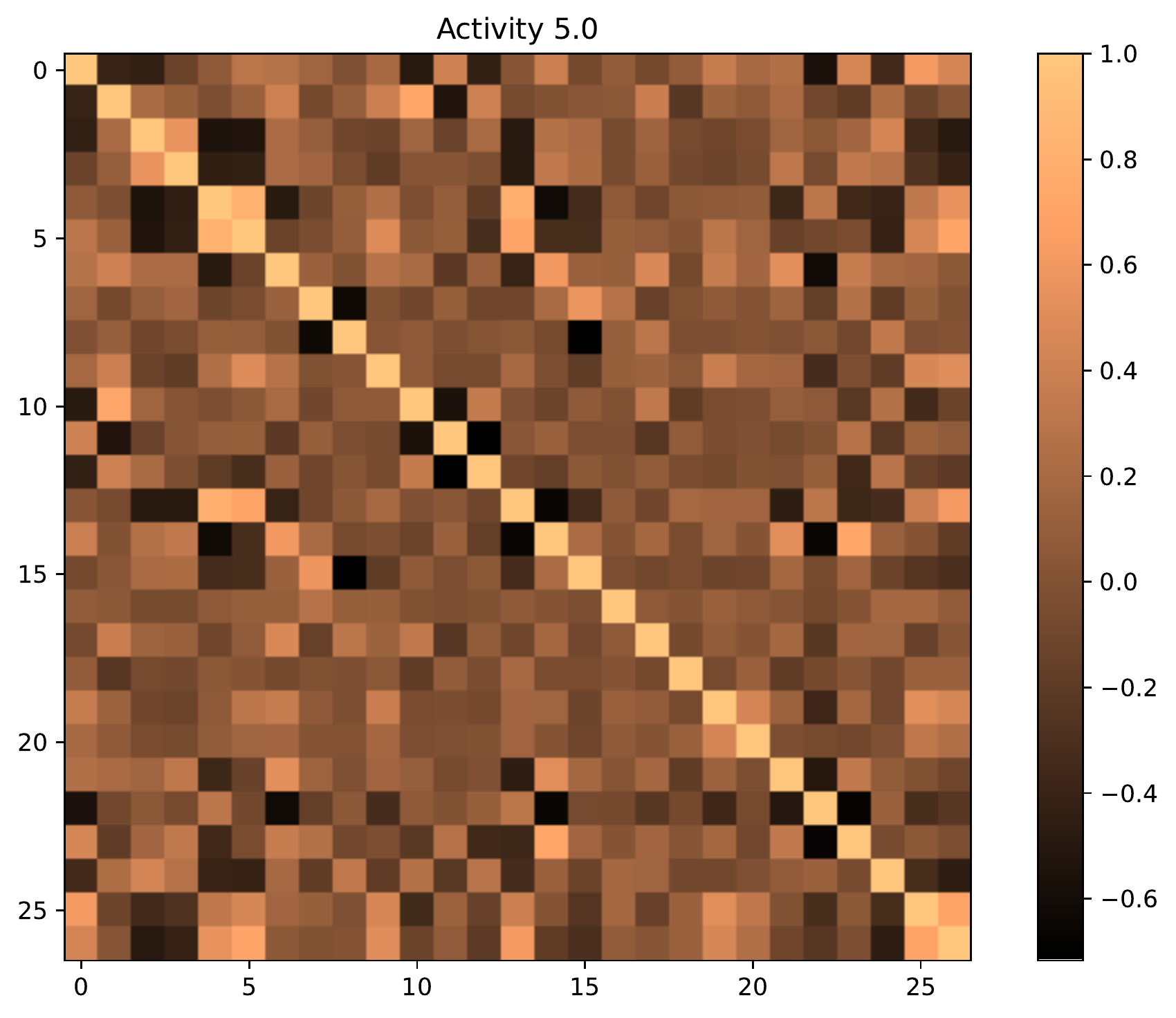}
    \caption{Activity 5}
    \end{subfigure}
    \hfill
        \begin{subfigure}[b]{0.3\textwidth}
         \centering
    \includegraphics[width=\linewidth, trim={0cm 0cm 3cm 0.8cm}, clip]{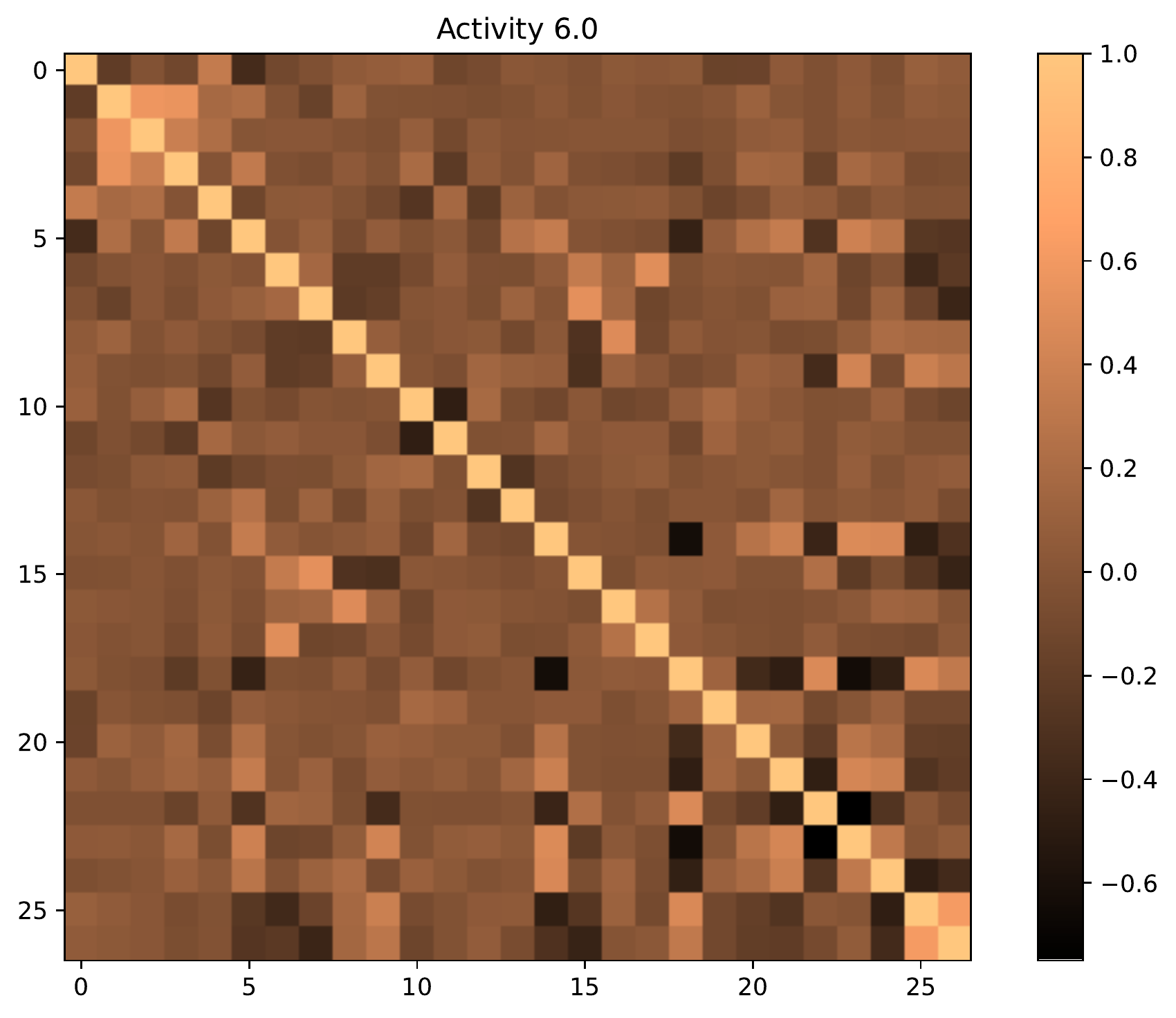}
    \caption{Activity 6}
    \end{subfigure}
    \hfill
    \begin{subfigure}[b]{0.3\textwidth}
         \centering
    \includegraphics[width=\linewidth, trim={0cm 0cm 3cm 0.8cm}, clip]{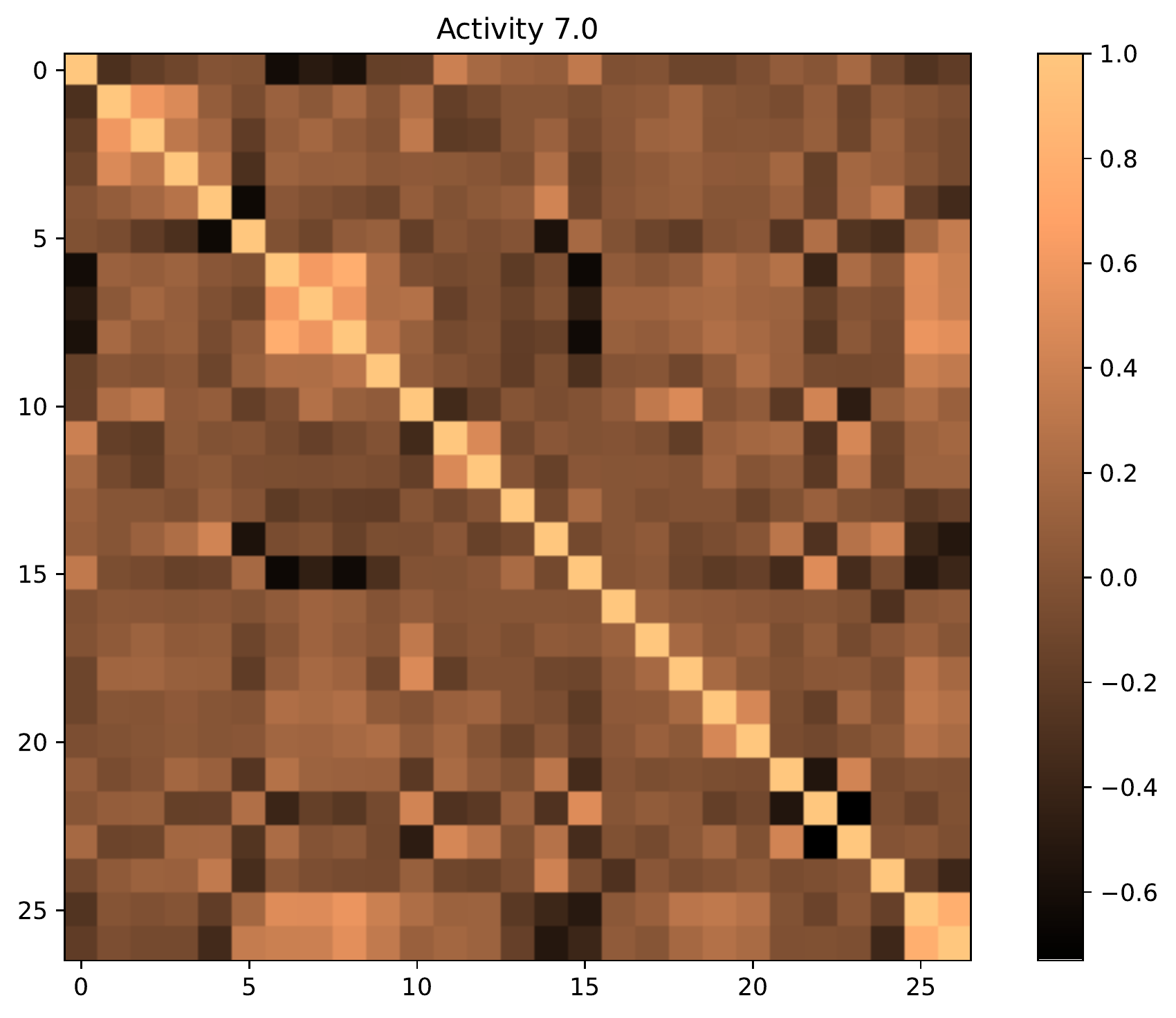}
    \caption{Activity 7}
    \end{subfigure}
    \hfill
    \begin{subfigure}[b]{0.3\textwidth}
         \centering
    \includegraphics[width=\linewidth, trim={0cm 0cm 3cm 0.8cm}, clip]{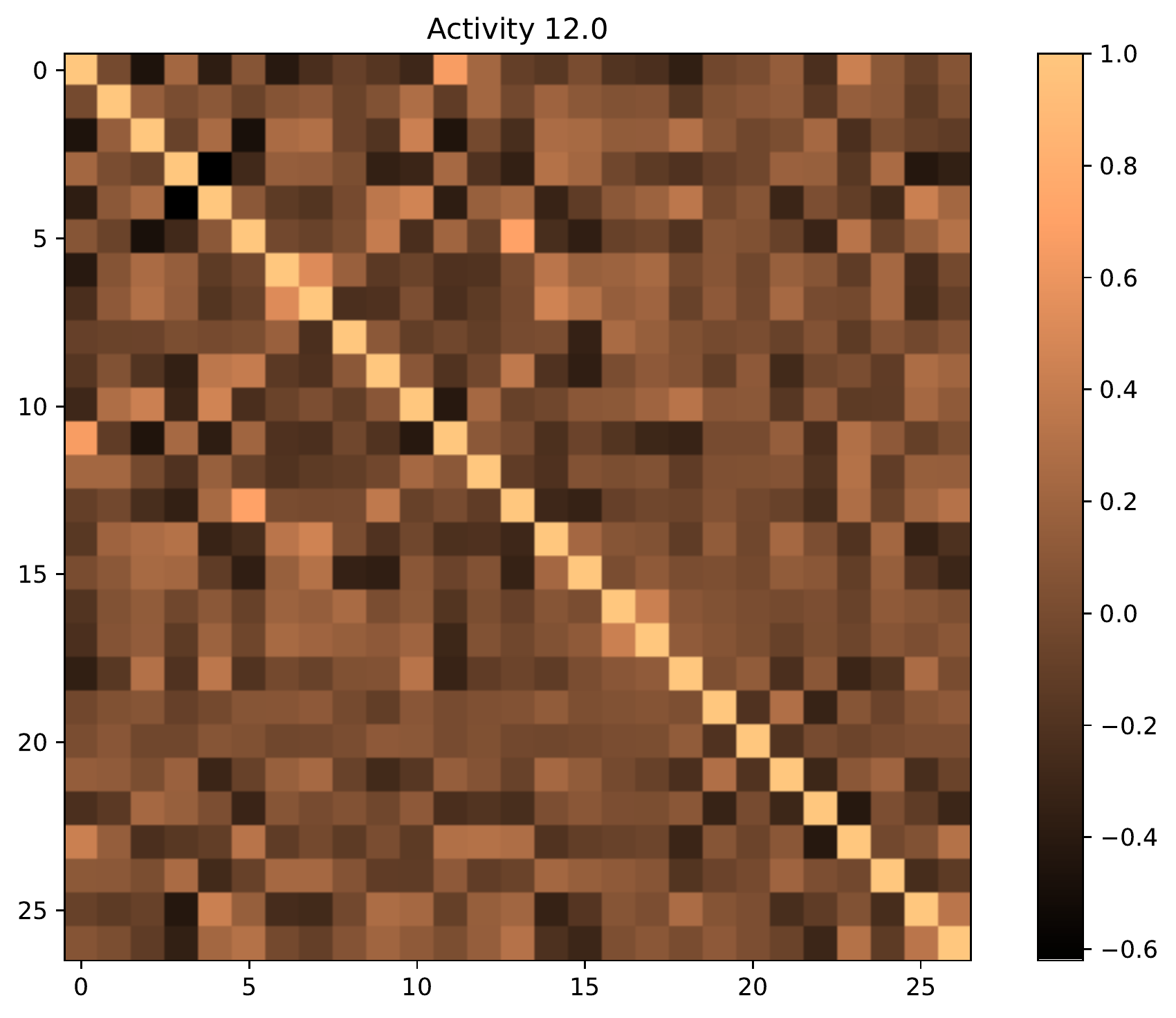}
    \caption{Activity 12}
    \end{subfigure}
    \hfill
            \begin{subfigure}[b]{0.3\textwidth}
         \centering
    \includegraphics[width=\linewidth, trim={0cm 0cm 3cm 0.8cm}, clip]{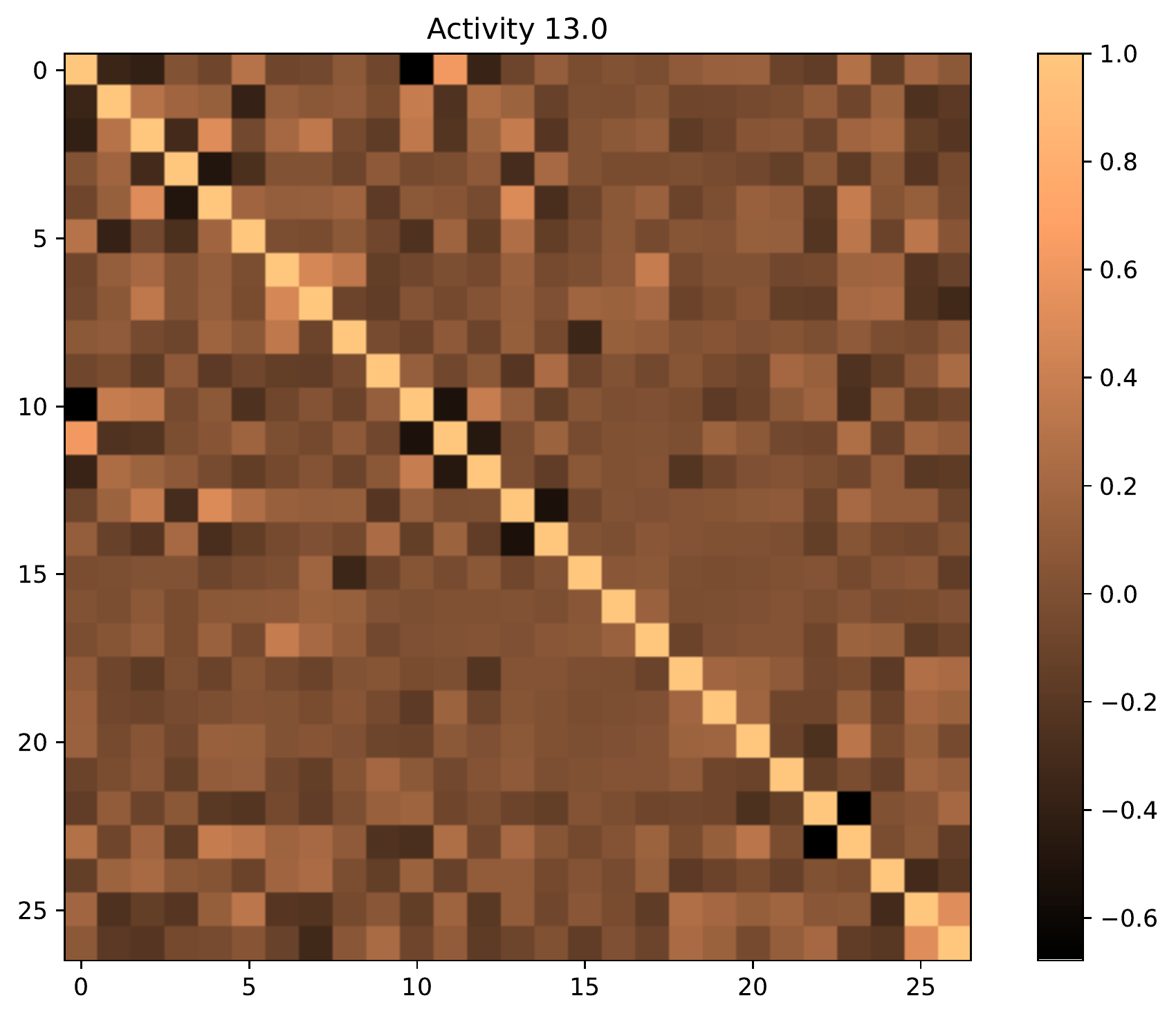}
    \caption{Activity 13}
    \end{subfigure}
    \hfill
    \begin{subfigure}[b]{0.3\textwidth}
         \centering
    \includegraphics[width=\linewidth, trim={1cm 0cm 4cm 0.8cm}, clip]{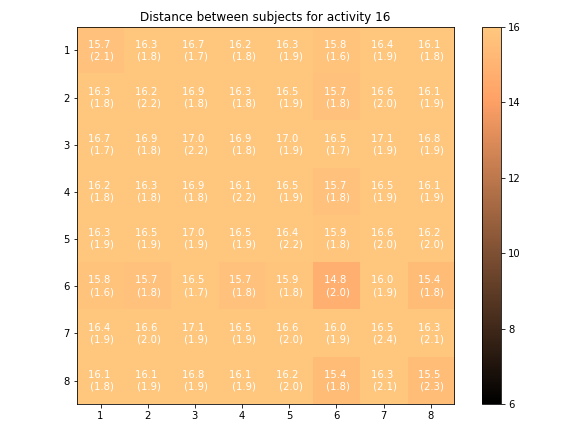}
     \caption{Activity 16}
    \end{subfigure}
    \hfill
    \begin{subfigure}[b]{0.3\textwidth}
         \centering
    \includegraphics[width=\linewidth, trim={0cm 0cm 3cm 0.8cm}, clip]{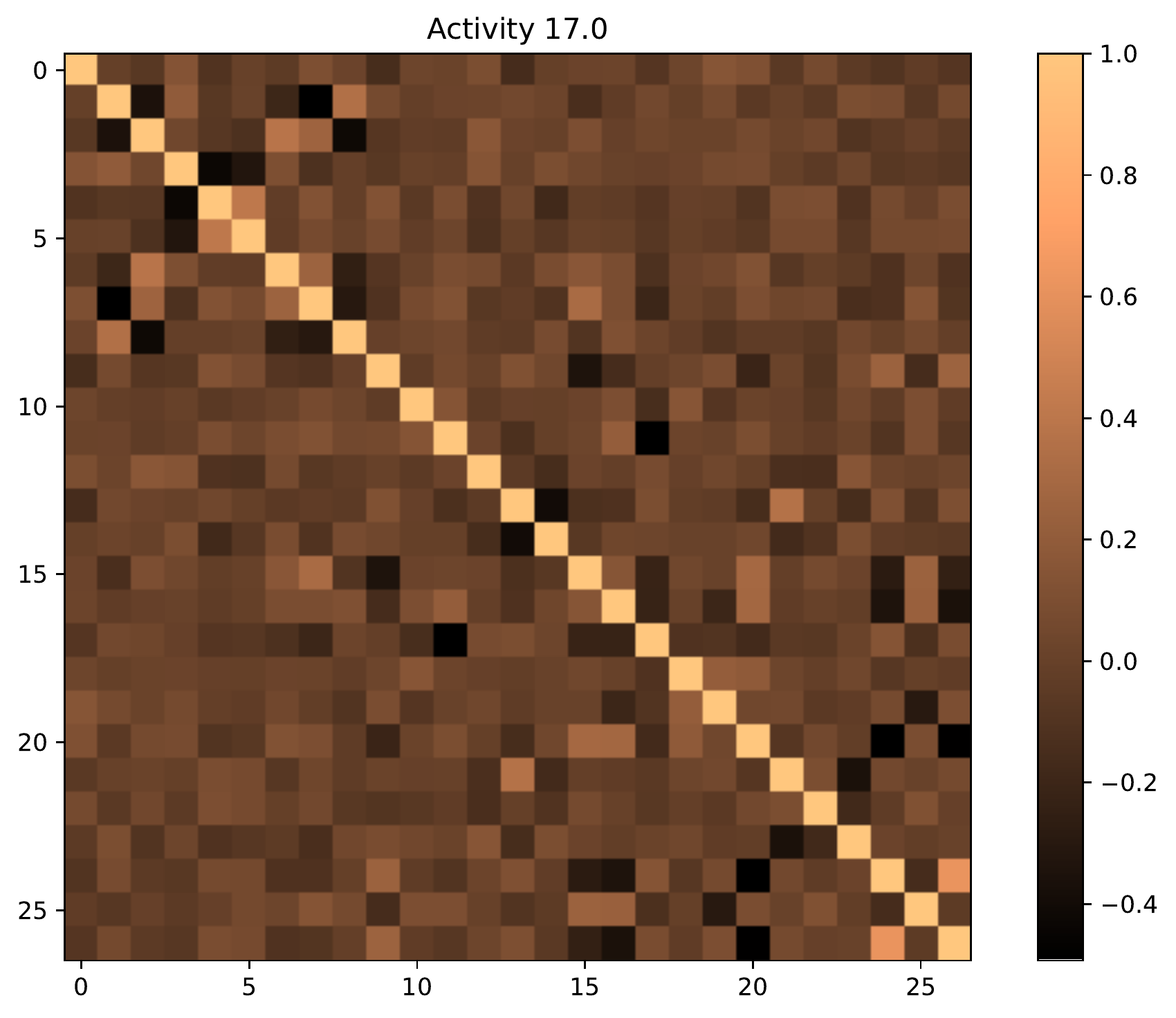}
     \caption{Activity 17}
    \end{subfigure}
    \hfill
    \begin{subfigure}[b]{0.3\textwidth}
         \centering
    \includegraphics[width=\linewidth, trim={0cm 0cm 3cm 0.8cm}, clip]{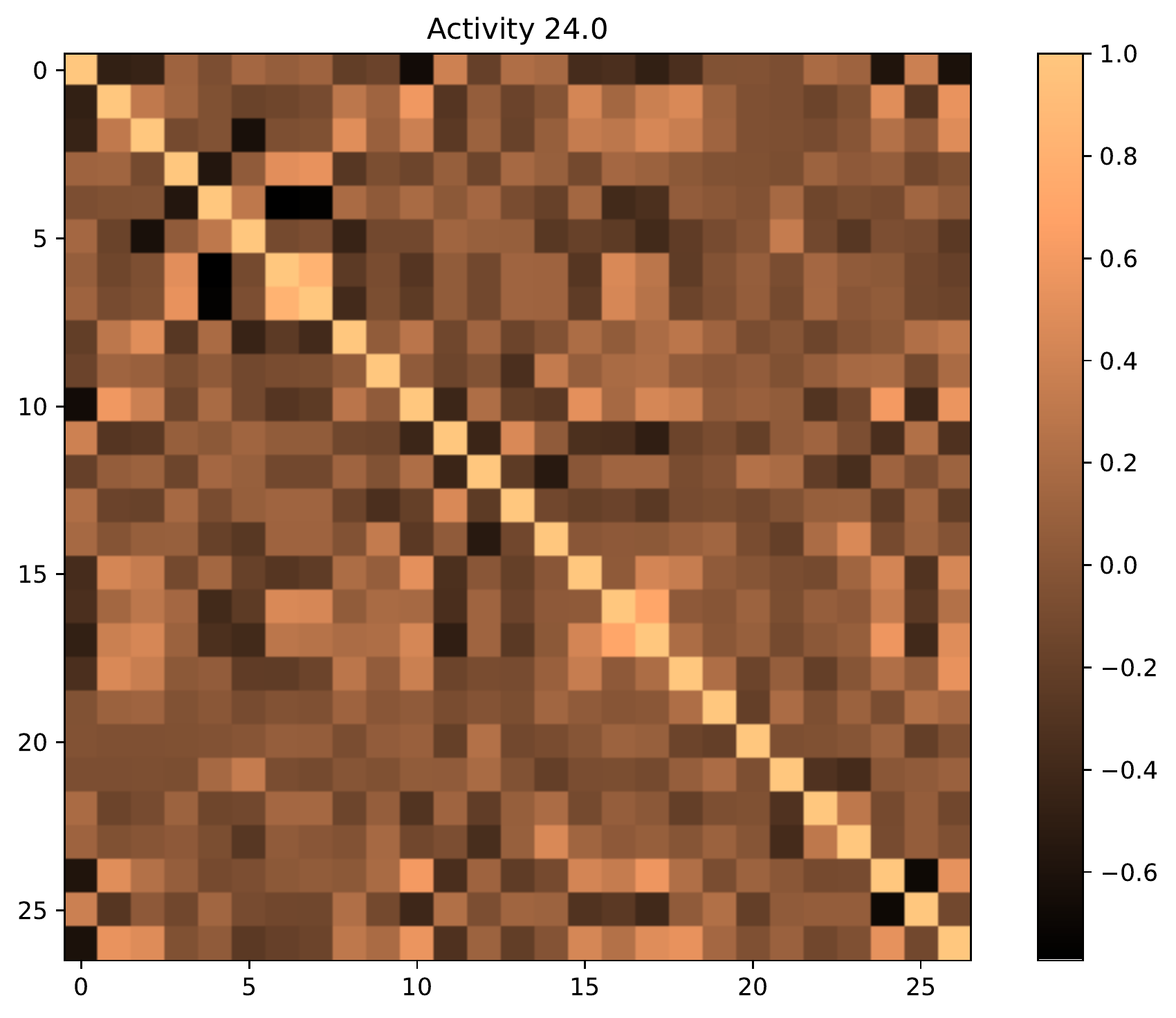}
    \caption{Activity 24}
    \end{subfigure}
    \hfill
 \caption{Average adjacency matrices for each of the 12 activities performed by Subject 1. }
   \label{fig:av_corr_0}
\end{figure}

\begin{figure}
    \centering
        \begin{subfigure}[b]{0.49\textwidth}
         \centering
    \includegraphics[width=\linewidth, trim={0cm 0cm 0cm 1cm}, clip]{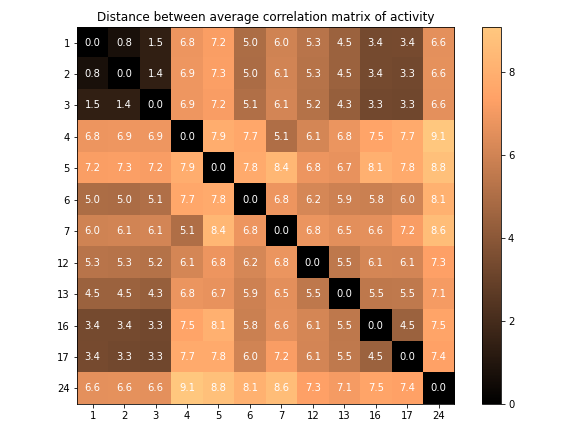}
    \caption{Distance between average adjacency matrices per activity.}
    \end{subfigure}
    \hfill
    \begin{subfigure}[b]{0.49\textwidth}
         \centering
    \includegraphics[width=\linewidth, trim={0cm 0cm 0cm 1cm}, clip]{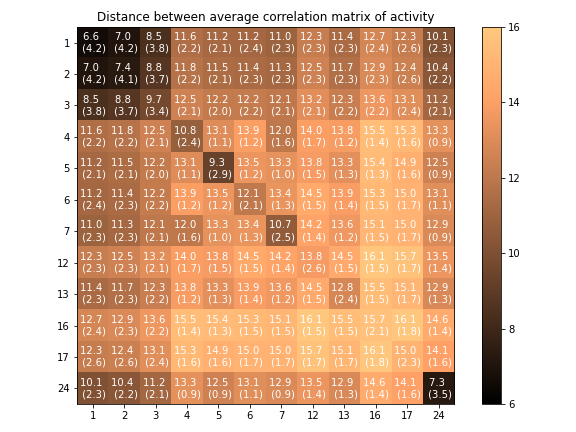}
    \caption{Average (and standard deviation) distance between graphs grouped by activities.}
    \end{subfigure}
    \hfill
    \caption{Frobenius distance between graphs in the dynamic network corresponding to Subject 1.} 
    \label{fig:av_dist_sub_0}
\end{figure}

\begin{figure}
    \centering
    \begin{subfigure}[b]{0.3\textwidth}
         \centering
    \includegraphics[width=\linewidth, trim={1cm 0cm 4cm 0.8cm}, clip]{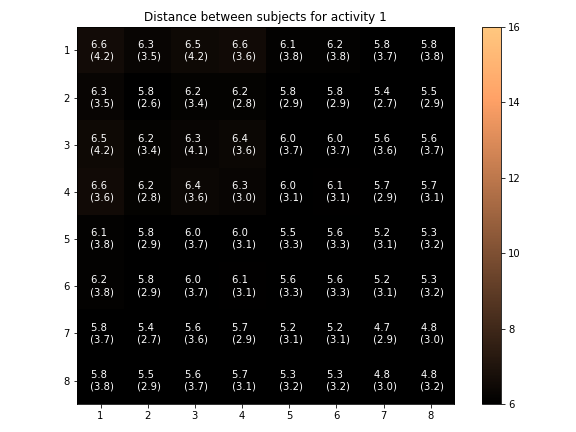}
    \caption{Activity 1}
    \end{subfigure}
    \hfill
    \begin{subfigure}[b]{0.3\textwidth}
         \centering
    \includegraphics[width=\linewidth, trim={1cm 0cm 4cm 0.8cm}, clip]{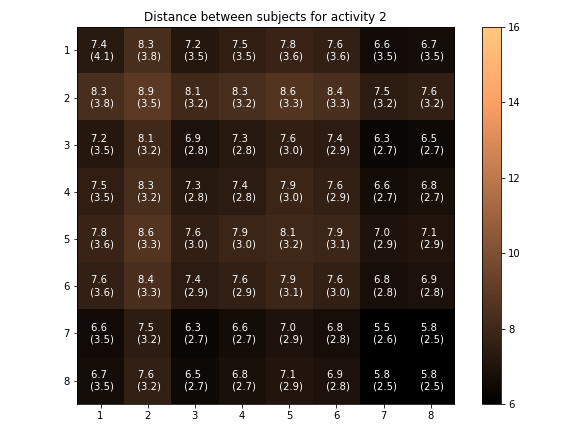}
    \caption{Activity 2}
    \end{subfigure}
    \hfill
    \begin{subfigure}[b]{0.34\textwidth}
         \centering
    \includegraphics[width=\linewidth, trim={1cm 0cm 2cm 0.8cm}, clip]{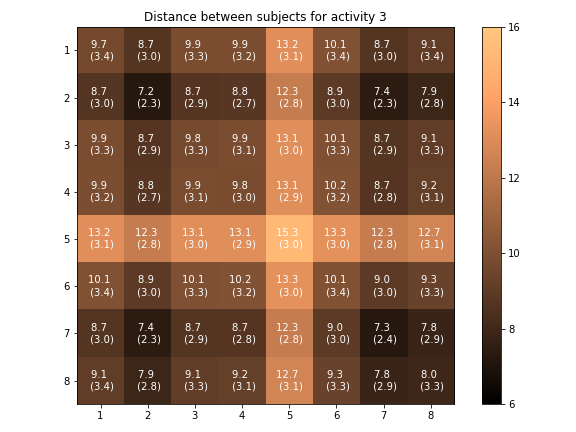}
    \caption{Activity 3}
    \end{subfigure}
    \hfill
        \begin{subfigure}[b]{0.3\textwidth}
         \centering
    \includegraphics[width=\linewidth, trim={1cm 0cm 4cm 0.8cm}, clip]{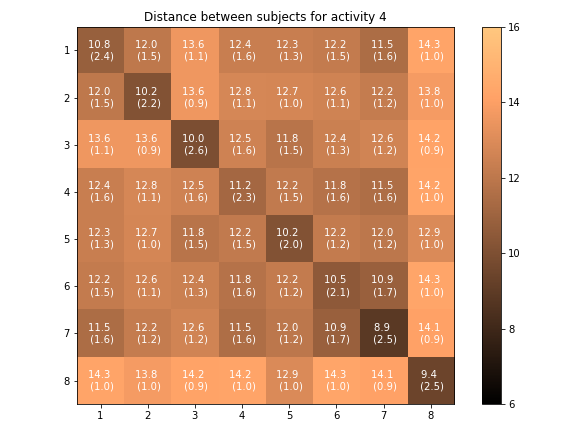}
    \caption{Activity 4}
    \end{subfigure}
    \hfill
        \begin{subfigure}[b]{0.3\textwidth}
         \centering
    \includegraphics[width=\linewidth, trim={1cm 0cm 4cm 0.8cm}, clip]{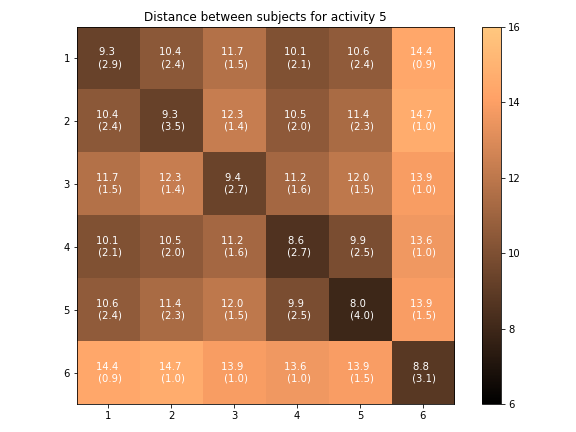}
    \caption{Activity 5}
    \end{subfigure}
    \hfill
        \begin{subfigure}[b]{0.3\textwidth}
         \centering
    \includegraphics[width=\linewidth, trim={1cm 0cm 4cm 0.8cm}, clip]{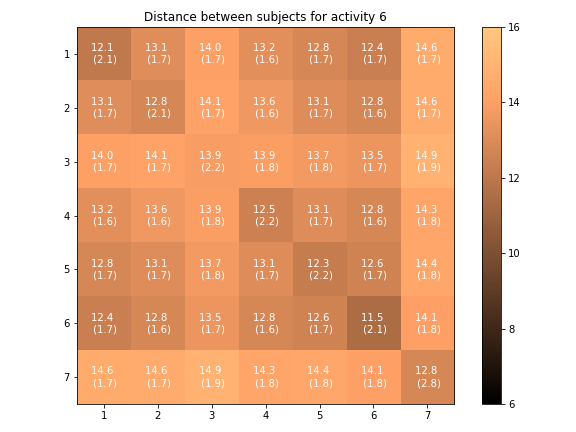}
    \caption{Activity 6}
    \end{subfigure}
    \hfill
    \begin{subfigure}[b]{0.3\textwidth}
         \centering
    \includegraphics[width=\linewidth, trim={1cm 0cm 4cm 0.8cm}, clip]{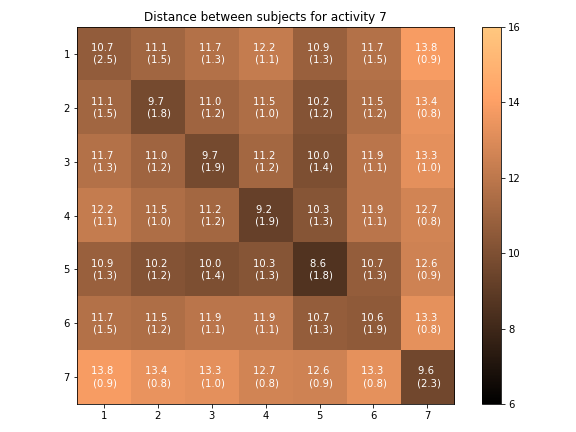}
    \caption{Activity 7}
    \end{subfigure}
    \hfill
    \begin{subfigure}[b]{0.3\textwidth}
         \centering
    \includegraphics[width=\linewidth, trim={1cm 0cm 4cm 0.8cm}, clip]{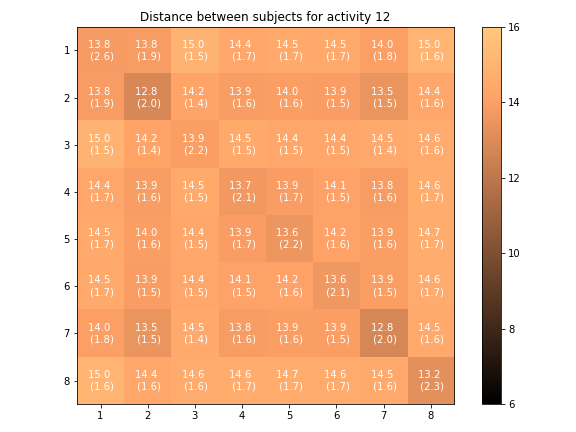}
    \caption{Activity 12}
    \end{subfigure}
    \hfill
            \begin{subfigure}[b]{0.3\textwidth}
         \centering
    \includegraphics[width=\linewidth, trim={1cm 0cm 4cm 0.8cm}, clip]{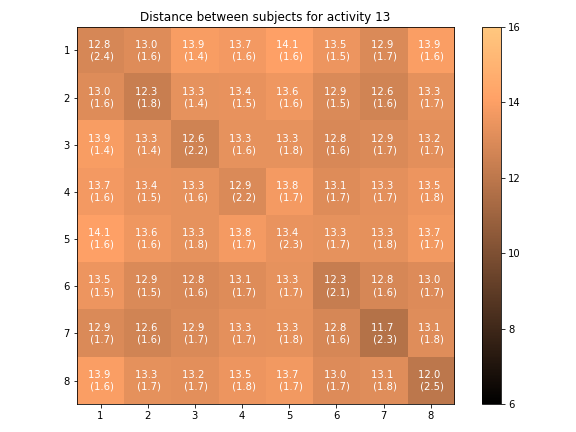}
    \caption{Activity 13}
    \end{subfigure}
    \hfill
    \begin{subfigure}[b]{0.3\textwidth}
         \centering
    \includegraphics[width=\linewidth, trim={1cm 0cm 4cm 0.8cm}, clip]{figures/phyact_dist_subjects_acti_16.png}
     \caption{Activity 16}
    \end{subfigure}
    \hfill
    \begin{subfigure}[b]{0.3\textwidth}
         \centering
    \includegraphics[width=\linewidth, trim={1cm 0cm 4cm 0.8cm}, clip]{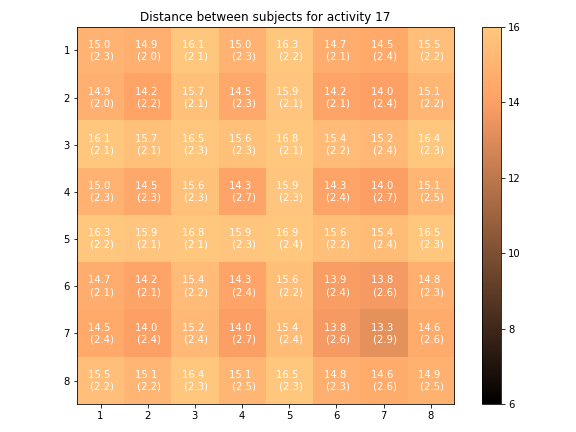}
     \caption{Activity 17}
    \end{subfigure}
    \hfill
    \begin{subfigure}[b]{0.3\textwidth}
         \centering
    \includegraphics[width=\linewidth, trim={1cm 0cm 4cm 0.8cm}, clip]{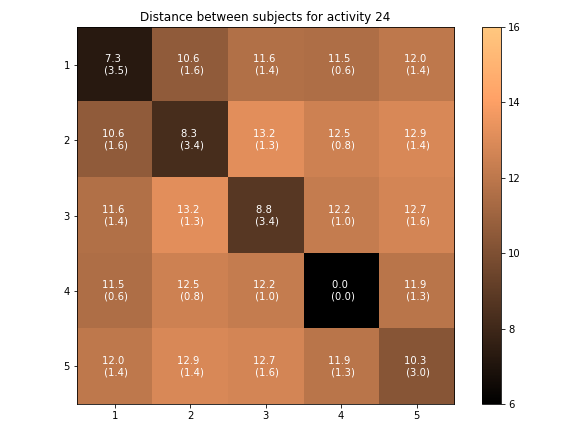}
    \caption{Activity 24}
    \end{subfigure}
    \hfill
    \caption{Average (and standard deviation) Frobenius distance between the graphs grouped by subjects for each activity label.}
    \label{fig:dist_act}
\end{figure}

\begin{figure}
    \centering
    \includegraphics[width=0.4\linewidth, trim={1cm 0cm 1cm 0.9cm}, clip]{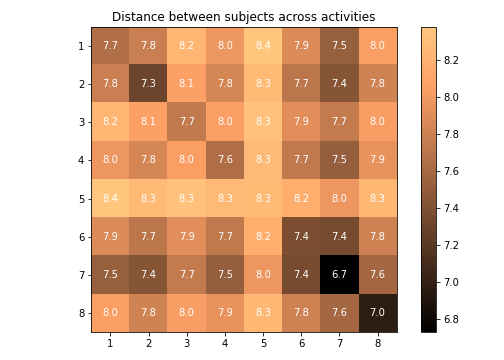}
    \caption{Average Frobenius distance between graphs with the same label grouped by subjects.}
    \label{fig:av_dist_subj}
\end{figure}

Additional results on the similarity between activities and subjects are presented in Figure~\ref{fig:av_corr_0}, Figure~\ref{fig:av_dist_sub_0}, Figure~\ref{fig:dist_act}, and Figure~\ref{fig:av_dist_subj}.

\subsection{ONCPD statistic}
\begin{figure}
    \centering
    \begin{subfigure}[b]{0.3\textwidth}
         \centering
         \includegraphics[width=\textwidth]{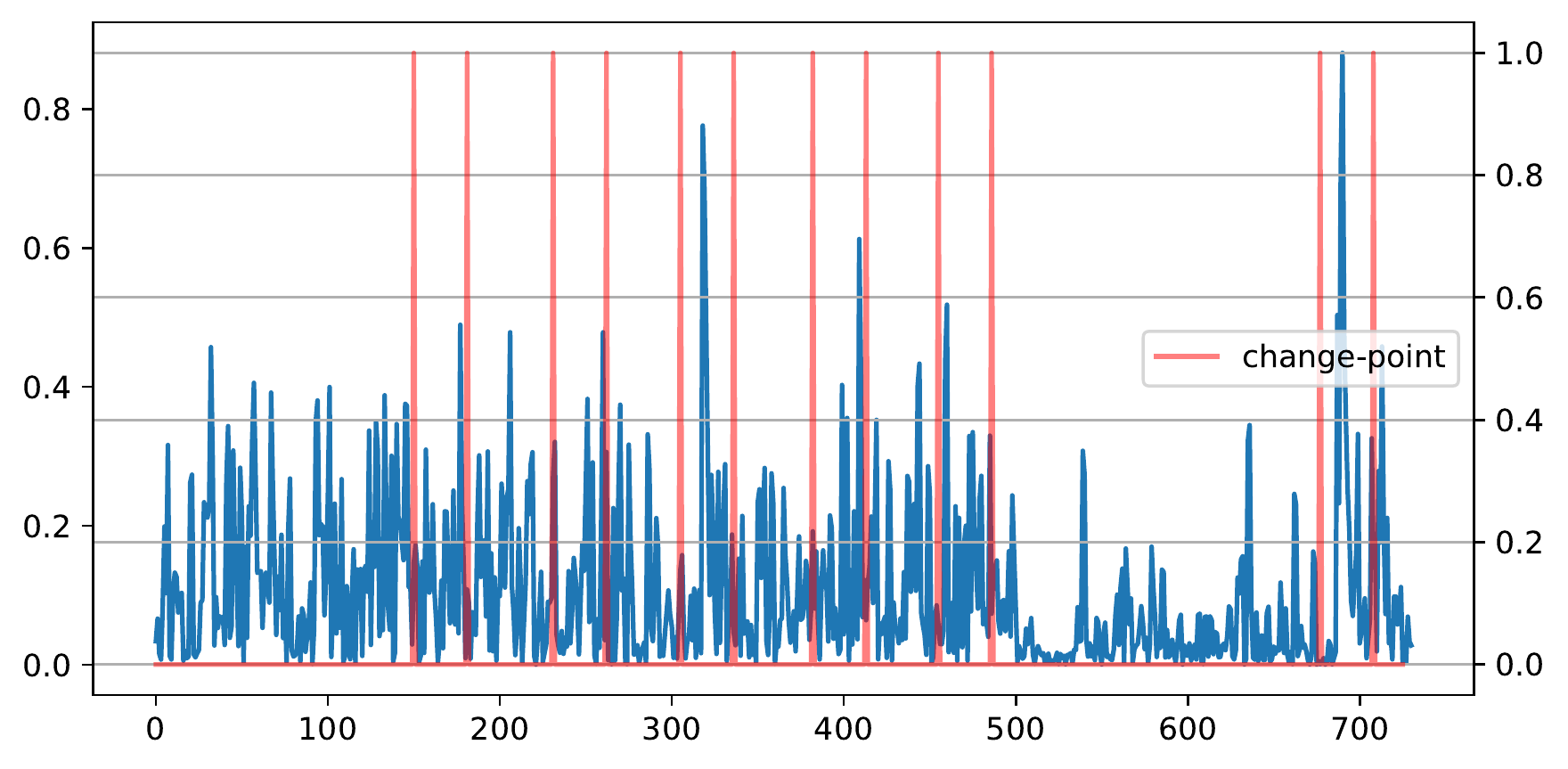}
         \caption{Subject 1}
         \label{}
     \end{subfigure}
     \hfill
         \begin{subfigure}[b]{0.3\textwidth}
         \centering
         \includegraphics[width=\textwidth]{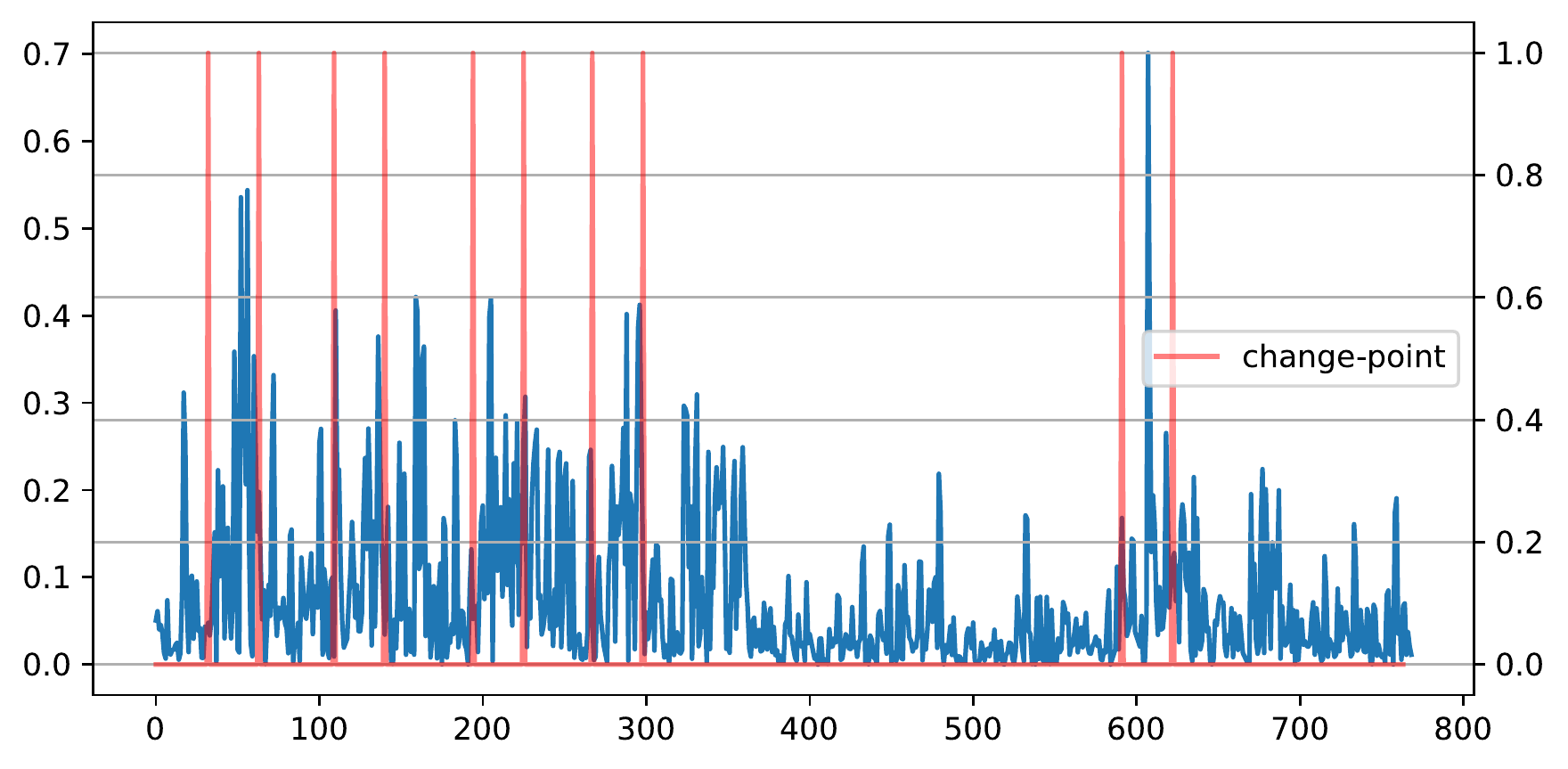}
         \caption{Subject 2}
         \label{}
     \end{subfigure}
     \hfill
     \begin{subfigure}[b]{0.3\textwidth}
         \centering
         \includegraphics[width=\textwidth]{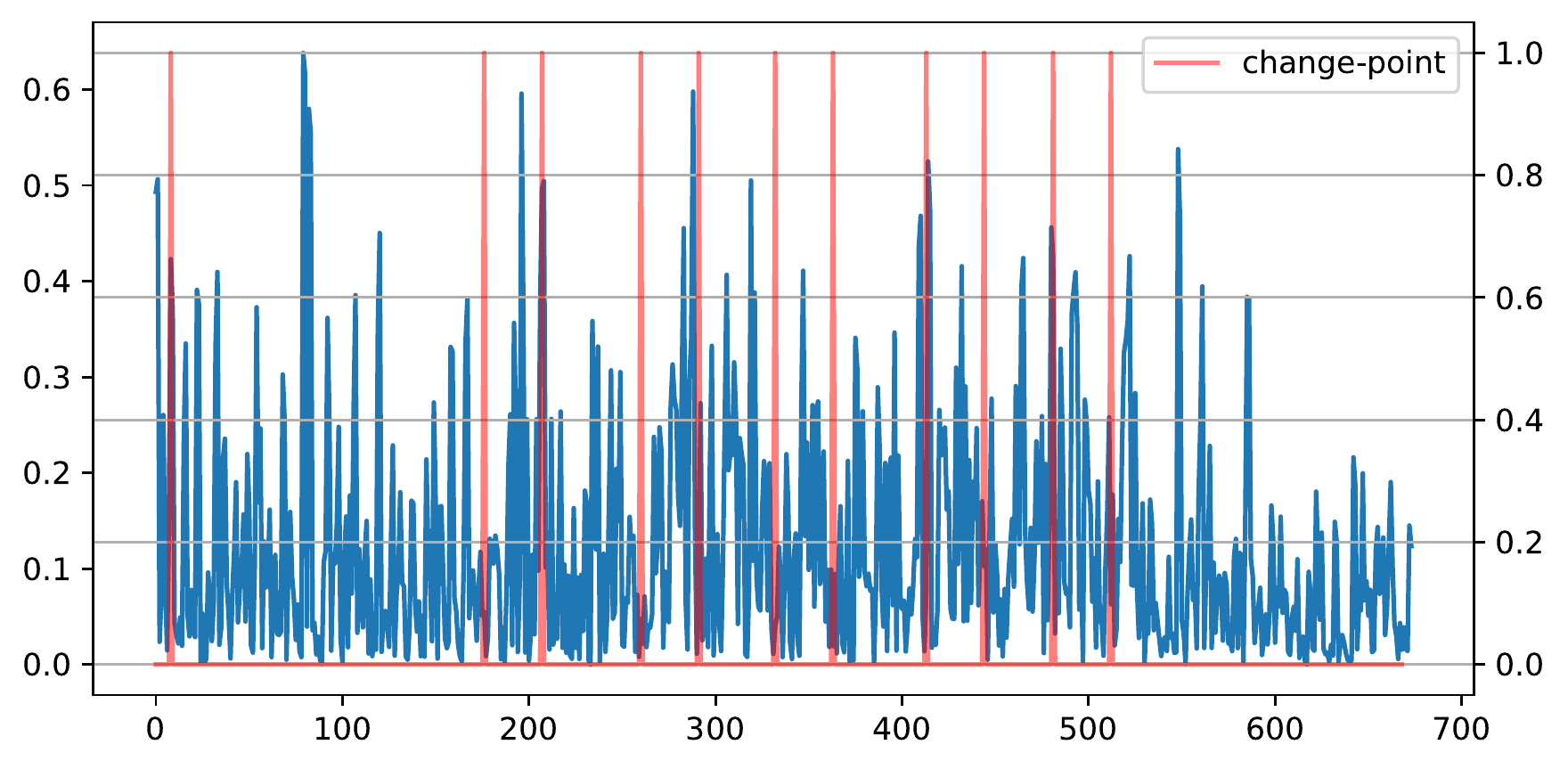}
         \caption{Subject 3}
         \label{}
     \end{subfigure}
     \hfill
    \begin{subfigure}[b]{0.3\textwidth}
         \centering
         \includegraphics[width=\textwidth]{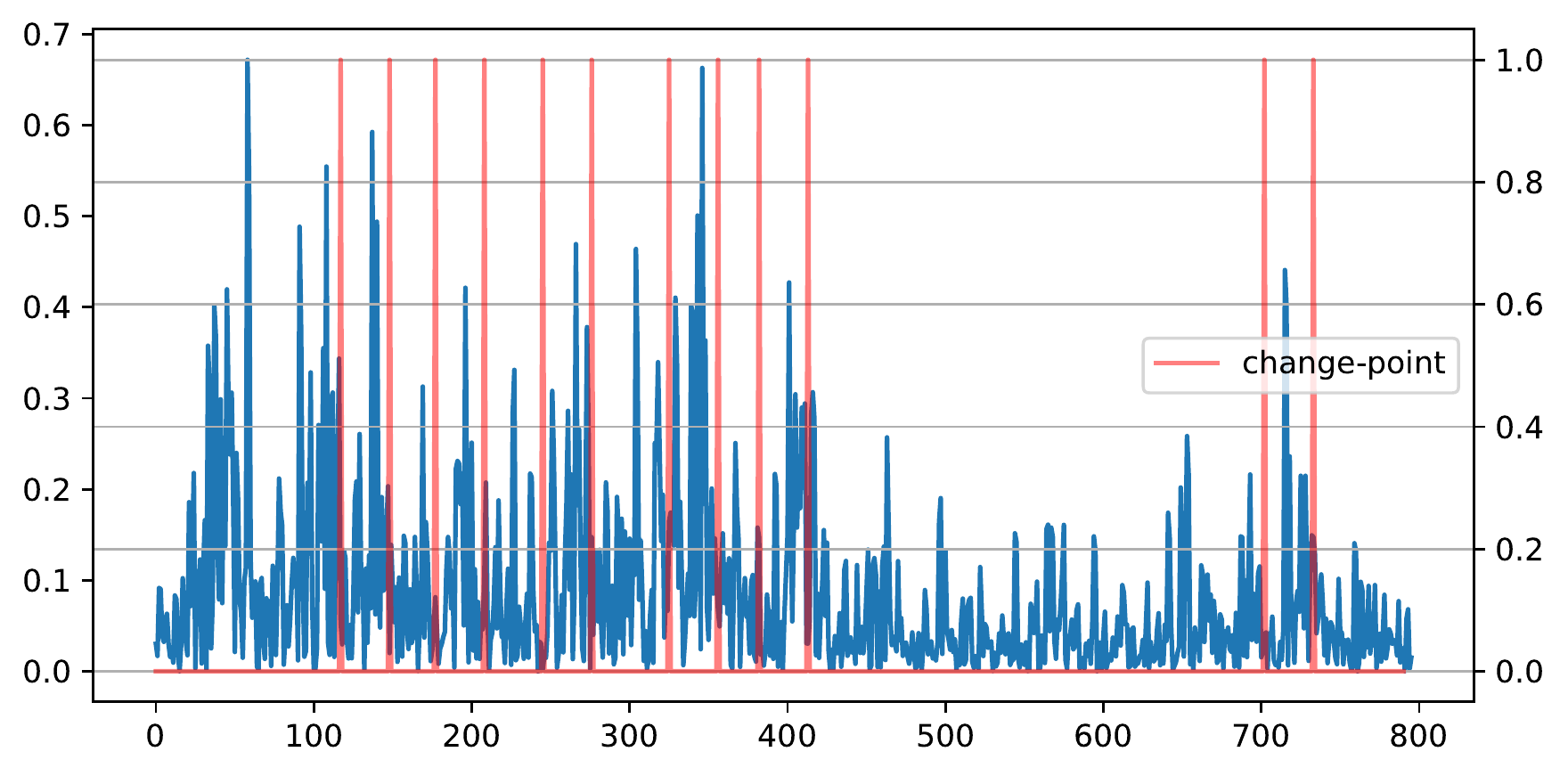}
         \caption{Subject 4}
         \label{}
     \end{subfigure}
     \hfill
     \begin{subfigure}[b]{0.3\textwidth}
         \centering
         \includegraphics[width=\textwidth]{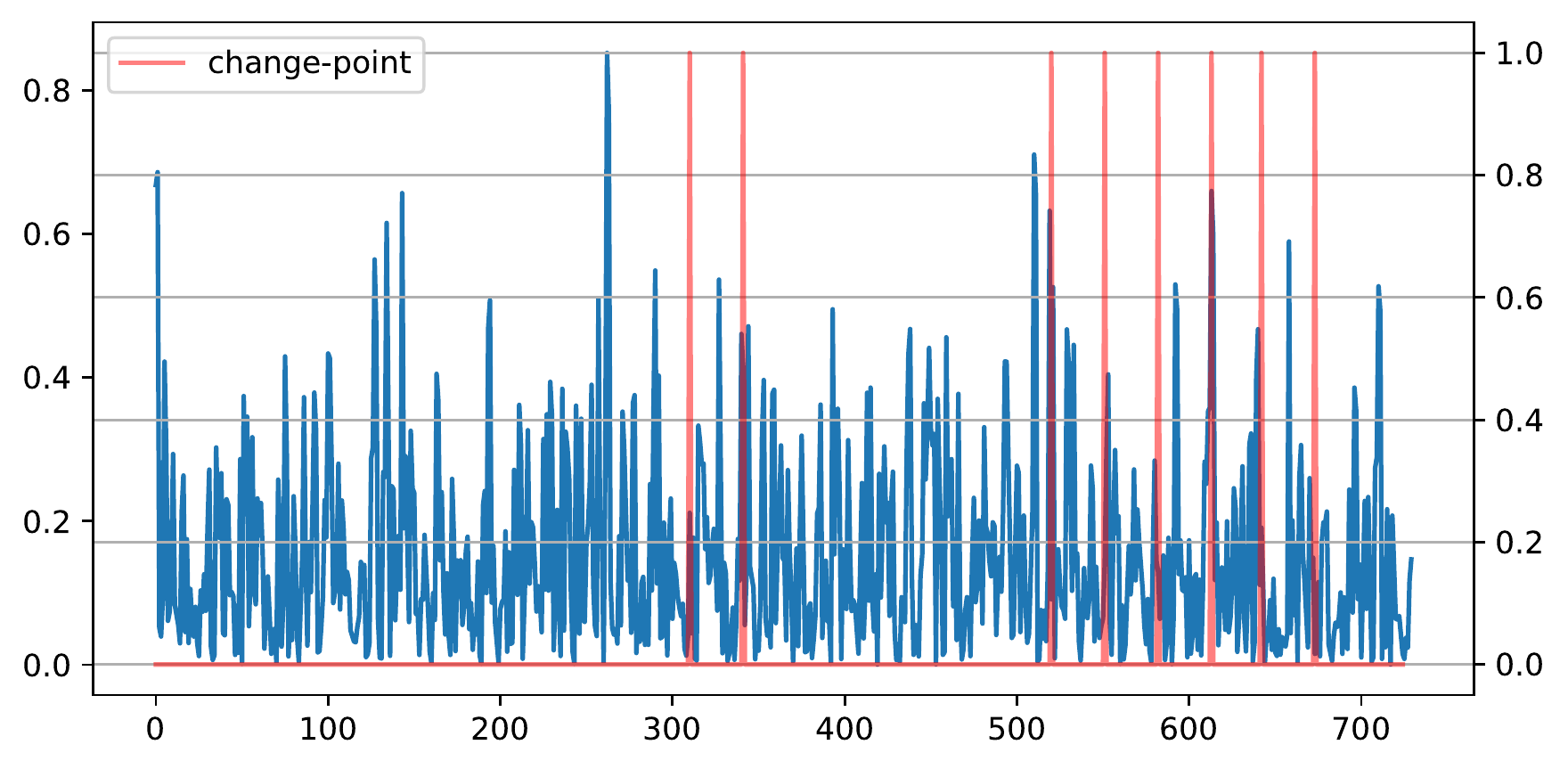}
         \caption{Subject 5}
         \label{}
     \end{subfigure}
     \hfill
     \begin{subfigure}[b]{0.3\textwidth}
         \centering
         \includegraphics[width=\textwidth]{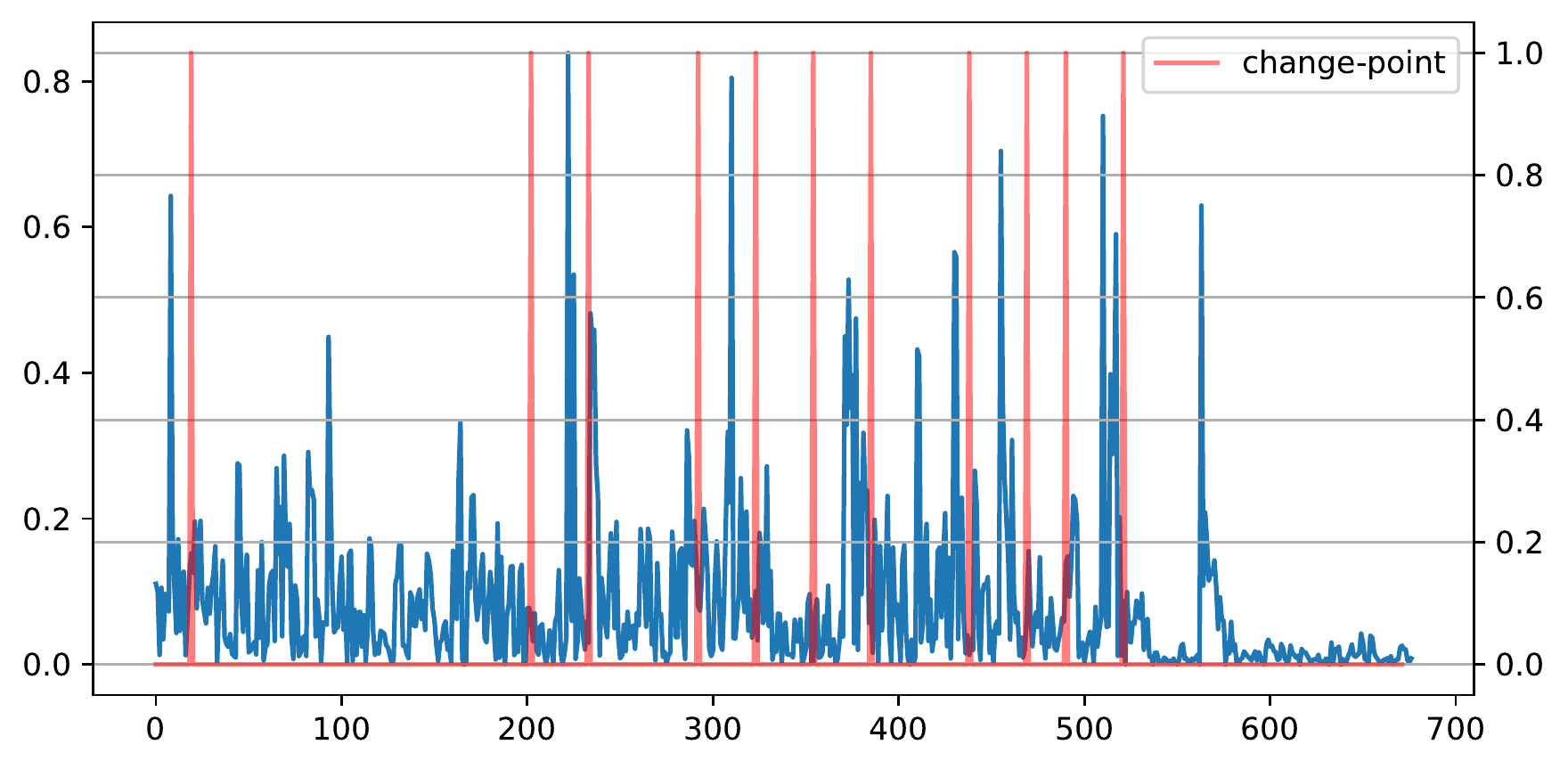}
         \caption{Subject 6}
         \label{}
     \end{subfigure}
     \hfill
     \caption{Change point detection statistic of our sGNN method and activity change points (in red) with a tolerance of $\pm 5$ timestamps for Subjects 1-6 a window size $L = 10$.}
    \label{fig:res_exp1_stat_gnn}
\end{figure}

\begin{figure}
    \centering
    \begin{subfigure}[b]{0.3\textwidth}
         \centering
         \includegraphics[width=\textwidth]{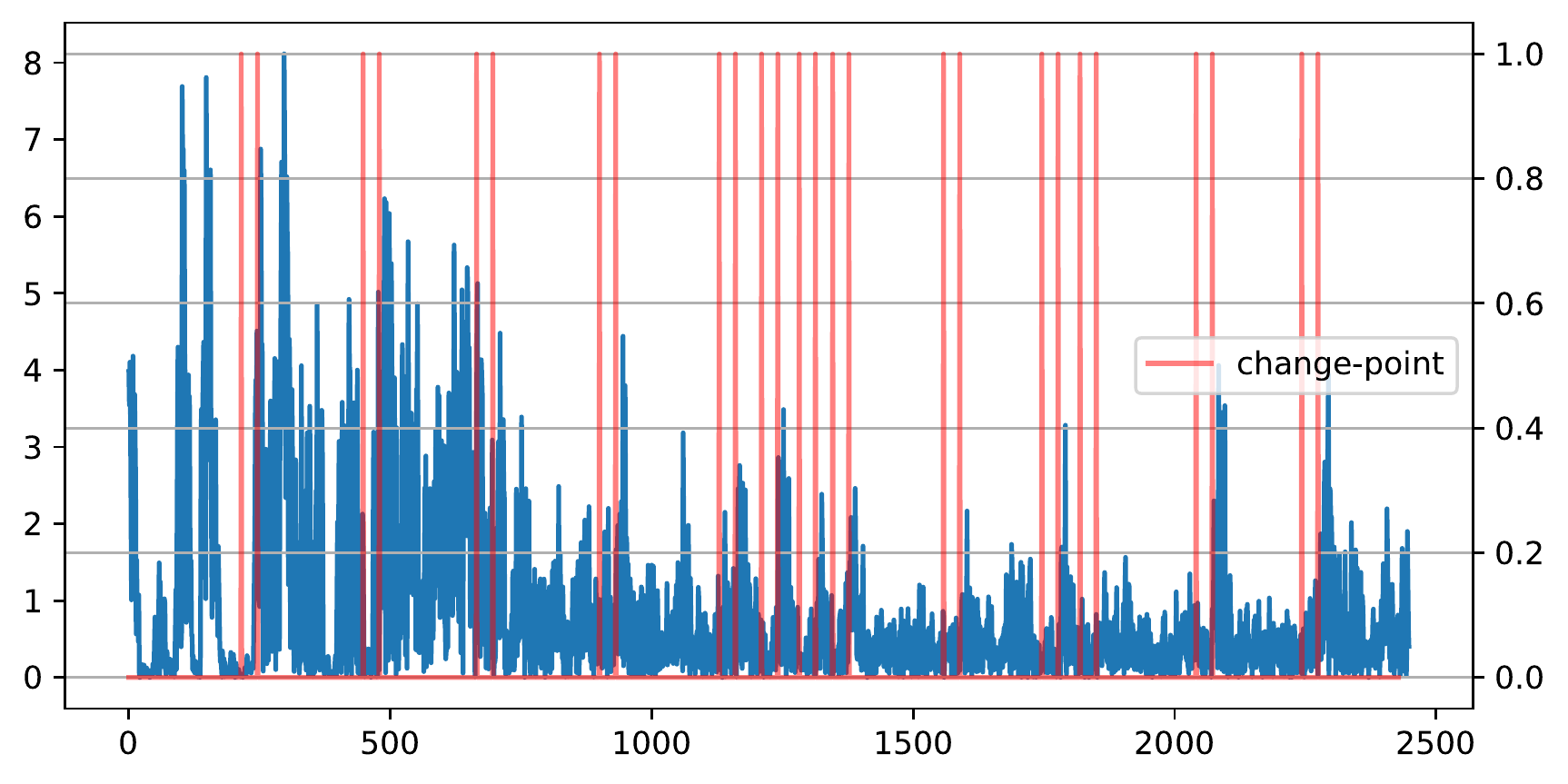}
         \caption{Subject 1}
         \label{}
     \end{subfigure}
     \hfill
         \begin{subfigure}[b]{0.3\textwidth}
         \centering
         \includegraphics[width=\textwidth]{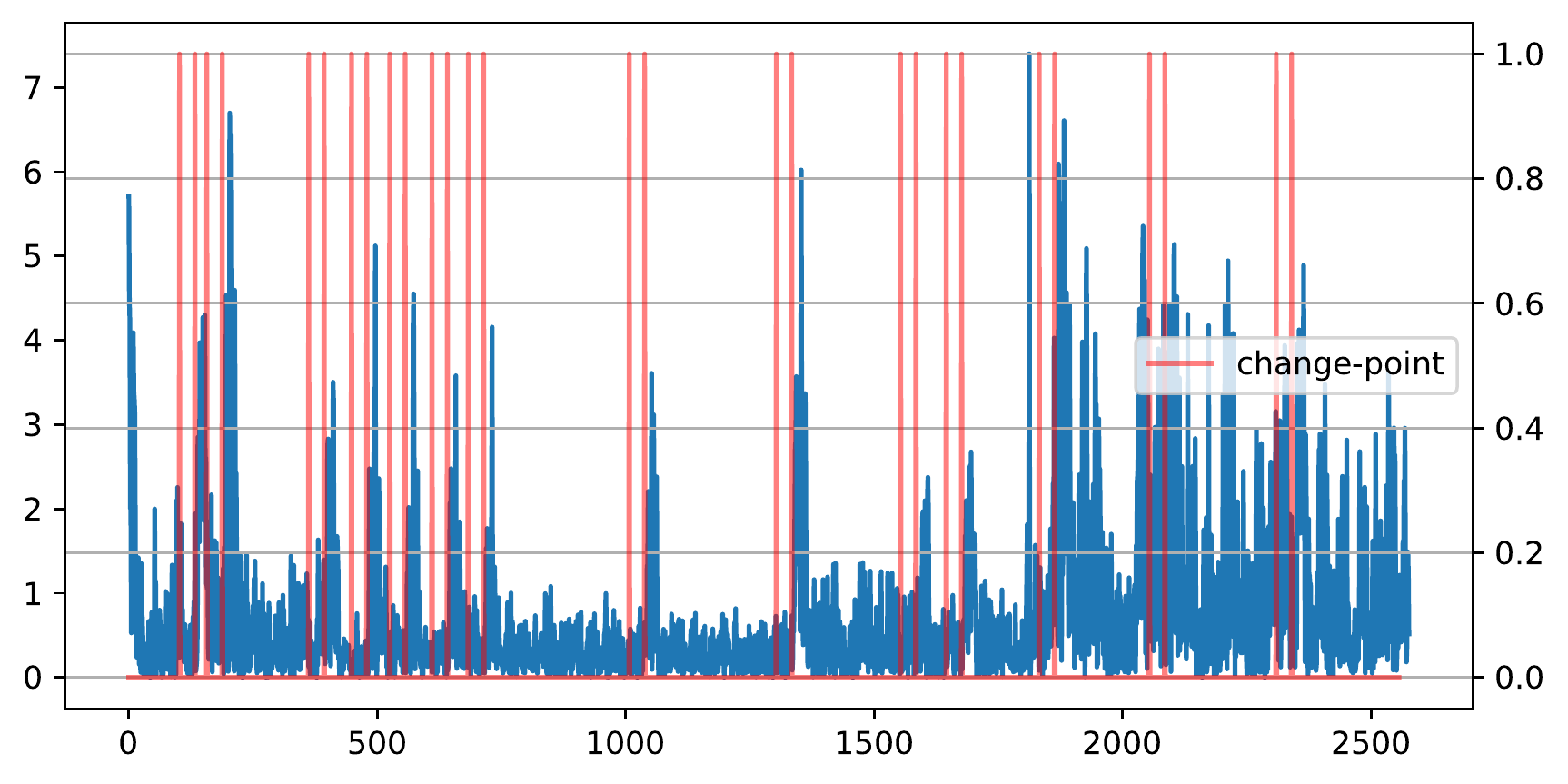}
         \caption{Subject 2}
         \label{}
     \end{subfigure}
     \hfill
     \begin{subfigure}[b]{0.3\textwidth}
         \centering
         \includegraphics[width=\textwidth]{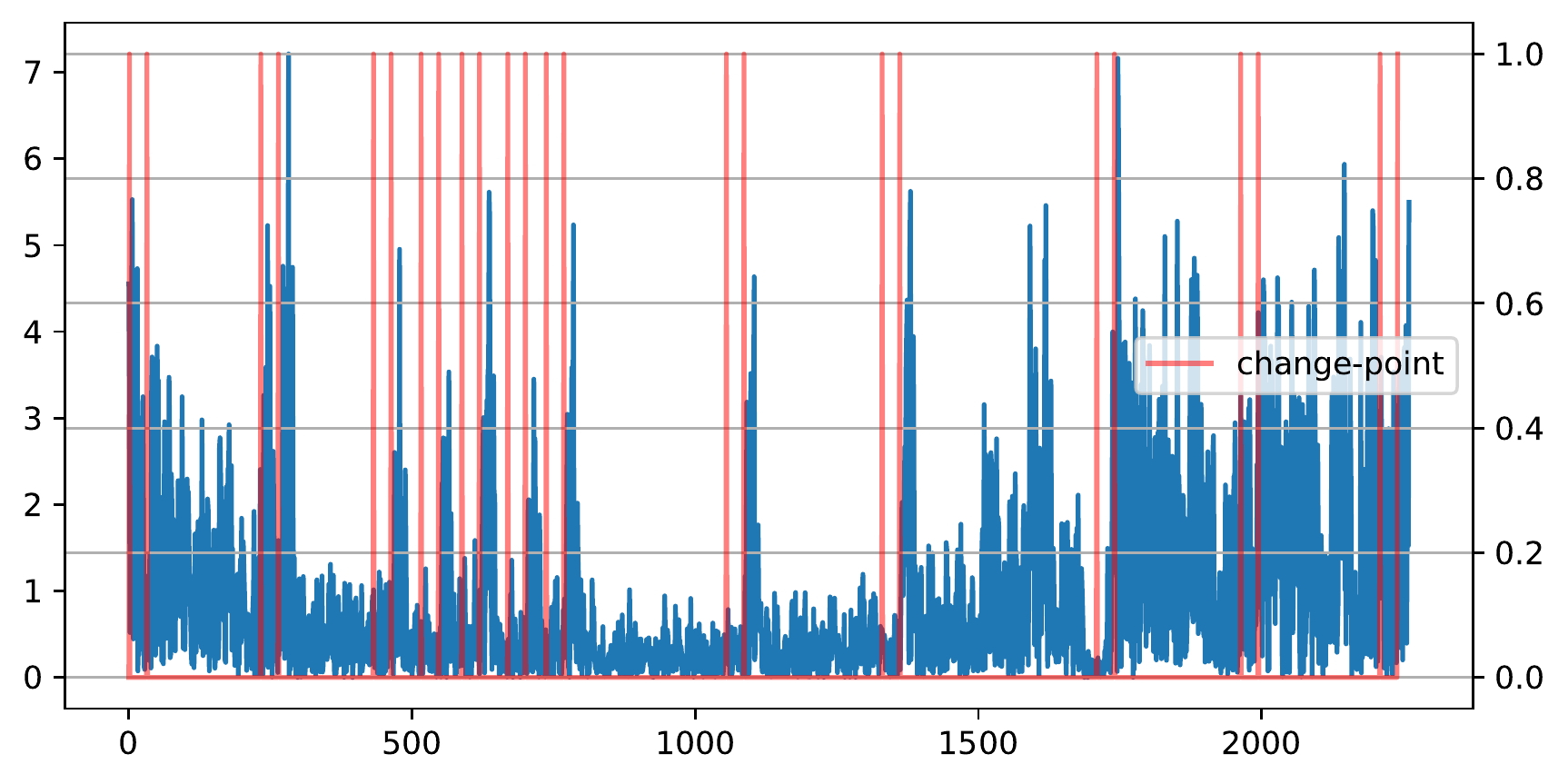}
         \caption{Subject 3}
         \label{}
     \end{subfigure}
     \hfill
    \begin{subfigure}[b]{0.3\textwidth}
         \centering
         \includegraphics[width=\textwidth]{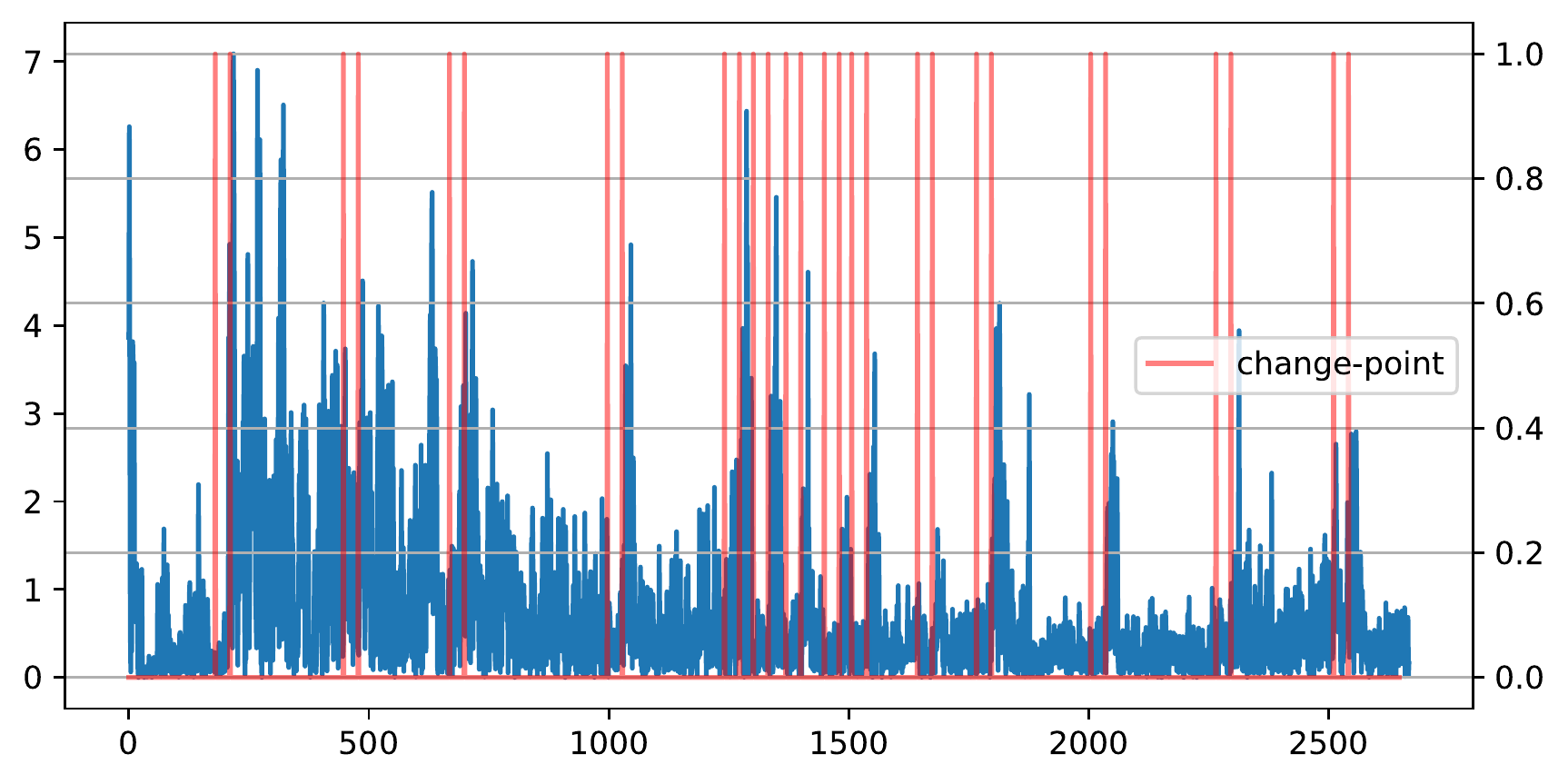}
         \caption{Subject 4}
         \label{}
     \end{subfigure}
     \hfill
     \begin{subfigure}[b]{0.3\textwidth}
         \centering
         \includegraphics[width=\textwidth]{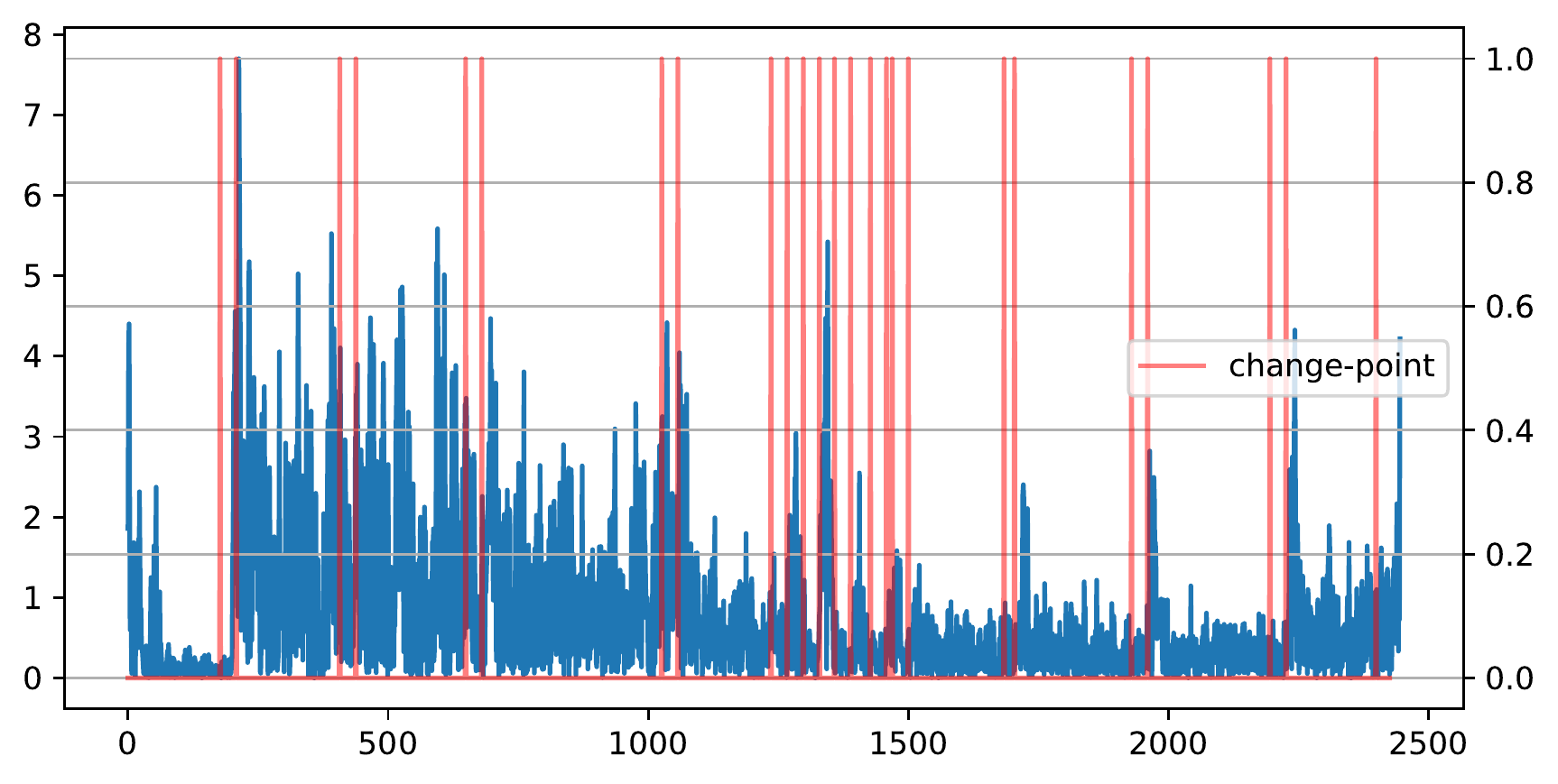}
         \caption{Subject 5}
         \label{}
     \end{subfigure}
     \hfill
     \begin{subfigure}[b]{0.3\textwidth}
         \centering
         \includegraphics[width=\textwidth]{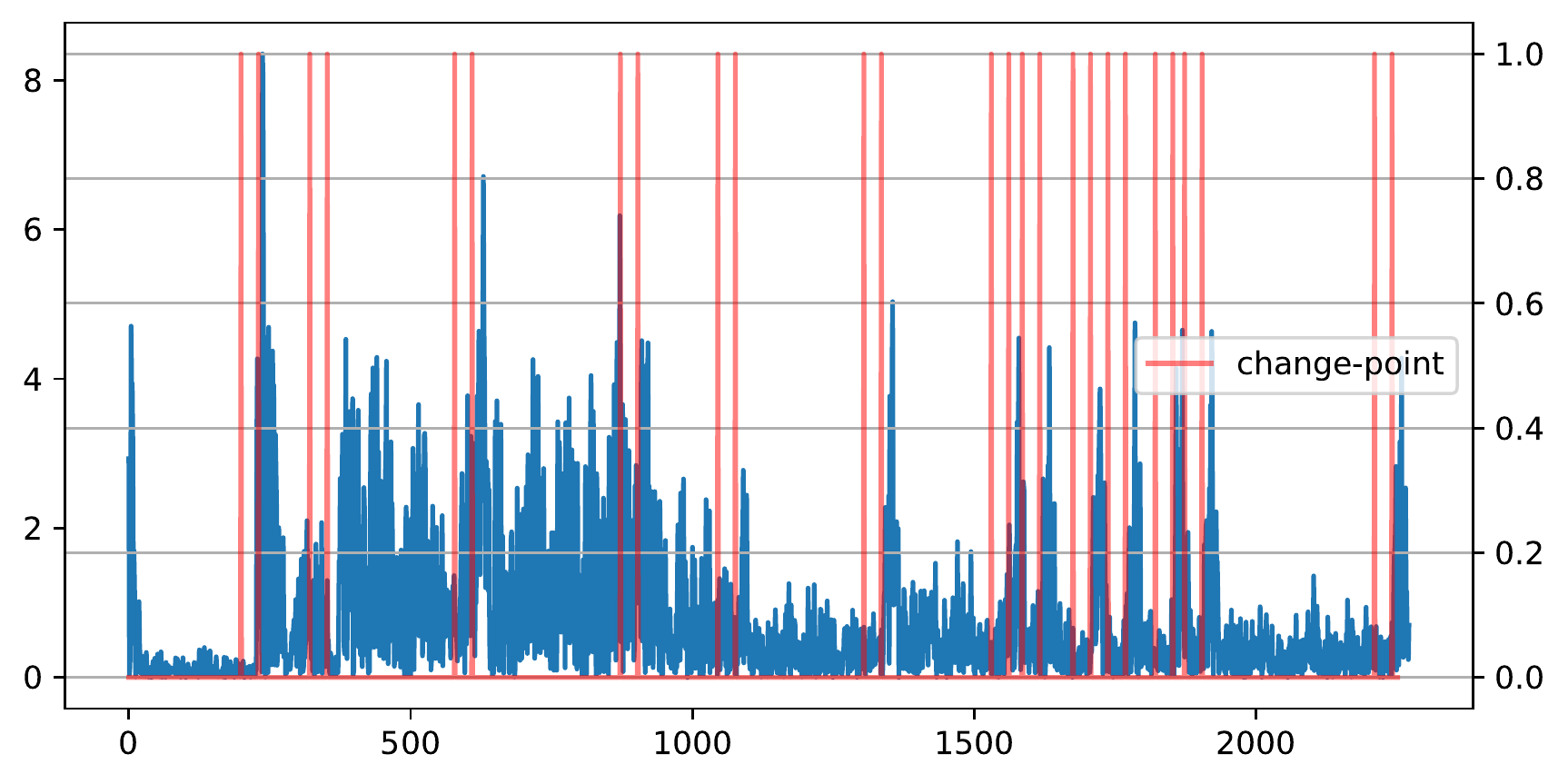}
         \caption{Subject 6}
         \label{}
     \end{subfigure}
     \hfill
     \caption{CUSUM 2 statistic and activity change points (in red) with a tolerance of $\pm 5$ timestamps for Subjects 1-6 a window size $L = 20$.}
    \label{fig:res_exp1_stat_cusum}
\end{figure}

\begin{figure}
    \centering
    \begin{subfigure}[b]{0.3\textwidth}
         \centering
         \includegraphics[width=\textwidth]{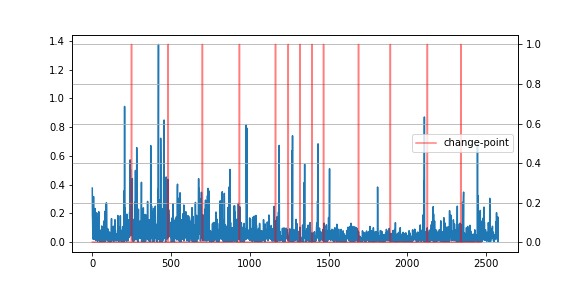}
         \caption{Subject 1}
         \label{}
     \end{subfigure}
     \hfill
         \begin{subfigure}[b]{0.3\textwidth}
         \centering
         \includegraphics[width=\textwidth]{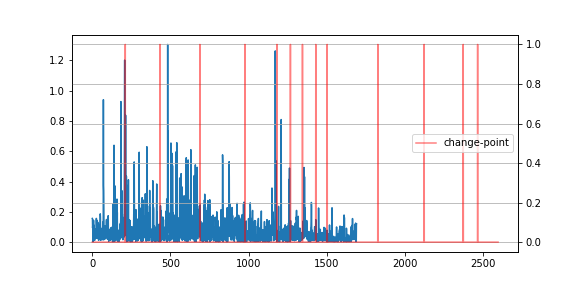}
         \caption{Subject 2}
         \label{}
     \end{subfigure}
     \hfill
     \begin{subfigure}[b]{0.3\textwidth}
         \centering
         \includegraphics[width=\textwidth]{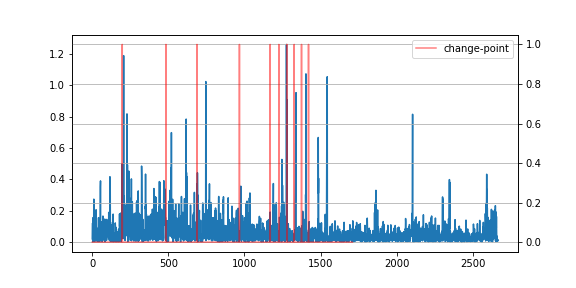}
         \caption{Subject 3}
         \label{}
     \end{subfigure}
     \hfill
    \begin{subfigure}[b]{0.3\textwidth}
         \centering
         \includegraphics[width=\textwidth]{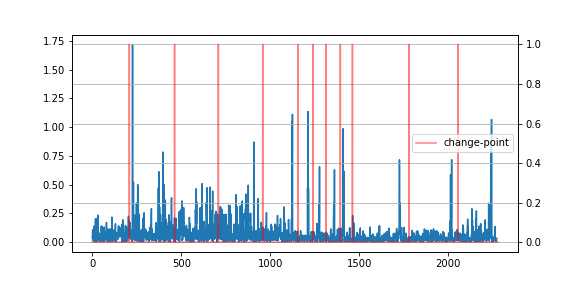}
         \caption{Subject 4}
         \label{}
     \end{subfigure}
     \hfill
     \begin{subfigure}[b]{0.3\textwidth}
         \centering
         \includegraphics[width=\textwidth]{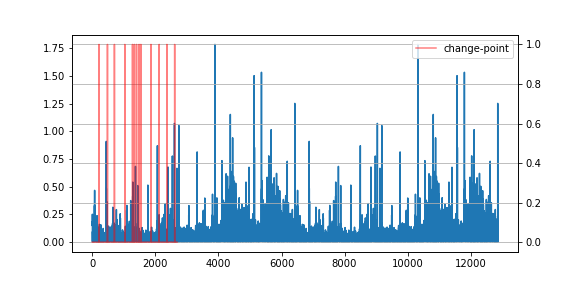}
         \caption{Subject 5}
         \label{}
     \end{subfigure}
     \hfill
          \begin{subfigure}[b]{0.3\textwidth}
         \centering
         \includegraphics[width=\textwidth]{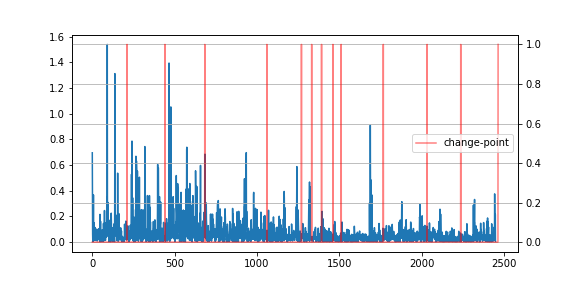}
         \caption{Subject 6}
         \label{}
     \end{subfigure}
     \hfill
     \caption{Change point detection statistic using the Frobenius distance and activity change points (in red) with a tolerance of $\pm 5$ timestamps for Subjects 1-6 a window size $L = 20$.}
    \label{fig:res_exp1_stat_fro}
\end{figure}

Additional results on ONCPD statistics are presented in  Figure~\ref{fig:res_exp1_stat_gnn}, Figure~\ref{fig:res_exp1_stat_cusum}, and Figure~\ref{fig:res_exp1_stat_fro}.

\subsection{Sensitivity to the tolerance level}

In this section, we investigate the sensitivity of our results presented in Table \ref{tab:res_exp1} to the tolerance level chosen to compute the adjusted F1-score. We consider the \textbf{Random split} experiment in the \textbf{Cross-Individual} task from Section \ref{sec:exp_physical}, and we reproduce this experiment for different levels of tolerance $\text{tol} = \{1,3,5,7\}$. We report the numerical results in Table \ref{tab:res_exp2_tolerance}. We note that our method has the best performance for all considered levels.  

\clearpage

\begin{table}
    \centering
    \begin{tabularx}{\textwidth}{ 
    >{\centering\arraybackslash}X 
   >{\centering\arraybackslash}X
   >{\centering\arraybackslash}X
   >{\centering\arraybackslash}X
   >{\centering\arraybackslash}X
   >{\centering\arraybackslash}X  }
          \toprule
 \multicolumn{6}{c}{\textbf{Cross-individual NCPD}} \\
  \midrule
         Tolerance   & s-GNN-I & Frobenius & SC-NCPD & CUSUM & CUSUM 2 \\
         \midrule
1 &  \textbf{0.60 (0.30)} &  0.53 (0.22) &  0.43 (0.32) &  0.33 (0.24) &  0.42 (0.30) \\
3 &  \textbf{0.87 (0.25)} &  0.68 (0.20) &  0.70 (0.31) &  0.53 (0.24) &  0.76 (0.29) \\
5 &  \textbf{0.61 (0.27)} &  0.53 (0.20) &  0.41 (0.32) &  0.27 (0.22) &  0.44 (0.31) \\
7 &  \textbf{0.85 (0.28)} &  0.71 (0.21) &  0.71 (0.30) &  0.56 (0.25) &  0.75 (0.29) \\
        \bottomrule
    \end{tabularx}
    \caption{Adjusted F1-score of our method (s-GNN) and  baselines in the \textbf{Cross-individual} NCPD task and the random split setting  on the physical activity monitoring data.  The bold values in each row  denote the top performing method. The values in the parentheses denote the standard deviation over 10 repetitions of the random splits train/validation/set. 
\mc{We remark that different rows corresponding to different tolerance levels essentially amounts to defining a different set of ground truth change-points, and hence we should not necessarily expect a monotonic relationship between tolerance versus the recovery accuracy of all methods; the main take-away message here is that the s-GNN method attains superior performance when compared to other baselines, for the same tolerance level.}}
    \label{tab:res_exp2_tolerance}
\end{table}

\end{appendix}

\end{document}